\documentclass[fleqn,10pt]{wlscirep}
\usepackage[utf8]{inputenc}
\usepackage[T1]{fontenc}

\usepackage[table]{xcolor}
\usepackage{multirow}
\usepackage{booktabs}
\usepackage{rotating}
\usepackage{float}
\usepackage{algorithm}
\usepackage{algpseudocode}
\usepackage{tcolorbox}
\usepackage{xurl}
\usepackage{hyperref} 
\usepackage{marvosym}
\usepackage{lineno}

\title{A Deployment-Friendly Foundational Framework for Efficient Computational Pathology}

\author[1]{Yu Cai}
\author[2]{Cheng Jin}
\author[3,4,5]{Zhengyu Zhang}
\author[2]{Jiabo Ma}
\author[2]{Fengtao Zhou}
\author[2]{Yingxue Xu}
\author[2]{Zhengrui Guo}
\author[2]{Yihui Wang}
\author[2]{Ling Liang}
\author[1]{Yonghao Tan}
\author[1]{Pingcheng Dong}

\author[6,7,8]{Du Cai}
\author[9]{On Ki Tang}
\author[3,11]{Chenglong Zhao}
\author[3]{Zhijian Cen}
\author[3]{Ying Tan}

\author[2]{Xi Wang}
\author[12]{Can Yang}

\author[13]{Yali Xu}
\author[11]{Jing Cui}
\author[14]{Zhenhui Li}
\author[9,10]{Ronald Cheong Kin Chan}
\author[15]{Yueping Liu}
\author[6,7,8]{Feng Gao}
\author[16]{Xiuming Zhang}
\author[3,4,5]{Li Liang}

\author[2,17,18,19,20,\Letter]{Hao Chen}
\author[1,2]{Kwang-Ting Cheng}

\affil[1]{Department of Electronic and Computer Engineering, The Hong Kong University of Science and Technology, Hong Kong SAR, China}
\affil[2]{Department of Computer Science and Engineering, The Hong Kong University of Science and Technology, Hong Kong SAR, China}

\affil[3]{Department of Pathology, Nanfang Hospital, School of Basic Medical Sciences, Southern Medical University, Guangzhou, China}
\affil[4]{Guangdong Province Key Laboratory of Molecular Tumor Pathology, Guangzhou, China}
\affil[5]{Jinfeng Laboratory, Chongqing, China}

\affil[6]{Department of General Surgery (Colorectal Surgery), The Sixth Affiliated Hospital, Sun Yat-sen University, Guangzhou, China}
\affil[7]{Guangdong Provincial Key Laboratory of Colorectal and Pelvic Floor Diseases, The Sixth Affiliated Hospital, Sun Yat-sen University, Guangzhou, China}
\affil[8]{Biomedical Innovation Center, The Sixth Affiliated Hospital, Sun Yat-sen University, Guangzhou, China}

\affil[9]{Department of Anatomical and Cellular Pathology, The Chinese University of Hong Kong, Hong Kong SAR, China}
\affil[10]{Pathology Artificial Intelligence Development and Assessment Laboratory, State Key Laboratory of Translational Oncology, The Chinese University of Hong Kong, Hong Kong SAR, China}

\affil[11]{Department of Pathology, The First Affiliated Hospital of Shandong First Medical University \& Shandong Provincial Qianfoshan Hospital, Ji’nan, China}
\affil[12]{Department of Mathematics, The Hong Kong University of Science and Technology, Hong Kong SAR, China}
\affil[13]{Department of Pathology, Shandong Provincial Hospital affiliated to Shandong First Medical University, Ji’nan, China}
\affil[14]{Department of Radiology, The Third Affiliated Hospital of Kunming Medical University, Yunnan Cancer Hospital, Kunming, China.}
\affil[15]{Department of Pathology, the Fourth Hospital of Hebei Medical University, Shijiazhuang, China}
\affil[16]{Department of Pathology, The First Affiliated Hospital, School of Medicine, Zhejiang University, Hangzhou, China}

\affil[17]{Department of Chemical and Biological Engineering, The Hong Kong University of
Science and Technology, Hong Kong SAR, China}
\affil[18]{Division of Life Science, The Hong Kong University of Science and Technology, Hong Kong SAR, China}
\affil[19]{State Key Laboratory of Nervous System Disorders, Hong Kong SAR, China}
\affil[20]{HKUST Shenzhen-Hong Kong Collaborative Innovation Research Institute, Futian, Shenzhen, China}

\affil[\Letter]{Corresponding author: jhc@ust.hk}
\begin{abstract} 
Pathology foundation models (PFMs) have enabled robust generalization in computational pathology through large-scale datasets and expansive architectures. However, the substantial computational cost of these models, particularly when analyzing gigapixel whole slide images, limits clinical accessibility and scalability. 
Here, we present \textbf{LitePath}, a deployment-friendly foundational framework designed to mitigate model over-parameterization and patch‑level redundancy. LitePath integrates \textbf{LiteFM}, a compact model distilled from three large PFMs (Virchow2, H-Optimus-1 and UNI2) using 190 million patches, and the \textbf{Adaptive Patch Selector (APS)}, a lightweight modular component for task-specific patch selection.
The framework reduces model parameters by 28$\times$ and lowers FLOPs by 403.5$\times$ relative to Virchow2, enabling deployment on low-power edge hardware such as the NVIDIA Jetson Orin Nano Super. On this device, LitePath achieves a processing speed of 208 slides per hour, 104.5$\times$ faster than Virchow2, and consumes 0.36 kWh per 3,000 slides, 171$\times$ lower than Virchow2 on a standard RTX 3090 GPU. % 
We validated accuracy using 45 multi-center clinical cohorts across four organs and 33 tasks, including 37 classification cohorts and 8 survival cohorts. These cohorts comprised 33 internal, 10 external and 2 prospective cohorts, with 17,837 slides from 9,977 patients disjoint from the pretraining data. Across all 45 cohorts, LitePath achieved the best average rank among 22 evaluated PFMs (6.56), slightly ahead of Virchow2 (6.58). Compared with Virchow2, LitePath retained 99.71\% of its Macro-AUC across 37 classification cohorts and achieved a 2.14-percentage-point higher mean C-index across 8 survival cohorts (71.91\% vs. 69.77\%).
To quantify the balance between accuracy and efficiency, we propose the Deployability Score (D-Score), a weighted geometric mean of normalized task performance and normalized FLOPs. LitePath obtained the highest overall D-Score (0.8455), exceeding H0-mini (0.7723) and Virchow2 (0.7297).
In a randomized paired crossover reader study involving four pathologists and 120 cases, LitePath assistance increased reader-level diagnostic accuracy by 4.1--15.8 percentage points and reduced mean case-level diagnostic time by 10.9--14.2\%.
Together, these results establish the technical feasibility of rapid, cost-effective and energy-efficient pathology image analysis on accessible hardware while maintaining competitive performance.
\end{abstract}

\begin{document}

% \flushbottom
\maketitle
 
\thispagestyle{empty}

% \linenumbers

\section*{Introduction}

\begin{figure}[hp]
\centering
\includegraphics[width=\linewidth]{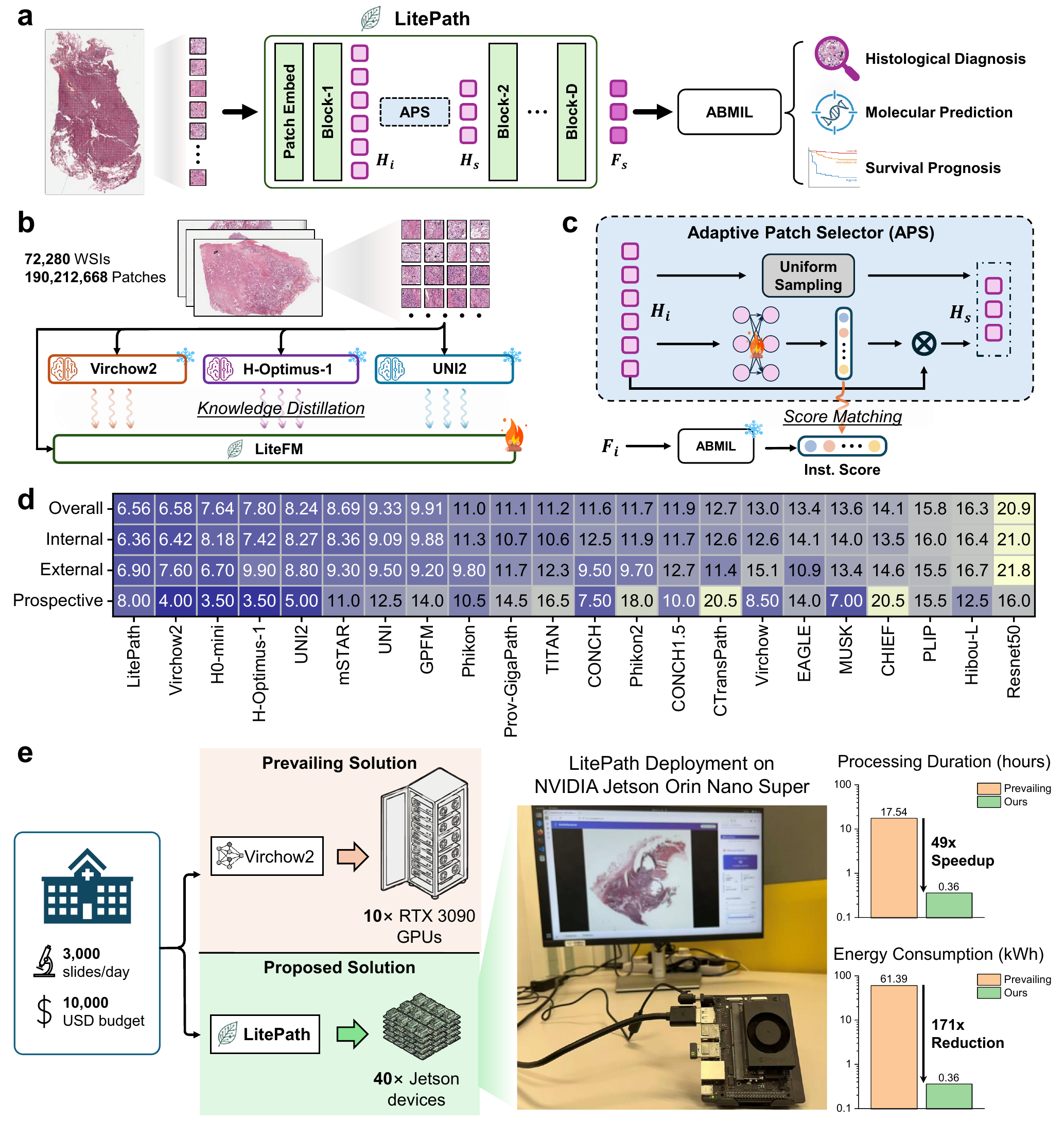}
\caption{\textbf{Overview of the LitePath framework.} LitePath is a deployment-friendly PFM framework designed to balance efficiency and diagnostic accuracy, consisting of LiteFM and APS.
\textbf{a}, The inference pipeline of LitePath. LiteFM extracts features, while APS selects patches based on indices and shallow features $\{\mathbf{H}_i\}_{i=1}^N$ from block-1. Only the selected features $\{\mathbf{H}_s\}_{s \in \mathcal{S}}$ are propagated through the network for final prediction.
\textbf{b}, LiteFM is distilled from Virchow2, H-Optimus-1 and UNI2 using approximately 190 million patches sourced from 72,280 WSIs.
\textbf{c}, APS combines uniform sampling and attention-based sampling for patch selection. The scoring network is trained on the shallow features to approximate the attention score distribution of the final ABMIL.
\textbf{d}, Average ranking scores based on task performance (Macro-AUC for classification and C-index for survival) for 22 PFMs across all cohorts, internal cohorts, external cohorts, and prospective cohorts, respectively.
\textbf{e}, Comparison of the prevailing deployment (Virchow2 on RTX 3090 GPUs) and the proposed solution (LitePath on Jetson Orin Nano Super devices) under an equivalent daily load and GPU budget.
}
\label{fig:main}
\end{figure}

Computational pathology (CPath) is a critical component of precision oncology, where deep learning approaches have improved diagnostic accuracy and clinical workflow efficiency. The confluence of whole slide image (WSI) digitization and developments in foundation models has driven the evolution of pathology foundation models (PFMs)\cite{Virchow2,gpfm,uni}. These models employ self-supervised learning strategies, such as DINOv2\cite{dinov2}, iBOT\cite{ibot}, and masked image modeling (MIM)\cite{mae}, to train on massive histopathology datasets. Recent extensions further enhance these models through multi-modal alignment with diagnostic reports and genomic data\cite{mstar}. Consequently, PFMs have achieved robust performance across various oncology applications ranging from tumor diagnosis to treatment planning and prognosis assessment.

The practical utility of PFMs, however, is constrained by the recurring computational cost of whole slide inference. Modern patch encoders require high-performance computing systems, substantial power and specialized infrastructure, while a gigapixel WSI typically contains tens of thousands of tissue patches. This burden arises from two coupled sources: \textit{(1) model overparameterization} and \textit{(2) patch-level redundancy}. At the model level, PathBench\cite{PathBench} showed that billion-parameter models, including H-Optimus-0 and Prov-GigaPath\cite{gigapath}, do not consistently outperform smaller architectures such as Virchow2\cite{Virchow2}, suggesting the potential to reduce model size while preserving downstream performance. At the slide level, both diagnostic workflows and attention-based multiple-instance learning\cite{abmil} indicate that predictions may depend on a limited subset of tissue regions. Nevertheless, conventional PFM pipelines encode all tissue patches indiscriminately, expending substantial computation on regions with limited relevance to the downstream task. Practical deployment therefore requires reducing both the computation applied to each patch and the number of patches undergoing full encoding. 

Existing efficiency strategies generally address only part of these bottlenecks. At the model level, H0-mini demonstrates that distillation can compress H-Optimus-0 into an 86M-parameter ViT-Base PFM\cite{h0_mini}, but it does not address patch redundancy, and the practical lower bound of such distillation remains unclear. At the slide level, Keshvarikhojasteh et al.~\cite{keshvarikhojasteh2024multiple} showed that sparse random sampling during training can preserve WSI-level performance, but all patches are still processed at inference, thereby not reducing inference-stage computational cost. Fu et al.~\cite{fu2024whole} explored self-learning sampling for WSI classification, but sampling is performed after full-patch encoding, leaving the dominant feature extraction cost unchanged. Other approaches used task-agnostic pre-screening\cite{neidlinger2025deep} or reduced-magnification inputs\cite{han2025towards}. EAGLE\cite{neidlinger2025deep}, for example, uses frozen CHIEF-derived representations to select a fixed number of 25 patches for subsequent Virchow2 encoding. Reduced-magnification methods\cite{han2025towards} lower the input burden by applying a common lower resolution, which may be less suitable for endpoints that depend on fine-scale morphology. Collectively, these studies suggest the feasibility of model compression and sparse patch processing, but do not fully eliminate both sources of recurring PFM inference cost.

To address these issues, we developed LitePath (Fig.~\ref{fig:main}a), a PFM framework that addresses model overparameterization and patch-level redundancy through LiteFM and the Adaptive Patch Selector (APS), respectively. LiteFM is a compact ViT-Small encoder \cite{vit,deit} obtained via knowledge distillation\cite{hinton2015distilling} (Fig.~\ref{fig:main}b). Drawing upon findings from PathBench \cite{PathBench}, we selected Virchow2\cite{Virchow2}, H-Optimus-1 and UNI2\cite{uni} as teachers based on their complementary performance across histological diagnosis, molecular prediction, and survival prognosis. To facilitate effective knowledge transfer, the distillation pretraining was performed on approximately 190 million patches extracted from 72,280 publicly available WSIs. APS is a lightweight, plug-and-play, and task-specific module that uses indices and shallow features from the first LiteFM block to guide patch selection (Fig.~\ref{fig:main}c). Its hybrid strategy combines uniform sampling for broad spatial coverage with attention-based sampling to prioritize potentially informative regions. Only the selected patches pass through the remaining LiteFM blocks, reducing full-encoding cost while retaining task-specific attention guidance. This design parallels the broad screening and focused inspection used in pathological examination.

Through the synergy of LiteFM and APS, LitePath achieves high efficiency while maintaining competitive performance (Fig.~\ref{fig:results_overall}e,f). With 22.5M parameters (28$\times$ smaller than Virchow2 and 50$\times$ smaller than H-Optimus-1) (Fig.~\ref{fig:results_efficiency}a) and efficient computation (Fig.~\ref{fig:results_efficiency}b--d), LitePath enables deployment on the NVIDIA Jetson Orin Nano Super (Fig.~\ref{fig:main}e), a user-friendly edge device with a 25W power rating and a price of USD \$249\cite{jetson}. In the standardized 30,000-patch WSI-equivalent benchmark, LitePath processed 208 WSI equivalents per hour on this device, 104.5$\times$ faster than Virchow2 under the same hardware setting (Fig.~\ref{fig:results_efficiency}d). Under the equivalent-cost deployment scenario, LitePath running on Jetson devices consumed 0.36 kWh per 3,000 WSI equivalents, 171$\times$ less energy than Virchow2 running on RTX 3090 GPUs (Fig.~\ref{fig:main}e). Across 45 cohorts (33 internal, 10 external, and 2 prospective cohorts) with 33 tasks, LitePath achieved the best average rank among 22 evaluated PFMs (6.56; Virchow2: 6.58; Fig.~\ref{fig:main}d), retained 99.71\% of Virchow2's Macro-AUC across 37 classification cohorts (Fig.~\ref{fig:results_overall}h), and achieved a 2.14-percentage-point higher mean C-index than Virchow2 across 8 survival cohorts (0.7191 vs. 0.6977; Fig.~\ref{fig:results_overall}i). To quantitatively evaluate the balance of task performance and computational efficiency, we defined the Deployability Score (D-Score) (Section \hyperref[sec:metrics]{Evaluation Metrics}) as the weighted geometric mean of normalized task performance and normalized FLOPs scores. LitePath achieved the highest average D-Score across the benchmark (0.8455, Fig.~\ref{fig:results_overall}g), exceeding H0-mini (0.7723) and Virchow2 (0.7297). These results support resource-efficient inference for the evaluated tasks. A randomized paired crossover reader study involving four pathologists and 120 cases further showed reader-level accuracy increases of 4.1--15.8 percentage points and mean case-level diagnostic-time reductions of 10.9--14.2\% with LitePath assistance (Fig.~\ref{fig:reader_study}).

\section*{Results}

\begin{figure}[!t] 
\centering 
\includegraphics[width=\linewidth]{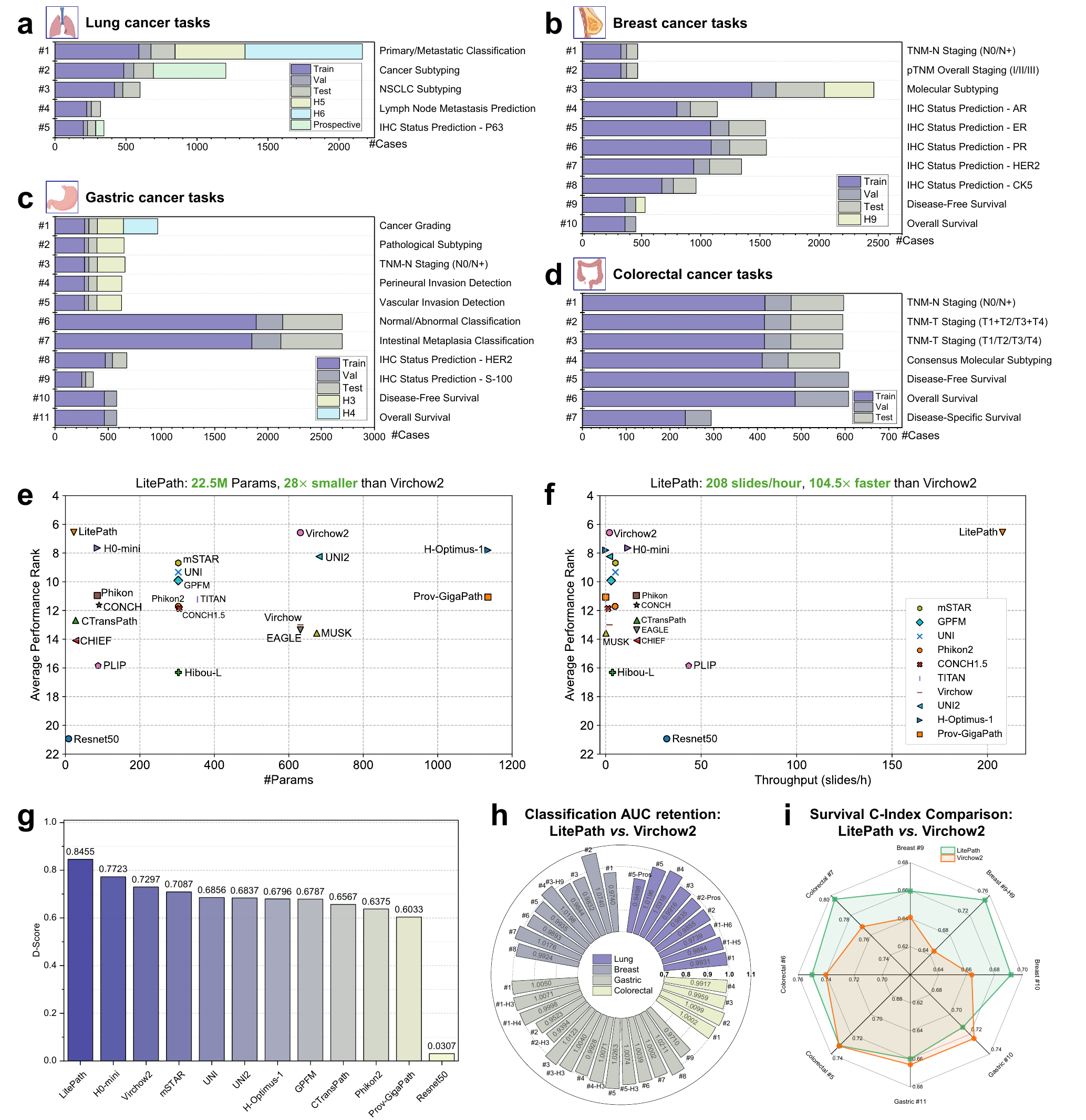} 
\caption{\textbf{Overall assessment of PFMs.} 
\textbf{a--d}, Composition of the multi-center evaluation dataset for lung, breast, gastric, and colorectal cancers, respectively (H: Hospital). 
\textbf{e}, Trade-off between the average task-performance rank and model parameters. \textbf{f}, Trade-off between the average task-performance rank and throughput. 
\textbf{g}, Average D-Score of the PFMs. 
\textbf{h}, AUC retention of LitePath relative to Virchow2 across 37 classification cohorts. 
\textbf{i}, C-index comparison of LitePath and Virchow2 across 8 survival cohorts. (\#[No.]: task identifiers in the corresponding organ, with detailed mappings provided in panels a--d. Pros: prospective. Hospital identifiers are omitted for internal evaluation cohorts.)} 
\label{fig:results_overall} 
\end{figure}

\begin{figure}[!t]
\centering
\includegraphics[width=\linewidth]{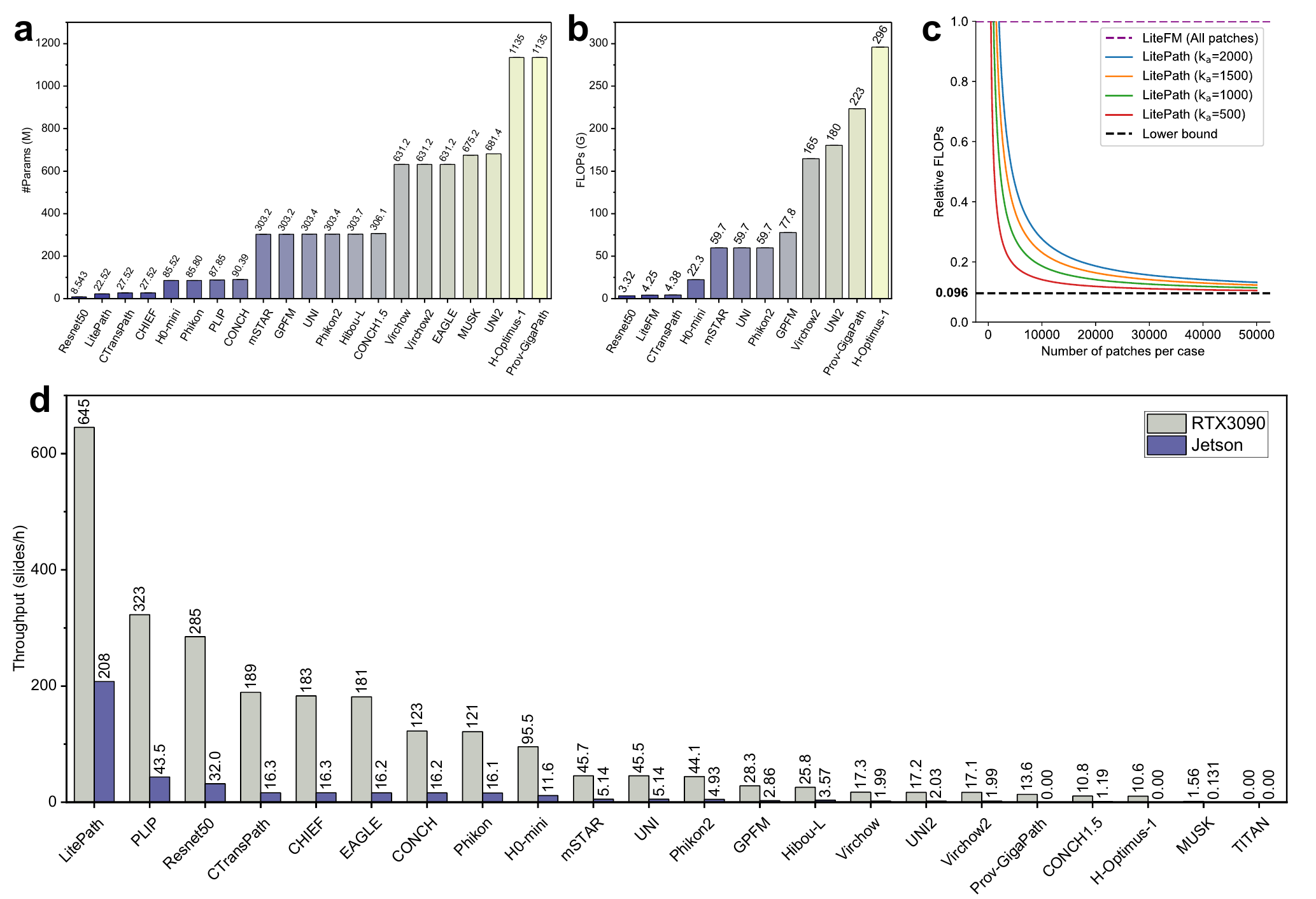}
\caption{\textbf{Efficiency comparison of PFMs.}
\textbf{a}, Number of parameters for each PFM. LitePath (including LiteFM and APS modules) contains $28\times$ fewer parameters than Virchow2.
\textbf{b}, Floating-point operations (FLOPs) required by each PFM to process a single input pathology patch. LiteFM requires $38.8\times$ fewer FLOPs than Virchow2. 
\textbf{c}, Relative FLOPs of LitePath compared to LiteFM as a function of the number of patches per case. Only results for attention-based sampling are presented, as uniform-based sampling scales FLOPs in a straightforward linear fashion. For a sufficiently large number of patches, LitePath achieves a relative FLOPs convergence to 0.096, representing a $10.4\times$ reduction. Therefore, LitePath can theoretically deliver up to a 403.5-fold ($38.8\times10.4$) reduction in computational cost compared to Virchow2. 
\textbf{d}, Throughput of PFMs on RTX 3090 and Jetson Orin Nano Super GPUs, evaluated using dummy slides containing 30,000 patches with half-precision computation. ``0.00'' indicates out-of-memory. For these experiments, APS selected 1,000 patches using attention-based strategy, with no uniform selection performed.}
\label{fig:results_efficiency}
\end{figure}

The LitePath framework was evaluated for its ability to balance efficiency and diagnostic accuracy. To assess performance across clinical tasks, we benchmark 22 PFMs based on their overall trade-off between efficiency and accuracy (Fig.~\ref{fig:results_overall}), as well as on efficiency (Fig.~\ref{fig:results_efficiency}) and accuracy (Fig.~\ref{fig:results_accuracy}) individually.

\subsection*{Overall Assessment}
To evaluate LitePath, this study leverages diverse pathology datasets and tasks alongside real-world deployment metrics. Specifically, we assessed 22 PFMs using 33 distinct tasks covering lung, breast, gastric, and colorectal cancers, incorporating 33 internal evaluation cohorts, 10 external cohorts, and 2 prospective cohorts (Fig.~\ref{fig:results_overall}a--d). The evaluated PFMs include LitePath, H-Optimus-1, H0-mini\cite{h0_mini}, EAGLE\cite{neidlinger2025deep}, UNI2\cite{uni}, MUSK\cite{MUSK}, CONCH1.5\cite{CONCH}, TITAN\cite{TITAN}, Phikon2\cite{Phikon2}, CHIEF\cite{CHIEF}, Virchow2\cite{Virchow2}, GPFM\cite{gpfm}, mSTAR\cite{mstar}, Hibou-L\cite{hibou}, Prov-GigaPath\cite{gigapath}, Virchow\cite{virchow}, UNI\cite{uni}, PLIP\cite{plip}, Phikon\cite{Phikon}, CONCH\cite{CONCH}, CTransPath\cite{ctranspath}, and Resnet50\cite{he2016deep}. The multi-center benchmark contains 17,837 slides from 9,977 patients across nine hospitals and is disjoint from public pretraining data, ensuring an impartial evaluation. Classification cohorts were evaluated using Macro-AUC, whereas survival cohorts were evaluated using C-index. Average task-performance ranks were computed by ranking PFMs within each cohort according to the corresponding metric and then averaging the ranks across cohorts. To emphasize practical relevance, PFMs were further deployed on the affordable and accessible NVIDIA Jetson device to measure throughput. We assess deployability by analyzing the trade-offs between average ranks and model parameters (Fig.~\ref{fig:results_overall}e), as well as between average ranks and throughput (Fig.~\ref{fig:results_overall}f). With 22.5M parameters (28$\times$ smaller than Virchow2 and 50$\times$ smaller than H-Optimus-1) and a throughput of 208 slides per hour on a Jetson device (104.5$\times$ faster than Virchow2), LitePath achieves the best average rank across all 45 cohorts (6.56), slightly ahead of Virchow2 (6.58). Additionally, we propose a new metric called D-Score to evaluate deployability by balancing task performance and computational efficiency (Section \hyperref[sec:metrics]{Evaluation Metrics}, Extended Data Table~\ref{tab:dscore}). The D-Score is a weighted geometric mean of normalized task-performance and normalized FLOPs scores, where a score of zero is assigned if the model ranks last in task performance and penalties are applied for excessive computational overhead. For comparison, the average D-Score across all cohorts is computed for LitePath and 11 other top-performing or most efficient PFMs (Fig.~\ref{fig:results_overall}g). LitePath achieved the highest average D-Score (0.8455), exceeding H0-mini (0.7723) and Virchow2 (0.7297). Given the importance of accuracy in clinical scenarios, we further compare LitePath with Virchow2 across classification and survival cohorts. Across the 37 classification cohorts, LitePath outperforms Virchow2 on 17 cohorts and retains 99.71\% of Virchow2's Macro-AUC on average (Fig.~\ref{fig:results_overall}h). Across the 8 survival cohorts, LitePath achieves a higher mean C-index than Virchow2 (0.7191 vs. 0.6977; Fig.~\ref{fig:results_overall}i). These findings support resource-efficient inference for the evaluated task-specific pipelines while maintaining competitive task performance.

\subsection*{Efficiency}
While numerous PFMs have been developed to facilitate AI-assisted precision oncology and have achieved impressive generalizability in diverse relevant tasks, the practical intractability of deploying such large models to analyze gigapixel WSIs has been largely overlooked. Take the case of Virchow2, a SOTA PFM with moderate computational demands among the evaluated PFMs: it processes 17.1 slides per hour on an RTX 3090 GPU (Fig.~\ref{fig:results_efficiency}d), which translates to more than 175 GPU hours to handle a tertiary hospital’s daily load of 3,000 slides. Consequently, investigating the computational efficiency of PFMs is essential for determining their clinical relevance. To this end, we analyze each PFM's parameter count (Fig.~\ref{fig:results_efficiency}a) and the required FLOPs for processing a single input pathology patch (Fig.~\ref{fig:results_efficiency}b). To illustrate the properties of the proposed APS, we present the relative FLOPs of LitePath (equipped with APS) compared to LiteFM (using all patches) as a function of the number of patches per case (Fig.~\ref{fig:results_efficiency}c). Moreover, PFMs were benchmarked on RTX 3090 and Jetson Orin Nano Super devices to assess hardware-level inference efficiency. For consistency, we measure their throughput (slides processed per hour) on CPU-generated dummy slides containing 30,000 patches each, with computations performed in half-precision (FP16) (Fig.~\ref{fig:results_efficiency}d). The detailed analysis is as follows.

\paragraph{LitePath reduces theoretical overhead by two orders of magnitude compared to Virchow2.} The model size and computational complexity represent key factors that determine the practical cost and speed of PFMs. To illustrate these characteristics, we report the number of parameters for the 22 PFMs (Fig.~\ref{fig:results_efficiency}a) and the theoretical FLOPs for the nine top-ranking PFMs as well as ResNet50 (Fig.~\ref{fig:results_efficiency}b). LitePath contains 22.5M parameters, making it 28 times smaller than Virchow2 (631M params) and 50 times smaller than H-Optimus-1 (1135M params). In terms of computational complexity, LiteFM (without APS) requires 4.25G FLOPs per input image, which is 38.8 times lower than Virchow2 (165G FLOPs) and 69.6 times lower than H-Optimus-1 (296G FLOPs). APS further minimizes computational overhead through hybrid sampling strategies: uniform sampling, which yields a linear reduction in FLOPs, and attention‑based sampling, where the relative FLOPs decrease inversely with the total number of patches per case (Fig.~\ref{fig:results_efficiency}c). For sufficiently large patch counts, the relative FLOPs of attention-based sampling in APS converge to 0.096, representing a 10.4-fold reduction. Consequently, by leveraging attention-based sampling in APS, LitePath can theoretically deliver up to a 403.5-fold (38.8 $\times$ 10.4) reduction in computational cost compared to Virchow2. Notably, in some cases, uniform sampling alone is sufficient to preserve accuracy, allowing the relative FLOPs to decrease linearly and approach smaller values.

\paragraph{LitePath enables efficient deployment on resource-constrained edge devices.} To quantify practical efficiency, PFMs were deployed on an NVIDIA RTX 3090 and an NVIDIA Jetson Orin Nano Super to evaluate performance under constrained hardware conditions. The Jetson device functions as a representative low-power edge platform (25W rated power, 8GB unified memory). Evaluating model performance on such a device provides a direct measure of deployability for widespread adoption in real-world settings. For this evaluation, we generate dummy slides containing 30,000 patches each and apply half‑precision (FP16) computation to measure PFM throughput (Fig.~\ref{fig:results_efficiency}d). LitePath achieves a throughput of 645 slides per hour on the RTX 3090 and 208 slides per hour on the Jetson device. On the Jetson device, LitePath (208 slides/h) is 104.5 times faster than Virchow2 (1.99 slides/h). Moreover, LitePath's throughput on the Jetson device is 12.2 times higher than Virchow2’s throughput on the RTX 3090 (17.1 slides/hour). Under the standardized 30,000-patch benchmark, LitePath was the only evaluated model whose measured edge-device throughput exceeded a reference load of 3,000 slides per day. In contrast, the lower throughput of the other PFMs limited their feasibility on the tested edge platform under the same standardized setting.

\paragraph{LitePath deployed on edge devices presents a scalable solution for high-speed, economical, and energy-efficient utilization of PFMs in clinical settings.} To characterize system-level scalability, we evaluated processing speed, cost-effectiveness and energy efficiency. High processing speed is necessary for timely analysis of routinely generated slides; cost-effectiveness enables broader adoption in remote and economically disadvantaged regions; and enhanced energy efficiency reduces carbon emissions and supports global environmental sustainability, particularly as adoption scales. To model a hospital-scale processing scenario under a constrained GPU budget, we compare these metrics using an equivalent-cost scenario (Fig.~\ref{fig:main}e). A budget sufficient to procure 10 RTX 3090 GPUs could instead be allocated to approximately 40 Jetson devices. Under this budget, LitePath deployed on 40 Jetson devices completes the analysis of 3,000 slides in 0.36 hours, which is 49 times faster than Virchow2 running on 10 RTX 3090 GPUs (17.54 hours). In terms of energy consumption, LitePath deployed on the Jetson devices consumes 0.36 kWh to process 3,000 slides, representing a 171 times reduction compared with the 61.39 kWh required by Virchow2 on the RTX 3090 GPUs.

\begin{figure}[!t]
\centering
\includegraphics[width=\linewidth]{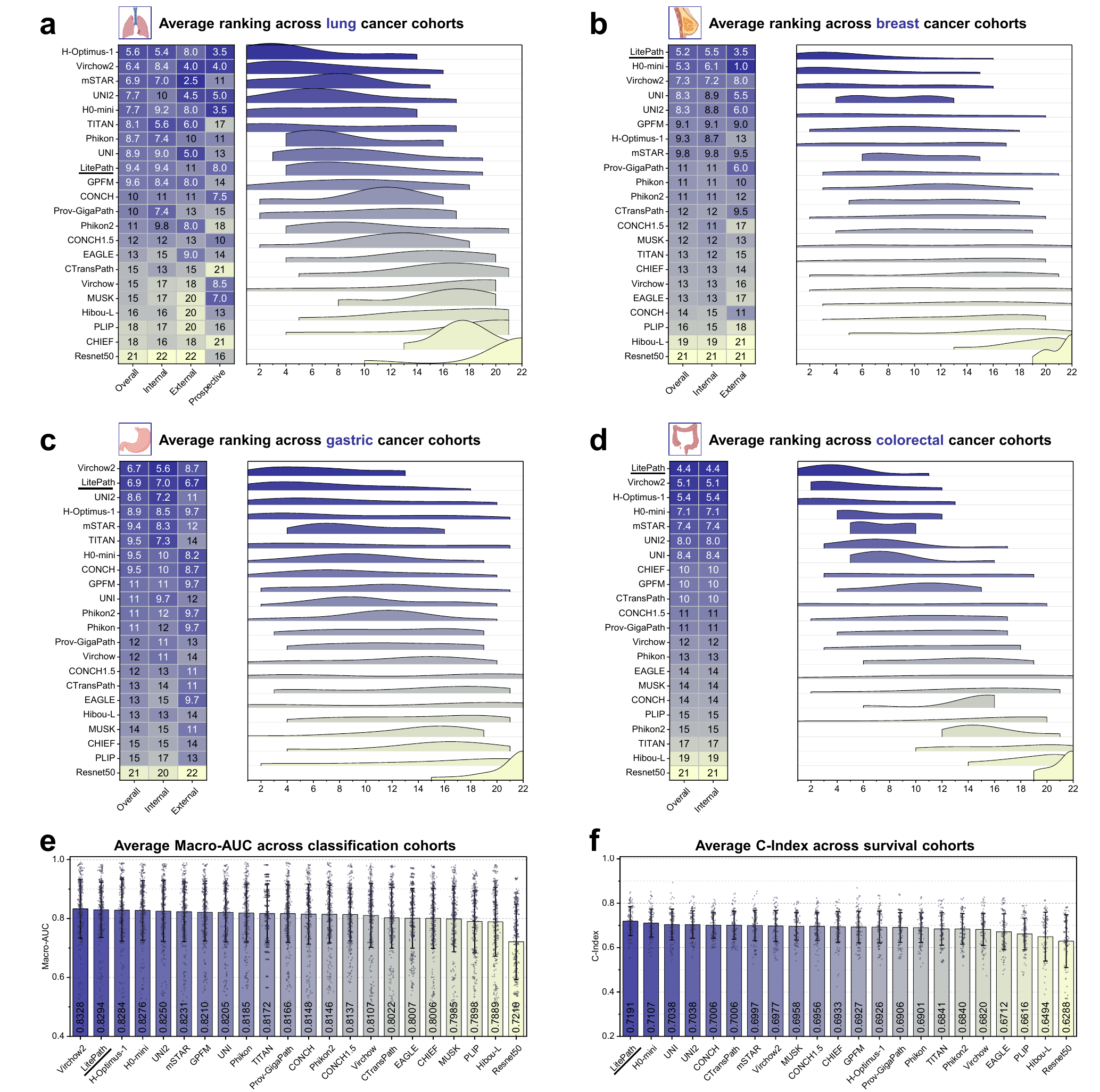}
\caption{\textbf{Overall accuracy of PFMs across four organs.} 
\textbf{a--d}, Performance of PFMs on lung, breast, gastric, and colorectal cancers, respectively. Each organ panel includes average ranking scores for all cohorts, internal cohorts, external cohorts (if applicable), and prospective cohorts (if applicable), together with the distribution of ranking scores for each PFM. 
\textbf{e}, Average Macro-AUC across classification cohorts. 
\textbf{f}, Average C-index across survival cohorts.}
\label{fig:results_accuracy}
\end{figure}

\subsection*{Accuracy}
To validate accuracy, we conduct experiments on 33 distinct tasks, incorporating 33 internal evaluation cohorts, 10 external cohorts, and 2 prospective cohorts (Fig.~\ref{fig:results_overall}a--d). The datasets cover four organs: lung, breast, gastric, and colorectal cancers. Lung cancer is the most common cancer worldwide in terms of both incidence and mortality\cite{cao2024comparative}. Breast cancer is the second most commonly diagnosed cancer worldwide and remains a leading cause of cancer-related mortality among women\cite{harbeck2019breast}. Gastric cancer is particularly prevalent in East Asia and is the fifth most commonly diagnosed cancer worldwide\cite{smyth2020gastric}. Colorectal cancer (CRC) is the third most commonly diagnosed cancer worldwide and a leading cause of cancer deaths\cite{marmol2017colorectal}, predominantly affecting individuals over 50. Macro-AUC is used for classification cohorts and C-index is used for survival cohorts. To assess overall PFM accuracy, we report average task performance, average ranking score, and the distribution of ranking scores across cohorts (Fig.~\ref{fig:results_accuracy}). The detailed Macro-AUC and C-index values with 95\% confidence intervals (CIs) for PFMs on each cohort are presented in Extended Data Figs.~\ref{fig:auc_hist1},\ref{fig:auc_hist2} and Tables~\ref{tab:nanfang_primary_metastatic}--\ref{tab:colorectal_survival}.

\paragraph{LitePath demonstrates comparable performance to state-of-the-art PFMs across four cancer types.} Across all 45 cohorts, LitePath achieves the best average rank among 22 PFMs (6.56), narrowly ahead of Virchow2 (6.58) and H0-mini (7.64) (Fig.~\ref{fig:main}d). Organ-specific analysis shows that LitePath performs particularly well in breast and colorectal cancer tasks, ranking first in both breast cancer (average rank: 5.2) and colorectal cancer (average rank: 4.4) (Fig.~\ref{fig:results_accuracy}b,d). In gastric cancer, LitePath ranks second (6.9), closely following Virchow2 (6.7) (Fig.~\ref{fig:results_accuracy}c). Lung cancer remains the most challenging setting for LitePath, where it ranks ninth (9.4) among 22 PFMs (Fig.~\ref{fig:results_accuracy}a). At the metric level, LitePath retains 99.71\% of Virchow2's Macro-AUC across the 37 classification cohorts and achieves a higher mean C-index than Virchow2 across the 8 survival cohorts (0.7191 vs. 0.6977). In summary, LitePath maintains performance comparable to state-of-the-art PFMs while offering substantially improved computational efficiency.

\paragraph{Prospectively collected cohorts provide task-specific tests of frozen LitePath pipelines.} The frozen LitePath pipelines were applied without retraining or hyperparameter tuning to two cohorts prospectively accrued at H1. For lung cancer subtyping, Macro-AUC was 0.8133 in the prospective cohort, compared with 0.8824 in the corresponding internal cohort. For P63 prediction, the respective values were 0.8716 and 0.8041 (Extended Data Tables~\ref{tab:nanfang_lung_finegrained} and~\ref{tab:nanfang_lung_p63}). Because only two tasks were evaluated and the subtyping cohorts differed substantially in class composition, these point-estimate differences should not be interpreted as a general improvement or decline under temporal shift. Instead, the results provide task-specific evidence that predefined LitePath pipelines can be applied to chronologically accrued data without model adaptation.

\paragraph{LitePath demonstrates strong potential for intraoperative analysis using frozen slides.} Accurate and rapid diagnosis of frozen sections is essential for guiding real-time surgical decision-making, making intraoperative pathology one of the most critical and demanding application scenarios. LitePath's highly efficient design is particularly suited for this environment, where time constraints and limited computational resources present significant challenges. To evaluate its performance in this context, we included a task focused on frozen sections: lymph node metastasis prediction in lung cancer (Fig.~\ref{fig:results_overall}a, Section~\hyperref[sec:lung_cancer]{Lung Cancer}). On this task, LitePath achieves an AUC of 77.29\% (95\% CI: 64.42\%--88.48\%), outperforming Virchow2 (74.90\%, 95\% CI: 59.07\%--88.11\%) and UNI2 (74.49\%, 95\% CI: 59.60\%--87.26\%) (Extended Data Table~\ref{tab:nanfang-lung-nsclc-lymph}). Balancing both accuracy and efficiency, LitePath attains a D-Score of 95.05\% (Extended Data Table~\ref{tab:dscore}), ranking first on this task. These results underscore LitePath's capability to deliver reliable AI-assisted intraoperative diagnoses.

\paragraph{LitePath maintains competitive performance in pre-cancer diagnosis.} Differentiating normal from abnormal tissue is crucial for early detection and prevention of malignant transformation, a process that requires large-scale screening of non-cancer cases. To assess LitePath's capabilities in this scenario, we conducted a normal/abnormal classification task on gastric tissue, using non-cancer slides that include diverse precancerous lesions such as Helicobacter pylori-associated chronic gastritis (HPACG), Autoimmune chronic gastritis with Helicobacter pylori (ACGHP), polyps, and ulcers (Fig.~\ref{fig:results_overall}c, Section~\hyperref[sec:gastric_cancer]{Gastric Cancer}). LitePath achieves an AUC of 92.68\% (95\% CI: 90.13\%--94.94\%), ranking second among 22 PFMs, with the best-performing model being H-Optimus-1 (93.19\%, 95\% CI: 90.17\%--95.59\%) (Extended Data Table~\ref{tab:stomach_four_tasks_h7}). With similar AUC and lower computational cost, LitePath achieved a higher D-Score than H-Optimus-1 (91.86\% vs 83.65\%). These results highlight LitePath's strength as a practical and effective tool for pre-cancer screening and early intervention.

\paragraph{LitePath generalizes competitively across open-access slide-level tasks, while LiteFM preserves patch-level representation quality.}
To assess generalization beyond the multi-center benchmark, we separately evaluated the models on 11 open-access slide-level tasks. Because these datasets may overlap with PFM pretraining corpora, they were excluded from the aggregate clinical ranking (Extended Data Fig.~\ref{fig:auc_hist3}a--d; Extended Data Tables~\ref{tab:public_bracs_subtyping}--\ref{tab:public_survival}). Across seven classification tasks, LitePath achieved an average Macro-AUC of 0.8653, comparable to H-Optimus-1 (0.8667) and LiteFM (0.8648). Across four survival tasks, LitePath achieved an average C-index of 0.6593, compared with 0.6613 for H-Optimus-1 and 0.6590 for LiteFM. Across all 11 tasks, LitePath ranked second (average rank, 5.0), after H-Optimus-1 (4.7), while LiteFM achieved an average rank of 5.7. We further assessed patch-level representation quality using linear probes on five ROI classification tasks. LiteFM achieved an average Macro-AUC of 0.9288, ranking third among 20 patch-level PFMs and closely matching UNI2 (0.9341) and Virchow2 (0.9332). These results indicate that the compact encoder retains competitive patch-level representations independently of APS and ABMIL (Extended Data Fig.~\ref{fig:auc_hist3}e; Extended Data Table~\ref{tab:roi_classification}).

\begin{figure}[hp]
\centering
\includegraphics[width=0.96\linewidth]{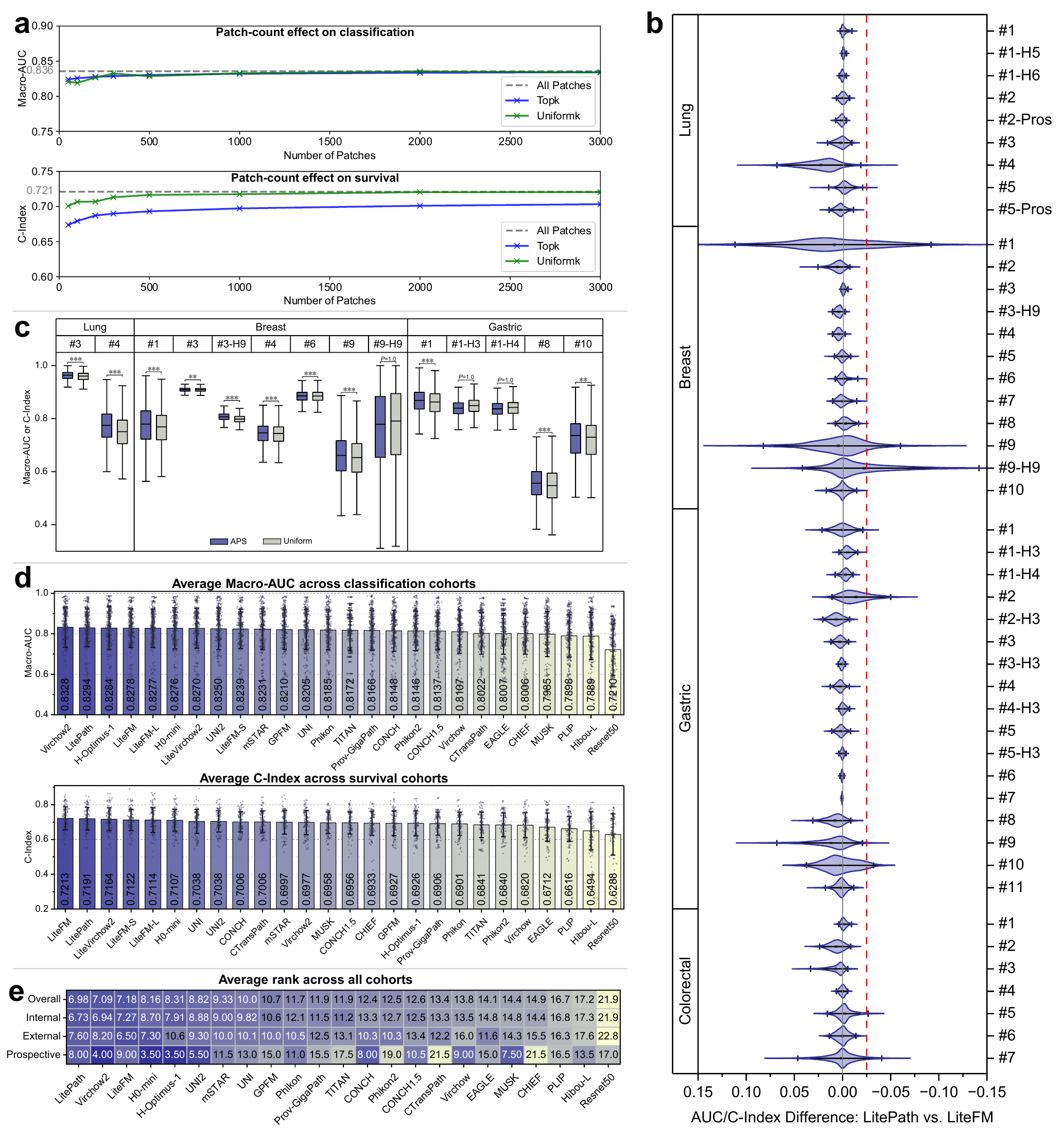}
\caption{\textbf{Ablation study.} 
\textbf{a}, Comparison of full-patch inference with top-$k$ patch selection based on final ABMIL attention scores and uniform-$k$ patch sampling. The average performance metric across internal evaluation cohorts is presented for classification and survival tasks, respectively. Task-wise results are provided in Extended Data Fig.~\ref{fig:topk}. 
\textbf{b}, Non-inferiority test of the Adaptive Patch Selector (APS). Violin plots show the distribution of performance differences between LitePath with APS and LiteFM using all patches. Mean differences and 95\% confidence intervals (CIs) of AUC or C-index are shown. The zero line and non-inferiority margin (-2.5\%) are indicated by gray and red dashed lines, respectively. Task identifier mappings are provided in Fig.~\ref{fig:results_overall}a--d. 
\textbf{c}, Comparison of APS and uniform sampling strategies. APS denotes the proposed hybrid strategy combining uniform and attention-based sampling, whereas the uniform strategy selects the same number of patches without attention guidance. Statistical significance was assessed using a paired one-sided Wilcoxon signed-rank test between APS and the uniform sampling strategy. Asterisks denote statistical significance: * $P < 0.05$, ** $P < 0.01$, and *** $P < 0.001$. 
\textbf{d}, Performance comparison of LiteFM variants with other PFMs. 
\textbf{e}, Average ranks comparison of LitePath, LiteFM, and other PFMs.}
\label{fig:ablation}
\end{figure}

\subsection*{Ablation Study}
To clarify the motivation and effectiveness of our method, we conducted ablation studies to assess the impact of partial-patch inference (Fig.~\ref{fig:ablation}a), APS-based patch reduction (Fig.~\ref{fig:ablation}b), attention-guided selection (Fig.~\ref{fig:ablation}c), and LiteFM distillation configurations (Fig.~\ref{fig:ablation}d,e) on model performance.

\paragraph{WSIs contain substantial patch redundancy.} WSIs typically comprise tens of thousands of patches, of which pathologists routinely examine only a subset for diagnosis. To evaluate how well MIL frameworks operate when given partial inputs, we performed inference with well-trained ABMIL models using a limited number of patches per slide, selected either by top attention scores or by uniform spatial sampling (Fig.~\ref{fig:ablation}a). Across 26 classification tasks and 7 survival tasks, we consistently observed that using thousands of patches per slide often yields performance comparable to using the full set of patches. This finding shows that computational cost can be reduced by roughly an order of magnitude via patch selection without sacrificing PFM performance. Notably, uniformly sampled patches outperform top‑score patches on several tasks (for example, Consensus Molecular Subtyping of colorectal cancer, Extended Data Fig.~\ref{fig:topk}d). This counterintuitive result highlights a limitation of ABMIL: attention weights can be biased and may fail to identify diagnostically informative regions in WSIs that lack obvious focal lesions. Collectively, these results motivate combining uniform and attention‑based sampling strategies to enable efficient patch selection that substantially lowers computation while preserving predictive performance.

\paragraph{APS preserves task performance after discarding redundant patches.} To evaluate whether APS preserves task performance after aggressive patch reduction, we performed non-inferiority tests on the distribution of Macro-AUC or C-index differences between LitePath (with APS) and LiteFM (without APS) (Fig.~\ref{fig:ablation}b). 
We pre-specified the non-inferiority margin at -2.5\% to ensure that any performance degradation remains clinically negligible. This threshold is more stringent than the 4\% diagnostic discordance rate typically accepted in digital pathology validation protocols as having no impact on patient management \cite{mukhopadhyay2018whole,bauer2013validation}.
The results show that APS successfully passes the non-inferiority test on 35 of 37 classification cohorts, indicating that accuracy loss introduced by patch reduction is generally acceptable, even in extreme cases. APS fails the test on two cohorts, where the lower bound of the CI falls below the non-inferiority margin: TNM‑N staging (N0/N+) of breast cancer on Internal‑H2 (Breast \#1), and Pathological Subtyping of gastric cancer on Internal‑H1 (Gastric \#2). For the breast cohort, although the lower bound of the confidence interval falls below the non-inferiority margin, the mean AUC difference is positive, suggesting that the point estimate of AUC does not decrease; the test failure likely reflects the high sensitivity of this task and the resulting variability. For the gastric subtyping task, APS fails on Internal‑H1 but passes and increases AUC on External‑H3 (Gastric \#2-H3), indicating that the observed failures are not systematic across cohorts.
At the cohort level, the overall mean AUC difference across the 37 classification cohorts is 0.16\% (95\% CI: -0.63\% to 1.29\%), with positive mean differences in 23 cohorts and negative mean differences in 14 cohorts. Across the eight survival cohorts, the mean C-index was 0.7191 for LitePath and 0.7213 for full-patch LiteFM, corresponding to a mean difference of -0.22 percentage points. Together, these results indicate that APS generalizes well across tasks and cohorts: it substantially reduces the number of inference patches while preserving overall task performance.

\begin{figure}[!t]
\centering
\includegraphics[width=\linewidth]{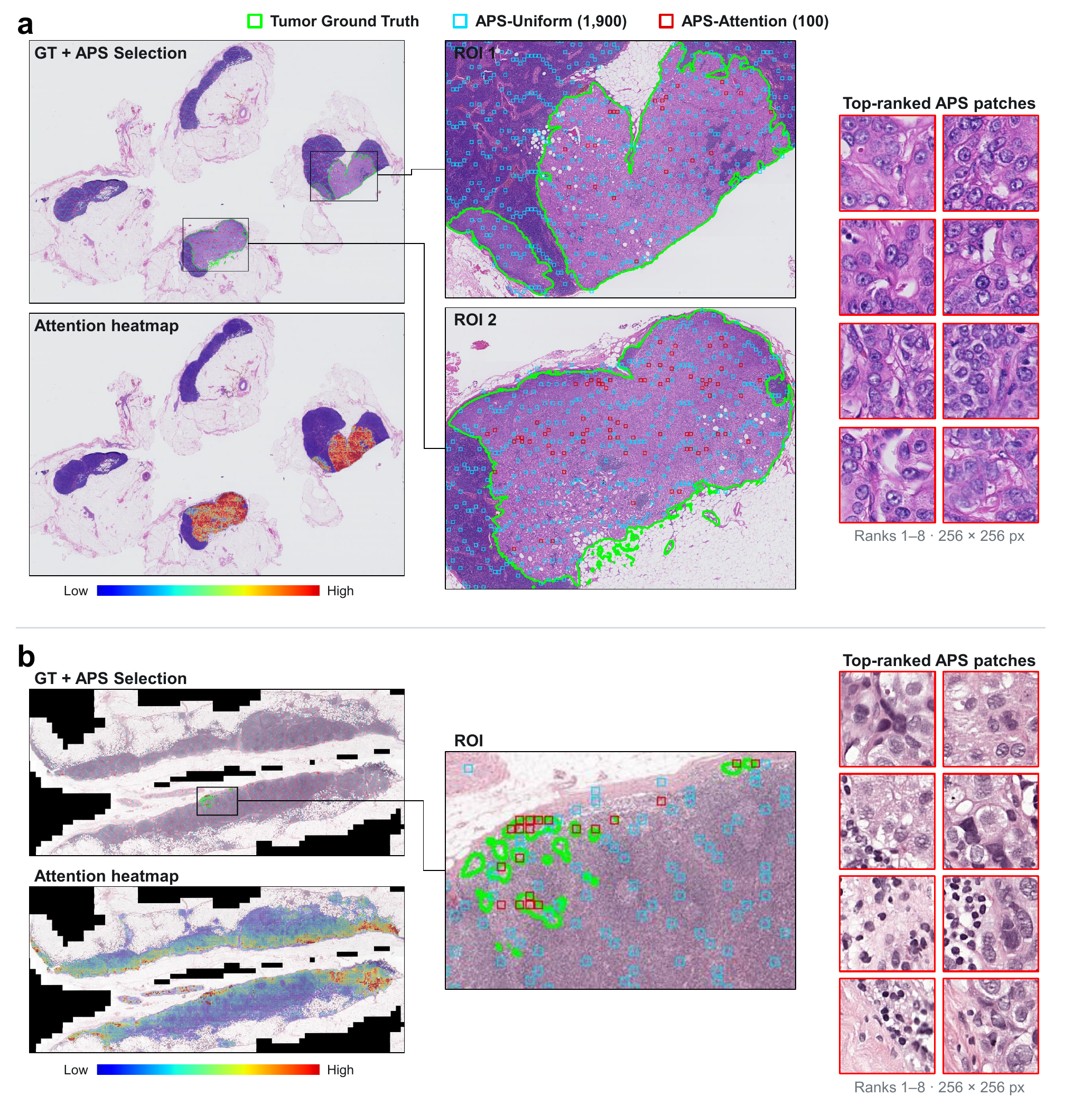}
\caption{\textbf{Qualitative assessment of APS-selected regions on the CAMELYON dataset.} 
\textbf{a}, Representative case. 
\textbf{b}, Challenging case with small tumor regions. 
In each case, the original WSI is shown with ground truth (GT) annotations and an overview of APS-selected patches, alongside the ABMIL attention heatmap computed from LiteFM features, highlighted regions of interest (ROIs), and the top-ranked patches selected by APS. Green, blue and red boxes denote GT regions, APS-uniform samples and APS-attention samples, respectively. The APS configuration for CAMELYON is $k_u=1,900$ and $k_a=100$ (Extended Data Table~\ref{tab:cohort_summary}).}
\label{fig:interpretability}
\end{figure}

\paragraph{The attention branch complements uniform sampling by recovering diagnostically informative regions.}
Uniform sampling alone was sufficient for many tasks, consistent with the widespread redundancy of WSIs. However, among the 14 task-cohort settings in which the attention branch was activated, APS significantly outperformed patch-budget-matched uniform selection, using the same total number of selected patches, in 11 settings (paired one-sided Wilcoxon signed-rank test; Fig.~\ref{fig:ablation}c). These results show that attention-guided selection adds measurable value in most attention-enabled settings, particularly when broad spatial coverage alone does not adequately capture task-relevant regions. We next examined the spatial distribution of APS-selected patches on two CAMELYON test slides, using this dataset because its region-level tumor annotations enabled direct comparison with diagnostically relevant regions (Fig.~\ref{fig:interpretability}). In the representative slide, attention-selected patches were concentrated within annotated tumor regions. In the challenging slide containing small tumor foci, uniform sampling sparsely covered the lesions and missed some annotated regions, whereas attention-selected patches preferentially overlapped these regions and the top-ranked patches contained tumor morphology. Together, the quantitative and qualitative analyses support the hybrid design of APS: uniform sampling provides broad and inexpensive coverage, while the attention branch helps recover focal diagnostic regions that may otherwise be undersampled or missed.

\paragraph{The selected LiteFM configuration balances accuracy and computational efficiency.} To evaluate student architecture and teacher composition, we compared LiteFM-S (ViT-Ti, 5.72M parameters), LiteFM (ViT-S, 22.06M), LiteFM-L (ViT-B, 86.59M), and LiteVirchow2 (ViT-S, 22.06M; distilled solely from Virchow2) (Extended Data Table~\ref{tab:litepath_family}; Fig.~\ref{fig:ablation}d). LiteFM achieved an average Macro-AUC of 0.8278 and C-index of 0.7213, compared with 0.8239 and 0.7122 for LiteFM-S, 0.8277 and 0.7114 for LiteFM-L, and 0.8270 and 0.7164 for LiteVirchow2, respectively. Thus, the ViT-S student with multi-teacher distillation maintained a favorable accuracy-size balance: increasing student size did not improve aggregate performance under the current pretraining setup, whereas the single-teacher variant performed slightly worse across classification and survival tasks. In the separate 23-system comparison that included full-patch LiteFM as an additional model, LiteFM achieved an average rank of 7.18, outperforming H0-mini (8.16) and GPFM (10.7) and remaining close to LitePath (6.98) and Virchow2 (7.09; Fig.~\ref{fig:ablation}e).

\subsection*{Reader study of LitePath-assisted diagnosis}

\begin{figure}[hp]
\centering
\includegraphics[width=\linewidth]{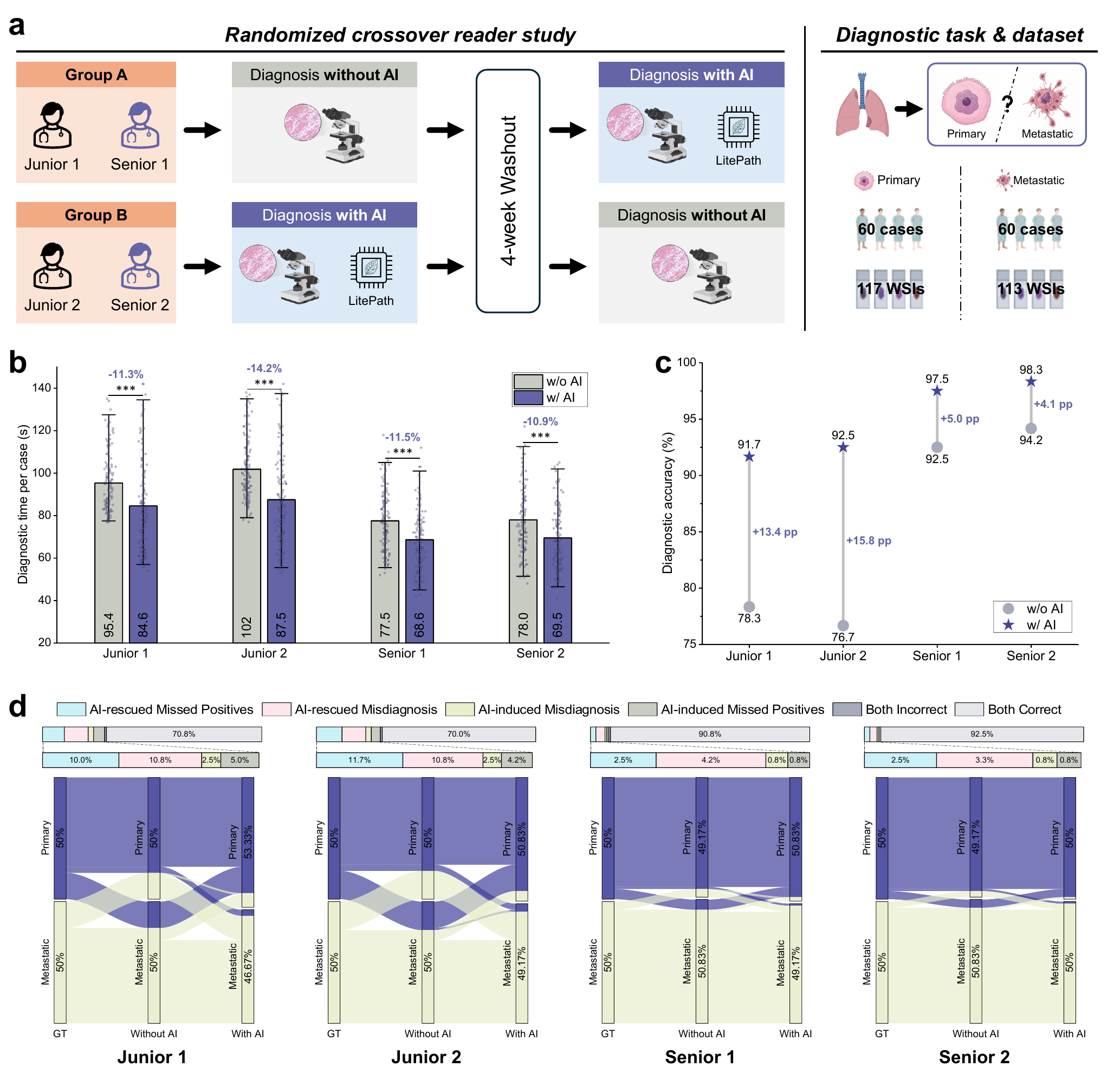}
\caption{\textbf{Randomized crossover reader study of LitePath-assisted diagnosis.}
\textbf{a}, Study design. Four pathologists (two junior and two senior) classified 120 cases as primary lung cancer or metastatic cancer under without-AI and with-AI conditions. Junior 1 and Senior 1 completed the without-AI condition first, followed by the with-AI condition after a 4-week washout period; Junior 2 and Senior 2 completed the conditions in the reverse order. The dataset comprised 60 primary cases (117 WSIs) and 60 metastatic cases (113 WSIs).
\textbf{b}, Case-level diagnostic time for each reader under the two conditions. Bars show the mean case-level diagnostic time, and error bars indicate the 2.5th--97.5th percentile interval for each reader and condition. Statistical significance was assessed separately for each reader using a paired one-sided Wilcoxon signed-rank test; *** $P<0.001$.
\textbf{c}, Diagnostic accuracy under the two conditions.
\textbf{d}, Composition of case-level diagnostic transitions. AI-rescued missed positives were metastatic cases classified as primary without AI and as metastatic with AI; AI-rescued misdiagnoses were primary cases classified as metastatic without AI and as primary with AI. AI-induced misdiagnoses were primary cases classified correctly without AI but as metastatic with AI; AI-induced missed positives were metastatic cases classified correctly without AI but as primary with AI. Both correct and both incorrect denote cases classified correctly or incorrectly, respectively, under both conditions. Alluvial plots show the case-level transitions from GT to without-AI and with-AI classifications.}
\label{fig:reader_study}
\end{figure}

Having established that LitePath preserved task performance while reducing computational cost, we next evaluated its use in an AI-assisted diagnostic workflow. In a randomized paired crossover reader study, four pathologists (two junior and two senior) classified the same 120 lung cancer cases as primary lung cancer or metastatic cancer under unaided and LitePath-assisted conditions. The dataset included 60 primary cases (117 WSIs) and 60 metastatic cases (113 WSIs). Junior 1 and Senior 1 first read the cases without AI and then with AI after a 4-week washout period, whereas Junior 2 and Senior 2 followed the reverse sequence (Fig.~\ref{fig:reader_study}a). Diagnostic accuracy and case-level diagnostic time were recorded for every reader and case.

As a result, mean diagnostic time decreased from 95.4s to 84.6s for Junior 1 (11.3\%), from 102s to 87.5s for Junior 2 (14.2\%), from 77.5s to 68.6s for Senior 1 (11.5\%), and from 78.0s to 69.5s for Senior 2 (10.9\%; Fig.~\ref{fig:reader_study}b). LitePath required approximately 8.15 s per case for inference ($k_u=2,000$, $k_a=0$) in this task on a Jetson device, substantially less than the time required for reader review. This low latency allowed LitePath to complete inference for the next case while the reader was reviewing the current case, ensuring that predictions were available without additional waiting and providing a workflow-level efficiency advantage over slower existing PFMs.

Diagnostic accuracy increased from 78.3\% (94/120) to 91.7\% (110/120) for Junior 1, from 76.7\% (92/120) to 92.5\% (111/120) for Junior 2, from 92.5\% (111/120) to 97.5\% (117/120) for Senior 1, and from 94.2\% (113/120) to 98.3\% (118/120) for Senior 2, corresponding to increases of 13.4, 15.8, 5.0 and 4.1 percentage points, respectively (Fig.~\ref{fig:reader_study}c). More specifically, LitePath assistance corrected 25, 27, 8 and 7 initially incorrect classifications for Junior 1, Junior 2, Senior 1 and Senior 2, respectively, while introducing 9, 8, 2 and 2 changes from correct to incorrect classifications. The numbers of cases correctly classified with and without AI by the four readers were 85, 84, 109 and 111, respectively. Conversely, only 1, 1, 1 and 0 cases were incorrectly classified under both conditions (with and without AI) (Fig.~\ref{fig:reader_study}d). These results demonstrate that LitePath-assisted workflows can enhance diagnostic efficiency and accuracy across readers with different levels of experience.

\section*{Discussion}
LitePath shows that jointly reducing encoder size and the number of fully processed patches can substantially lower recurring PFM inference cost while preserving competitive performance across the evaluated downstream tasks. LiteFM and APS address these two sources of cost through multi-teacher distillation and adaptive patch selection, respectively. Across 45 cohorts and 33 tasks, this combination maintained competitive classification and survival performance while enabling inference on low-power edge hardware. These results position computational efficiency as a system-level design objective for PFMs and yield two complementary insights into how WSI inference can be made more resource efficient.

The first insight concerns how WSIs are sampled. The APS results indicate that exhaustive patch encoding is unnecessary for many slide-level tasks, but also show that reducing patch count should not be equated with uniform subsampling alone. Uniform sampling provided sufficient broad tissue coverage for many tasks, whereas the attention branch improved performance in 11 of 14 attention-enabled settings and recovered focal tumor regions that were sparsely covered or missed by uniform sampling. The two branches therefore serve complementary functions: uniform sampling establishes efficient whole-slide coverage, while attention guidance adds sensitivity to localized diagnostic evidence. This complementarity suggests that hybrid selection can reduce WSI computation while retaining sensitivity to focal diagnostic regions.

The second insight concerns how PFM representations are compressed. Using direct $\ell_1$ regression of the final teacher features, the 22.06M-parameter full-patch LiteFM achieved a stronger aggregate rank than H0-mini and GPFM, despite their larger encoders and more elaborate distillation objectives. This comparison demonstrates the empirical sufficiency of directly matching the final teacher representations under the evaluated conditions, rather than the universal superiority of $\ell_1$ regression. PFM performance nevertheless remained task-dependent: LitePath performed strongly overall but did not lead the lung cancer tasks. Differences in pretraining corpora and downstream tasks across PFMs prevent reliable attribution of this pattern to model size or pretraining scale. This task dependence underscores the importance of evaluating compact PFMs across diverse tasks and institutions rather than inferring general performance from any single organ or endpoint.

Together, these efficiencies translate into a practical centralized-development and distributed-inference model. For the evaluated tasks, the frozen LiteFM encoder, task-specific ABMIL and APS weights, and validation-selected patch configuration can be distributed as a complete inference package. An end-site hospital therefore needs only to preprocess incoming WSIs and execute the packaged pipeline, without on-site model training, hyperparameter tuning or configuration selection. The feasibility of this deployment model was supported by applying the same frozen pipelines, without further adaptation, to 10 external cohorts and two prospectively collected cohorts, demonstrating cross-site transfer and task-specific performance under chronological accrual. Beyond technical portability, the controlled four-reader primary-versus-metastatic study showed that LitePath assistance improved diagnostic accuracy and reduced diagnostic time for readers with different levels of experience, providing pathologist-facing evidence that its computational efficiency can translate into practical workflow benefits. This deployment model may be particularly valuable to regional and medium-sized hospitals with limited access to high-performance computing infrastructure.

Reducing model-side computation also changes the next systems-level optimization target. As feature-extraction cost decreases, WSI reading, patch retrieval, data transfer and caching may account for a larger fraction of end-to-end latency. Further gains will therefore depend on data-loading and preprocessing pipelines alongside model efficiency. Improving these system components could further increase the throughput achievable by LitePath in routine use.

More broadly, these findings establish LitePath as a resource-efficient framework for slide-level classification and survival analysis. Future work could extend adaptive patch selection to other computational pathology tasks, including segmentation, dense prediction and quantitative estimation, with selection strategies tailored to their distinct spatial requirements. By maintaining competitive PFM performance while substantially reducing recurring inference cost, LitePath improves the practical deployability of WSI analysis in computationally constrained settings.

\section*{Methods}
\subsection*{Inference Pipeline of LitePath}
In conventional PFM-based frameworks, WSIs are partitioned into numerous non-overlapping pathology patches. A PFM processes these patches to produce patch-level embeddings, and a multiple-instance learning (MIL) model (e.g., ABMIL~\cite{abmil}) aggregates those embeddings into a slide-level prediction. This paradigm, however, ignores the substantial redundancy among WSI patches and is therefore inefficient in practice. To enable efficient and reliable inference, we propose an alternative inference pipeline in LitePath (Fig.~\ref{fig:main}a). LitePath consists of a compact PFM, LiteFM, built on the ViT-S architecture~\cite{vit,deit}, and a lightweight, plug-and-play module, APS, which incorporates uniform-based and attention-based strategies for patch selection. Rather than computing full embeddings for every patch, we first obtain shallow features by running each patch only through the Patch Embedding layer and the first transformer block of LiteFM. APS then adaptively selects a small set of representative patches using their indices and these shallow features. Only the hidden features of the selected patches are forwarded through the remaining LiteFM layers to produce the final prediction.

Specifically, given a slide consisting of $N$ patches $\{\mathbf{x}_i\}_{i=1}^N$, each patch is processed by the Patch Embedding layer and the first attention block, resulting in shallow hidden features $\{\mathbf{H}_i\}_{i=1}^N$. To efficiently identify representative patches, APS employs uniform-based and attention-based selection strategies. In the uniform-based strategy, $k_u$ patch indices are selected via uniform sampling, yielding the set 
$\mathcal{U} = \left\{ i_m \mid i_m = \left\lfloor \frac{(m-1)N}{k_u} \right\rfloor + 1,\ m=1,\ldots, k_u \right\}$.
In the attention-based strategy, the hidden features are fed into a lightweight scoring network to estimate attention scores $\mathbf{\hat{A}}_i$. The top $k_a$ patches with the highest scores (excluding those already selected by $\mathcal{U}$) are chosen, forming the set 
$\mathcal{A} = \left\{ i_n \mid i_n \in \arg\max_{i \notin \mathcal{U}}^{(k_a)} \mathbf{\hat{A}}_{i} \right\}$.
The final set of selected patch indices is given by $\mathcal{S} = \mathcal{U} \cup \mathcal{A}$. The corresponding features of these selected patches, $\{\mathbf{F}_s\}_{s \in \mathcal{S}}$, are subsequently extracted and fed into the MIL model for downstream task prediction. A comparison of this process with the conventional PFM pipeline is outlined in Extended Data Algorithms \ref{alg:litepath*} and \ref{alg:PFM}. The selection numbers $k_u$ and $k_a$ are determined through a grid search on the held-out validation set for each task (Extended Data Table~\ref{tab:cohort_summary}).

\subsection*{Model Training}
To enable the efficient inference pipeline of LitePath, three components require training: the feature extractor LiteFM, the APS scoring network, and the ABMIL. We adopt a three-stage training scheme,  dedicating one stage to each component. In the first stage, LiteFM is pretrained through knowledge distillation from three large, SOTA PFMs using a curated database. In the second stage, patch features extracted by LiteFM are used to train an ABMIL model for the target downstream task. In the third stage, with the LiteFM and ABMIL fixed, the APS scoring network is trained to predict ABMIL’s final attention scores from LiteFM’s shallow features, enabling identification of important patches using partial inference. Note that both the ABMIL and the APS scoring network are task-specific. This staged procedure ensures each component is optimized for the proposed efficient inference pipeline.

\subsubsection*{Stage 1 - Distillation Pretraining}
Several benchmark studies have demonstrated that no single PFM consistently outperforms all others across downstream tasks~\cite{gpfm,PathBench}. This observation motivates the integration of multiple powerful models into a single generalizable framework via knowledge distillation. To enhance the performance of our lightweight PFM, we select three expert models--Virchow2, H-Optimus-1, and UNI2--that exhibit competitive results across diverse tasks, and distill their knowledge into LiteFM, as illustrated in Fig.~\ref{fig:main}b. For distillation pretraining, we followed GPFM~\cite{gpfm} to curate 33 public datasets. We used the CLAM toolkit~\cite{CLAM} for tissue segmentation and extraction of non-overlapping $512 \times 512$ patches at level 0, while preserving the original WSI resolution to retain scale diversity. This yielded 190,212,668 patches from 72,280 WSIs (Extended Data Table~\ref{tab:dataset_distributation}). Using this database, knowledge distillation is performed by enforcing $\ell_1$ consistency between the embeddings extracted by LiteFM and those from the teacher models. Specifically, given an input image, we obtain the embeddings of the three teachers, $\mathbf{F}^{[\text{Virchow2}]}$, $\mathbf{F}^{[\text{H-Optimus-1}]}$, and $\mathbf{F}^{[\text{UNI2}]}$, as well as the student embedding, $\mathbf{F}^{[\text{LiteFM}]}$. Three projection heads, $\varphi_1$, $\varphi_2$, and $\varphi_3$, are adopted to project the student embedding to the same dimensionality as each teacher, respectively. The optimization objective is to minimize the loss function $L_{KD} = \alpha \cdot \ell_1\big(\varphi_1(\mathbf{F}^{[\text{LiteFM}]}), \mathbf{F}^{[\text{Virchow2}]}\big) 
+ \beta \cdot \ell_1\big(\varphi_2(\mathbf{F}^{[\text{LiteFM}]}), \mathbf{F}^{[\text{H-Optimus-1}]}\big)
+ \gamma \cdot \ell_1\big(\varphi_3(\mathbf{F}^{[\text{LiteFM}]}), \mathbf{F}^{[\text{UNI2}]}\big)$, where $\alpha, \beta, \gamma$ are the weights assigned to each of the three teacher models. The teacher-specific projection heads accommodate differences in output dimensionality and representation space, allowing the same direct $\ell_1$ regression objective to align LiteFM with the three final teacher embeddings. The distillation target is the final patch representation used by downstream models. By directly matching these outputs, LiteFM is optimized to inherit the information encoded in the teacher representations without separately reconstructing each teacher's internal tokens, attention maps or pretraining objective. Details of distillation pretraining are presented in Extended Data Table~\ref{tab:pretraining}.

\subsubsection*{Stage 2 - ABMIL Training}
In CPath, multiple instance learning (MIL) serves as the standard framework for downstream task analysis using PFM embeddings. Although several MIL methods are available~\cite{abmil,shao2021transmil,zhang2022dtfd,lin2023interventional,zhang2024attention}, ABMIL\cite{abmil} stands out for its simplicity and consistently competitive performance with PFM embeddings\cite{uni,virchow,gpfm}. Consequently, following previous studies, we adopt ABMIL for PFM performance evaluation. Given the patch features $\mathbf{F} \in \mathbb{R}^{N\times D}$ of a slide, ABMIL first projects the feature dimension to 512, resulting in $\mathbf{F}' \in \mathbb{R}^{N\times 512}$. Subsequently, two fully connected layers map $\mathbf{F}'$ to $\mathbf{A} \in \mathbb{R}^N$, which represents the attention score for each patch feature. The attention scores $\mathbf{A}$ are then used to aggregate the projected features $\mathbf{F}'$, producing a slide-level representation $\mathbf{Z}$. Finally, a projection head generates the final prediction based on $\mathbf{Z}$. We followed the configurations used in PathBench~\cite{PathBench}: for classification, ABMIL was optimized using cross-entropy loss; for survival analysis, event times were discretized into four quantile-based intervals defined by the event-time distribution among uncensored cases, and the model predicted the hazard for each interval. The model was optimized using the censored discrete-time negative log-likelihood (\texttt{NLLSurvLoss}, $\alpha=0$). Survival probabilities were obtained by cumulatively multiplying one minus the predicted hazards, and the risk score was defined as the negative sum of the survival probabilities across the four intervals. The architecture and training details of ABMIL are presented in Extended Data Table~\ref{tab:abmil}.

\subsubsection*{Stage 3 - Score Matching}
The well-trained LiteFM and ABMIL together support the conventional inference pipeline of PFM for pathology analysis, which incorporates all patches in the computation. To enable our APS to identify representative patches with minimal computational cost, we propose training a scoring network $\phi$ to estimate the final attention scores using shallow features, thereby allowing patch selection to be performed via partial inference. Specifically, given a slide consisting of $N$ patches $\{\mathbf{x}_i\}_{i=1}^N$, we compute the hidden features after the first block $\{\mathbf{H}_i\}_{i=1}^N$ and the corresponding attention scores $\{\mathbf{A}_i\}_{i=1}^N$. 
Then we process the hidden features by $\mathbf{H}_i^{\text{[Concat]}} = \text{Concat}\left[\mathbf{H}_i^{\text{[CLS]}}; \text{Mean}(\mathbf{H}_i^{[\text{Patch}]})\right]$.
The training objective is to enforce consistency between the estimated scores $\mathbf{\hat{A}}_i = \phi(\mathbf{H}_i^{\text{[Concat]}}), i=1,...N,$ and the true attention scores $\{\mathbf{A}_i\}_{i=1}^N$. We measure this consistency using a soft cross-entropy loss 
$L_{\text{score}} = -\sum_{i=1}^N p_i \log \hat{p}_i$,
where $p_i = \text{softmax}\left(\frac{\mathbf{A}}{\tau}\right)_i$ and $\hat{p}_i = \text{softmax}\left(\frac{\mathbf{\hat{A}}}{\tau}\right)_i$, with $\tau$ denoting the temperature parameter. Details of the APS scoring network are presented in Extended Data Table~\ref{tab:aps}.

\subsection*{Multi-Center Clinical Evaluation Tasks}
To comprehensively evaluate the accuracy of LitePath, we performed experiments across a diverse array of cancers involving four organs, encompassing 33 distinct tasks and 45 cohorts (33 internal, 10 external, and 2 prospective cohorts). The benchmark includes 26 classification tasks and seven survival-analysis tasks, summarized in Extended Data Tables~\ref{tab:task_summary} and~\ref{tab:survival_task_summary}, respectively. Task-specific APS configurations for all clinical cohorts are provided in Extended Data Table~\ref{tab:cohort_summary}. The dataset includes 17,837 slides from 9,977 patients across nine hospitals, designated H1 through H9. Notably, the prospective dataset contains 1,897 slides from 525 patients at H1, collected between April and August 2025. Some WSIs were included in multiple downstream tasks within the same organ. Accordingly, task-specific slide counts are not additive; the dataset-level totals reported here represent unique WSIs after accounting for cross-task overlap.

The multi-center benchmark was derived from an earlier implementation of PathBench but was not intended to exhaustively reproduce every task in the current benchmark. We selected four clinically important organ systems with sufficient task diversity, prioritized a central subtyping task within each organ, and included representative diagnostic, staging, molecular, immunohistochemical and prognostic endpoints. Brain cancer was not included because the PathBench contained only three eligible tasks, which was insufficient for a comparable organ-level assessment. We additionally included two prospectively collected cohorts beyond the PathBench version used in this study to strengthen evaluation under temporal distribution shift.

\subsubsection*{Lung Cancer} 
\phantomsection
\label{sec:lung_cancer}
\paragraph{Primary and Metastatic Cancer Classification.}
Accurate distinction between primary and metastatic lung tumors is a key diagnostic challenge in precision oncology, as it directly informs patient management and prognosis. To evaluate the performance of PFMs in clinically relevant scenarios, we built an internal cohort that contains 389 primary cases (686 slides) and 457 metastatic cases (736 slides) from Hospital H1. We built two external cohorts: The Hospital H5 cohort contains 237 primary cases (237 slides) and 256 metastatic cases (256 slides); The Hospital H6 cohort contains 465 primary cases (744 slides) and 361 metastatic cases (678 slides).

\paragraph{Cancer Subtyping.}
Precise subtyping of lung cancer represents a foundational aspect of pathological diagnosis. In this study, we performed classification among six lung cancer subtypes: Lung Neuroendocrine Tumor (LNET), Minimally Invasive Adenocarcinoma (MIA), Lung Adenosquamous Carcinoma (ASC), Invasive Adenocarcinoma (IAC), Lung Squamous Cell Carcinoma (LUSC), and Lung Adenocarcinoma In Situ (AIS). Specifically, we constructed an internal cohort from Hospital H1 that contains 131 LNET cases (170 slides), 121 MIA cases (167 slides), 19 ASC cases (34 slides), 150 IAC cases (181 slides), 123 LUSC cases (255 slides), and 150 AIS cases (271 slides). Additionally, we constructed a prospective cohort from Hospital H1 that contains 5 LNET cases (23 slides), 150 MIA cases (331 slides), 7 ASC cases (34 slides), 150 IAC cases (647 slides), 70 LUSC cases (400 slides), and 127 AIS cases (367 slides).

\paragraph{NSCLC Subtyping.}
Accurate discrimination between lung adenocarcinoma (LUAD) and lung squamous cell carcinoma (LUSC) among non-small cell lung cancer (NSCLC) subtypes is of paramount importance in pathology, given its direct impact on clinical management and tailored treatment strategies. For this task, we constructed an internal cohort from Hospital H1 that contains 300 LUAD cases (300 slides) and 300 LUSC cases (300 slides). 

\paragraph{Lymph Node Metastasis Prediction.}
Reliable prediction of lymph node metastasis is critical in the pathological evaluation of lung cancer, informing disease staging and guiding surgical and adjuvant treatment decisions. For this task, we constructed an internal cohort from Hospital H1 that contains 71 lymph node metastasis-positive cases (231 slides) and 250 lymph node metastasis-negative cases (828 slides).

\paragraph{IHC Status Prediction.}
Assessment of immunohistochemistry (IHC) status plays an essential role in lung cancer pathology, facilitating biomarker-driven stratification and the implementation of targeted therapeutic interventions. In this task, we focused on the prediction of P63. We constructed an internal cohort from Hospital H1 that contains 197 P63-negative cases (249 slides) and 90 P63-positive cases (105 slides). Additionally, we constructed a prospective validation cohort from Hospital H1 that contains 24 P63-negative cases (126 slides) and 34 P63-positive cases (171 slides).

\subsubsection*{Breast Cancer}
\paragraph{TNM-N Staging (N0/N+).}
Accurate staging of axillary lymph node involvement, defined by the TNM-N staging, is fundamental for prognostication and treatment planning in breast cancer. For this task, we constructed an internal cohort from Hospital H2 that contains 343 N0 cases (916 slides) and 125 N+ cases (381 slides).

\paragraph{pTNM Overall Staging (I/II/III).}
The pTNM overall stage, synthesized from postoperative pathological examination of the primary tumor and regional lymph nodes, represents the definitive prognostic benchmark and is the principal determinant for adjuvant therapy recommendations in breast cancer. For this task, we constructed an internal cohort from Hospital H2 that contains 192 stage I cases (451 slides), 232 stage II cases (727 slides), and 43 stage III cases (116 slides).

\paragraph{Molecular Subtyping.}
Identification of molecular subtypes is critical for personalized cancer therapy, as it enables the selection of targeted treatments and improves prognostic accuracy. We constructed an internal cohort from Hospital H2 that contains 307 Luminal A cases (310 slides), 614 Luminal B1 cases (618 slides), 243 Luminal B2 cases (268 slides), 589 TNBC cases (1,932 slides), and 292 HER-2 cases (323 slides). We built an external cohort from H9 that includes 102 Luminal A cases, 89 Luminal B1 cases, 24 Luminal B2 cases, 101 TNBC cases, and 102 HER-2 cases, where each case contains only one slide.

\paragraph{IHC Status Prediction.}
Assessing immunohistochemical marker status directly impacts clinical management and patient prognosis. To evaluate the performance of foundation models on biomarker prediction, we constructed cohorts from Hospital H2 to perform the prediction of 5 biomarkers: AR, ER, PR, HER2, and CK5, respectively. For the prediction of AR, the cohort contains 463 AR negative cases (731 slides) and 677 AR positive cases (841 slides). For ER, it contains 767 ER negative cases (1,264 slides) and 781 ER positive cases (786 slides). For PR, it contains 623 PR negative cases (1,108 slides) and 933 PR positive cases (950 slides). For HER2, it contains 511 HER2 negative cases (743 slides) and 833 HER2 positive cases (975 slides). For CK5, it contains 753 CK5 negative cases (859 slides) and 208 CK5 positive cases (379 slides).

\paragraph{Disease-Free Survival Analysis.}
Prediction of disease-free survival (DFS) is important for postoperative monitoring and recurrence risk assessment in breast cancer. For this task, we constructed a cohort from Hospital H2 that contains 380 disease-free patients (1,066 slides) and 71 recurred patients (204 slides). For external validation, we built a cohort from H9 that contains 76 disease-free patients (76 slides) and 3 recurred patients (3 slides).

\paragraph{Overall Survival Analysis.}
Accurate prediction of overall survival (OS) is essential for prognostic stratification and long-term treatment planning in breast cancer. For this task, we constructed a cohort from Hospital H2 that contains 392 censored patients (1,089 slides) and 59 deceased patients (181 slides).

\subsubsection*{Gastric Cancer}  
\phantomsection
\label{sec:gastric_cancer}
\paragraph{Cancer Grading.}
Histological grading in gastric cancer provides vital insights into tumor differentiation and is a key factor in prognosis and treatment selection. For this task, we constructed an internal cohort from H1 that contains 81 well/moderately differentiated (G1+G2) cases (82 slides) and 318 poorly differentiated (G3) cases (319 slides). For external validation, we constructed two cohorts. The Hospital H3 cohort contains 55 G1+G2 cases and 190 G3 cases, while the Hospital H4 cohort provides 62 G1+G2 cases and 258 G3 cases, with each case containing one slide.

\paragraph{Pathological Subtyping.}
Classifying gastric tumors by pathological subtype enables tailored therapeutic approaches and supports more precise outcome prediction. We constructed an internal cohort from Hospital H1 that contains 163 Signet Ring Cell Carcinoma (SRCC) cases (163 slides), 166 Tubular Adenocarcinoma (TAC) cases (167 slides), and 66 non-specified Stomach Adenocarcinoma (NOS) cases (67 slides). We built an external cohort from H3 that includes 59 SRCC cases and 195 NOS cases, with each case containing only one slide.

\paragraph{TNM-N Staging (N0/N+).}
Assessment of lymph node involvement (N stage) informs both disease staging and the likelihood of recurrence, directly impacting surgical and adjuvant therapy planning. We constructed an internal cohort from Hospital H1 that contains 186 N0 cases (188 slides) and 212 N+ cases (212 slides). We built an external cohort from H3 that contains 85 N0 cases and 175 N+ cases, with each case containing only one slide. 

\paragraph{Perineural Invasion Detection.}
Detection of perineural invasion serves as an indicator of aggressive disease behavior and is associated with poorer prognosis in gastric cancer. For this task, we built an internal cohort from H1 that contains 255 PNI-positive cases (256 slides) and 141 PNI-negative cases (142 slides). We built an external cohort from H3 that contains 156 PNI-positive cases and 76 PNI-negative cases, with each case containing only one slide.

\paragraph{Vascular Invasion Detection.}
Identifying vascular invasion is crucial for estimating the risk of metastasis and guiding postoperative management strategies. We built an internal cohort from Hospital H1 that contains 197 VI-positive cases
(198 slides) and 198 VI-negative cases (199 slides). We built an external cohort from H3 that contains 140 VI-positive and 90 VI-negative cases, with each case containing only one slide. 

\paragraph{Normal/Abnormal Classification.}   
Differentiating between normal and abnormal gastric tissue aids in early detection and prevention of malignant transformation. For this task, we constructed an internal cohort from H7 that contains 733 normal slides and 1,967 abnormal slides.

\paragraph{Intestinal Metaplasia Classification.} Recognizing intestinal metaplasia (IM) is important for surveillance and risk assessment, as it represents a precancerous change in gastric mucosa. We constructed an internal cohort from H7 comprising 270 IM slides and 2,430 non-IM slides. 

\paragraph{IHC Status Prediction of HER2 and S-100.}
Predicting immunohistochemical marker status supports targeted therapy selection and refines diagnostic accuracy. We focused on HER2 and S-100, which are especially crucial for gastric cancer. To evaluate the prediction of HER2, we constructed an internal cohort from H1, H3, and H4 that contains 549 low-HER2 cases (IHC 0/1+) and 126 high-HER2 cases (IHC 2+/3+). For S-100, we constructed an internal cohort from H1, H3, and H4 that contains 90 IHC 0 slides and 270 IHC 1+ slides. 

\paragraph{Disease-Free Survival Analysis.}
Prediction of disease-free survival (DFS) is crucial for postoperative surveillance and adjuvant therapy planning in gastric cancer. For this task, we assembled a cohort from Hospitals H3 and H4 comprising 580 patients, including 448 disease-free cases (448 slides) and 132 cases with recurrence (132 slides). Specifically, the H3 cohort included 157 disease-free patients (157 slides) and 103 patients with recurrence (103 slides), whereas the H4 cohort included 291 disease-free patients (291 slides) and 29 patients with recurrence (29 slides).

\paragraph{Overall Survival Analysis.}
Accurate prediction of overall survival (OS) is essential for guiding treatment decisions and prognostic stratification in gastric cancer. For this task, we assembled a cohort from Hospitals H3 and H4 comprising 580 patients, including 374 censored cases (374 slides) and 206 uncensored cases (206 slides). Specifically, the H3 cohort included 172 censored cases (172 slides) and 88 uncensored cases (88 slides), whereas the H4 cohort included 202 censored cases (202 slides) and 118 uncensored cases (118 slides).

\subsubsection*{Colorectal Cancer}
\paragraph{TNM-N Staging (N0/N+)}
Accurate evaluation of lymph node involvement is crucial for determining disease progression and guiding postoperative treatment in colorectal cancer. We constructed an internal cohort from Hospital H8 that contains 367 N0 cases (1,848 slides) and 230 N+ cases (871 slides).

\paragraph{TNM-T Staging (T1+T2/T3+T4)}
Assessing the depth of tumor invasion helps stratify patients into risk groups and informs decisions regarding surgical and adjuvant therapies. For this task, we constructed an internal cohort from H8 that contains 76 T1+T2 cases (319 slides) and 519 T3+T4 cases (2,391 slides).

\paragraph{TNM-T Staging (T1/T2/T3/T4)}
To further elucidate the TNM-T staging, we expanded the task to a more fine-grained four-class classification. For this purpose, the internal cohort from H8 includes 20 T1 cases (75 slides), 56 T2 cases (244 slides), 440 T3 cases (2,130 slides), and 79 T4 cases (261 slides).

\paragraph{Consensus Molecular Subtyping}
Identifying consensus molecular subtypes enables a more refined understanding of tumor biology, supporting individualized management and prognostication. We constructed an internal cohort that contains 76 CMS1 (372 slides), 239 CMS2 (1,061 slides), 86 CMS3 (393 slides), and 187 CMS4 (857 slides).

\paragraph{Disease-Free Survival Analysis.}
Accurate prediction of disease-free survival (DFS) is crucial for postoperative surveillance and adjuvant therapy planning in colorectal cancer. For this task, we constructed a cohort from Hospital H8 comprising 608 cases (2,779 slides), including 389 disease-free patients (1,875 slides) and 219 recurred or progressed patients (904 slides).

\paragraph{Overall Survival Analysis.}
Accurate prediction of overall survival (OS) is essential for risk stratification and long-term outcome assessment in colorectal cancer. For this task, we constructed a cohort from Hospital H8 that contains 608 patients (2,779 slides), including 440 living patients (2,081 slides) and 168 deceased patients (698 slides).

\paragraph{Disease-Specific Survival Analysis.}
Disease-specific survival (DSS) captures tumor-related mortality and provides complementary prognostic information in colorectal cancer. For this task, we constructed a cohort from Hospital H1 that contains 294 patients (301 slides), including 252 living, or dead but tumor-free, patients (259 slides) and 42 dead patients with tumor (42 slides).

\subsection*{Open-Access Evaluation Tasks}

To complement the multi-center clinical evaluation and enhance reproducibility, we additionally evaluated PFMs on open-access pathology tasks. Given the potential overlap between these datasets and PFM pretraining corpora, these tasks were analyzed separately from the multi-center clinical evaluation and excluded from the aggregate ranking. The task-specific dataset composition and evaluation protocols are described below.

\subsubsection*{Slide-level Tasks}
\paragraph{BRACS for Breast Carcinoma Subtyping (3 classes \& 7 classes).}
Breast carcinoma subtyping requires the recognition of both broad diagnostic categories and fine-grained epithelial abnormalities. We used the BRACS dataset~\cite{Bracs}, which contains 545 H\&E WSIs from 187 patients after quality control. We evaluated both a three-class task, including 263 benign, 89 atypical, and 193 malignant slides, and a seven-class task, including 43 normal, 147 pathological benign, 73 usual ductal hyperplasia, 41 flat epithelial atypia, 48 atypical ductal hyperplasia, 61 ductal carcinoma in situ, and 132 invasive carcinoma. The dataset is stratified into training, validation, and test sets in a 7:1:2 ratio for both tasks, respectively.

\paragraph{CAMELYON for Breast Metastasis Detection (2 classes).}
Detection of breast cancer metastasis in lymph node WSIs is important for tumor staging and treatment planning. We used the CAMELYON cohort assembled from CAMELYON16~\cite{CAMELYON16} and CAMELYON17~\cite{CAMELYON17}. After removing one corrupted slide, the dataset contains 557 normal slides and 341 metastasis slides. The training, validation, and testing sets include 629, 90, and 179 slides, respectively. The pixel-level tumor annotations provided with CAMELYON were used for the qualitative APS analysis described below.

\paragraph{TCGA NSCLC Subtyping (2 classes).}
Accurate discrimination between lung adenocarcinoma (LUAD) and lung squamous cell carcinoma (LUSC) is fundamental for NSCLC diagnosis. We used diagnostic H\&E WSIs from the TCGA-NSCLC cohort~\cite{TCGA}, including 541 LUAD slides and 512 LUSC slides after tissue segmentation. The training, validation, and testing sets include 664, 100, and 289 slides, respectively.

\paragraph{TCGA-LUAD for TP53 and EGFR Gene Mutation Prediction (2 classes).}
Prediction of TP53 and EGFR mutation status from H\&E-stained LUAD WSIs is relevant to molecular stratification and targeted therapy. We used 469 TCGA-LUAD~\cite{TCGA} slides for these two tasks. For TP53, the dataset includes 248 mutant and 221 wild-type slides. For EGFR, it includes 71 mutant and 398 wild-type slides. The dataset is stratified into training, validation, and test sets in a ratio close to 7:1:2 for both tasks, respectively.

\paragraph{TCGA-CRC Molecular Subtyping (4 classes).}
Consensus molecular subtyping (CMS) of colorectal cancer captures clinically relevant differences in tumor biology. We used TCGA colon adenocarcinoma (COAD) and rectum adenocarcinoma (READ) cohorts~\cite{TCGA}, comprising 74 CMS1 slides, 211 CMS2 slides, 68 CMS3 slides, and 139 CMS4 slides. The training, validation, and testing sets include 342, 47, and 103 slides, respectively.

\paragraph{TCGA Survival Analysis.}
We additionally evaluated slide-level survival prediction on TCGA colorectal cancer (TCGA-CRC), head and neck squamous cell carcinoma (TCGA-HNSC), lung adenocarcinoma (TCGA-LUAD), and skin cutaneous melanoma (TCGA-SKCM). Task-specific APS configurations are provided in Extended Data Table~\ref{tab:cohort_summary}, and the corresponding results are reported in Extended Data Table~\ref{tab:public_survival}.

\subsubsection*{Patch-level Tasks}
Patch-level ROI classification tasks provide a direct assessment of patch representation quality. To evaluate PFMs on these tasks, we trained an independent linear probe for each model, thereby addressing the concern that slide-level performance may mask differences in embedding quality across PFMs. We included five patch-level tasks in this study: CRC-MSI\cite{kather2019deep}, PCAM\cite{veeling2018rotation}, PanCancer-TIL\cite{saltz2018spatial}, UniToPatho\cite{barbano2021unitopatho}, and WSSS4LUAD\cite{han2022wsss4luad,han2022multi}.

\paragraph{CRC-MSI for MSI Screening\cite{kather2019deep}.}
Microsatellite instability (MSI) status is an important molecular marker in colorectal cancer, with implications for prognosis, hereditary cancer screening, and immunotherapy response. We used the CRC-MSI dataset for binary MSI screening from H\&E-stained colorectal cancer images. The dataset contains 51,918 ROI images of size $512 \times 512$ from TCGA, with patient-level MSI status grouped into MSI-high and non-MSI-high (MSI-low or microsatellite-stable) categories. The training and testing sets include 19,557 and 32,361 ROIs, respectively.

\paragraph{PCAM for Metastatic Tissue Classification\cite{veeling2018rotation,CAMELYON16}.}
Detection of metastatic tissue in lymph node histology is a canonical patch-level task for evaluating whether image representations capture diagnostically relevant tumor morphology. The PCAM dataset consists of 327,680 images of size $96 \times 96$ extracted from CAMELYON16, each annotated with a binary label indicating the presence or absence of metastatic tissue. The training, validation, and testing sets include 262,144, 32,768, and 32,768 ROIs, respectively.

\paragraph{TIL classification\cite{saltz2018spatial}.}
Tumor-infiltrating lymphocytes (TILs) reflect the tumor immune microenvironment and are associated with prognosis and treatment response across cancer types. We used the PanCancer-TIL dataset to evaluate binary TIL classification. The dataset contains 304,097 ROI images of size $100 \times 100$ pixels at 0.5 $\mu$m per pixel, labeled as TIL-positive when at least two TILs are present in the image and TIL-negative otherwise. The dataset includes 54,910 TIL-positive and 249,187 TIL-negative ROIs. The training, validation, and testing sets include 209,221, 38,601, and 56,275 ROIs, respectively.

\paragraph{UniToPatho for CRC Polyp Classification\cite{barbano2021unitopatho}.}
Colorectal polyp classification requires recognition of fine-grained glandular and dysplastic morphology. The UniToPatho dataset contains 9,536 H\&E-stained patches extracted from 292 WSIs and annotated into six colorectal polyp categories: normal tissue (950 ROIs), hyperplastic polyp (545 ROIs), tubular adenoma with high-grade dysplasia (454 ROIs), tubular adenoma with low-grade dysplasia (3,618 ROIs), tubulo-villous adenoma with high-grade dysplasia (916 ROIs), and tubulo-villous adenoma with low-grade dysplasia (2,186 ROIs). The training and testing sets include 6,270 and 2,399 ROIs, respectively.

\paragraph{WSSS4LUAD for Lung Adenocarcinoma Tissue Classification\cite{han2022wsss4luad,han2022multi}.}
Lung adenocarcinoma tissue classification evaluates whether patch representations distinguish tumor epithelium from tumor-associated stroma and normal tissue. The WSSS4LUAD dataset was collected from Guangdong Provincial People's Hospital and TCGA and contains 10,091 images annotated into three tissue categories: tumor epithelial tissue (6,579 ROIs), tumor-associated stroma (1,680 ROIs), and normal tissue (1,832 ROIs). As individual images may contain multiple tissue categories, a single label was assigned to each image according to diagnosability, prioritizing tumor epithelial tissue over tumor-associated stroma and normal tissue. A label-stratified 8:2 training and testing split was used, resulting in 8,072 training ROIs and 2,019 testing ROIs.

\subsection*{Downstream Evaluation Protocol}

For downstream WSI analyses, slide preprocessing followed the standardized PathBench pipeline~\cite{PathBench}. Only foreground tissue patches were analyzed, and all WSIs were processed at level 0 using $256 \times 256$-pixel patches for 20$\times$ WSIs and $512 \times 512$-pixel patches for 40$\times$ WSIs, ensuring consistent tissue coverage of 0.25 $\mu$m$^2$/pixel. Multi-center classification cohorts were stratified by label into fixed training, validation and test sets at a ratio of 7:1:2, whereas survival analyses used five-fold cross-validation. Public classification tasks used the dataset-specific splits described above.

All downstream model development was restricted to the corresponding training and validation data. For classification tasks, ABMIL models were trained on the training set, whereas model selection, early stopping where applicable, and selection of the task-specific APS configuration were performed on the validation set. For survival tasks, model fitting and model selection were performed independently within each cross-validation fold. The internal held-out test sets, external cohorts and prospective cohorts were reserved exclusively for final evaluation, without additional training or hyperparameter tuning. External survival cohorts, where available, were excluded from model fitting and selection and evaluated using the corresponding fold-specific models. External cohorts were independently collected from hospitals distinct from the training hospital, whereas prospective cohorts were defined by chronological accrual and evaluated as independent test sets.

The APS selection numbers $k_u$ and $k_a$ were determined by grid search on the validation set for each task. Because the candidate configurations differed only in the numbers of uniform and attention-selected patches, this search used the frozen LiteFM, ABMIL and APS models and required inference only.

\paragraph{Model-specific downstream implementations.} With the exception of TITAN~\cite{TITAN} and EAGLE~\cite{neidlinger2025deep}, the slide-level PFMs were evaluated using the common ABMIL architecture and task-specific objectives described in Stage 2. For classification tasks, TITAN and EAGLE were evaluated using their official downstream architectures and optimization settings on the benchmark's fixed training, validation and test splits. The TITAN linear probe fitted fixed slide embeddings using $\ell_2$-regularized logistic regression and the L-BFGS solver, with a maximum of 500 iterations. Following the default official code, the regularization coefficient was selected by minimizing validation log loss over 45 logarithmically spaced values between $10^{-5}$ and $10^{6}$. The EAGLE classifier was a multilayer perceptron (MLP) with a 256-dimensional hidden layer, SiLU activation and dropout of 0.5. It was trained with a batch size of 64 for up to 32 epochs using class-weighted cross-entropy, AdamW (learning rate, $1\times10^{-4}$; weight decay, $1\times10^{-2}$) and a one-cycle learning-rate policy. Checkpoints were selected by validation loss, with early stopping after eight epochs without improvement. At the time of evaluation, neither official repository provided a survival-analysis implementation. For TITAN, we therefore followed its published survival formulation by fitting an $\ell_2$-regularized linear Cox proportional-hazards model to the fixed slide embeddings using scikit-survival; the regularization coefficient was searched over 25 logarithmically spaced values between $10^{1}$ and $10^{5}$. For EAGLE, we retained the official MLP architecture and optimization settings, replaced the classification output with four interval-specific hazard logits, and replaced cross-entropy with the same \texttt{NLLSurvLoss} used for the survival ABMIL models.

\paragraph{Patch-level evaluation.} Data splits followed the GPFM evaluation suite~\cite{gpfm}. For each task, we extracted fixed ROI embeddings using each PFM and trained an independent linear probe on the corresponding training set. The linear probe was optimized with cross-entropy loss using the hyperparameters listed in Extended Data Table~\ref{tab:linear_probe}, and performance was evaluated on the held-out test set using Macro-AUC. Results are summarized in Extended Data Table~\ref{tab:roi_classification}.

\paragraph{APS analyses.} To isolate the contribution of attention guidance, the complete hybrid APS configuration was compared with patch-budget-matched uniform sampling using the same total number of selected patches. We also assessed whether APS preserved performance relative to full-patch LiteFM. For qualitative spatial assessment, we visualized two CAMELYON test slides, including a challenging slide containing small metastatic foci, and overlaid the ground-truth tumor annotations with patches selected by the uniform and attention branches. The visualization further included the ABMIL attention heatmap derived from LiteFM features and the eight highest-ranked attention-selected patches. For both cases, APS used $k_u=1,900$ and $k_a=100$ (Extended Data Table~\ref{tab:cohort_summary}).

\paragraph{Statistical analysis.}
Each slide-level classification experiment was repeated with ten random seeds, followed by non-parametric bootstrapping with 1,000 resamples per run; the resulting 10,000 bootstrap replicates were used to estimate 95\% confidence intervals. The five-fold survival evaluation was repeated with three random seeds, yielding 15 held-out-fold evaluations. Non-parametric bootstrapping with 1,000 resamples per held-out-fold evaluation yielded 15,000 bootstrap replicates for estimation of the 95\% confidence intervals. Each patch-level experiment was repeated with three random seeds and 1,000 bootstrap resamples per run, yielding 3,000 bootstrap replicates. For the comparison of APS with full-patch LiteFM, non-inferiority was defined using a prespecified margin of $-2.5\%$ and was established when the lower bound of the 95\% confidence interval for the performance difference exceeded this margin. APS and patch-budget-matched uniform sampling were compared using a paired one-sided Wilcoxon signed-rank test.

\subsection*{Reader Study Design}

We conducted a randomized paired crossover reader study of primary lung cancer versus metastatic cancer classification. The evaluation set contained 120 cases and 230 WSIs, comprising 60 primary cases (117 WSIs) and 60 metastatic cases (113 WSIs). The case-level primary-versus-metastatic label supplied with the evaluation set served as the ground truth (GT). All reader-study cases were strictly held out from model training, validation, checkpoint selection and APS configuration. The frozen LiteFM encoder, task-specific ABMIL and APS weights, and validation-selected patch configuration were used without retraining or hyperparameter tuning.

Four pathologists participated, including two junior (<5 years experience) and two senior readers (>5 years experience). The readers were randomized between two seniority-balanced, counterbalanced diagnostic sequences. Group A comprised Junior 1 and Senior 1, who completed unaided reading in the first period and LitePath-assisted reading in the second period. Group B comprised Junior 2 and Senior 2, who completed the assisted condition first and the unaided condition second. The two study periods were separated by a 4-week washout period.

In the unaided condition, readers reviewed the WSIs without model output. In the assisted condition, the same WSIs were presented together with the LitePath prediction. Case-level diagnostic time was measured from presentation of a case to submission of the reader's final primary-versus-metastatic classification. Because the LitePath inference time was much shorter than the typical reader-review interval, predictions for upcoming cases were generated in the background while the reader reviewed the current case, effectively concealing computational latency during the workflow. All 960 reader--case--condition observations were complete and included in the analysis.

AI effects were defined from GT and the paired unaided and assisted classifications. AI-rescued missed positives were metastatic cases classified as primary without AI and as metastatic with AI, whereas AI-rescued misdiagnoses were primary cases classified as metastatic without AI and as primary with AI. AI-induced misdiagnoses were primary cases classified correctly without AI but as metastatic with AI, whereas AI-induced missed positives were metastatic cases classified correctly without AI but as primary with AI. Both correct and both incorrect denoted cases classified correctly or incorrectly, respectively, under both conditions. Accuracy was summarized as counts and proportions, and the difference between conditions was reported in percentage points. Diagnostic time was summarized using the mean, with the 2.5th--97.5th percentile interval calculated from the case-level time distribution for each reader and condition. For each reader, diagnostic times under the two conditions were compared using a paired one-sided Wilcoxon signed-rank test, with the alternative hypothesis that diagnostic time was shorter with LitePath assistance.

\subsection*{Evaluation Metrics} 
\phantomsection
\label{sec:metrics}
Model accuracy on downstream tasks was evaluated using Macro-AUC for classification and C-index for survival analysis. Inference efficiency was assessed using FLOPs and throughput. For aggregate comparisons, PFMs were ranked within each cohort according to the corresponding task-performance metric, and the cohort-level ranks were then averaged.
To balance accuracy and efficiency for deployment, we introduce a new metric called the Deployability Score (D-Score). The metric for assessing the deployability of PFMs in clinical practice is designed to adhere to the following principles: \\
1. Accuracy, as the top priority for clinicians, should take precedence over efficiency. \\
2. Models with unacceptably low accuracy should receive the minimum score, regardless of their efficiency.\\
3. Models with high computational costs should be penalized in the score to reflect their impact on practical usability.

Based on these principles, we define D-Score as a weighted geometric mean of the normalized task-performance and normalized FLOPs scores. The task-performance metric is Macro-AUC for classification cohorts and C-index for survival cohorts. The FLOPs are measured at the slide-level and averaged across each cohort. Given a set of PFMs denoted by $\{m_i\}_{i=1}^{M}$, the D-Score for the $i$-th model is computed as: $d_i = \overline{p}_i^\alpha \cdot \overline{f}_i^{(1-\alpha)}$, where $\overline{p}_i$ and $\overline{f}_i$ are the normalized task-performance score and the normalized FLOPs score, respectively, and $\alpha$ is the weighting factor for task performance. To align with the first principle, we set $\alpha=0.9$ to prioritize performance. The normalized task-performance score, $\overline{p}_i$, is derived using min-max normalization across evaluated PFMs, ensuring that the lowest-performing model receives a score of zero, which results in a zero D-Score and satisfies the second principle.

To account for the exponentially growing nature of FLOPs, we first transform FLOPs into a logarithmic scale, which compresses the broad range of values while maintaining their relative differences. Next, we normalize these log-FLOPs to the [0, 1] range using a Sigmoid function centered at the median log-scale FLOPs of the evaluated models. The normalized FLOPs are calculated as: $\overline{f}_i = 1 - \frac{1}{1 + e^{-(\log{f_i} - \log{T})} }$. Here $f_i$ represents the raw FLOPs value, and $T$ denotes the median FLOPs value among the evaluated models. The use of ``$1-$'' ensures that larger scores correspond to better efficiency. The subtraction of $\log{T}$ in the Sigmoid function aligns the normalized scores with the moderate baseline, where models with median FLOPs scores serve as the baseline for relative efficiency evaluation. This design emphasizes small or moderate-efficiency models while assigning diminishing scores to models with excessively high FLOPs, reflecting practical constraints in deployment and satisfying the third principle.

\subsection*{Deployment}
To facilitate real-world application, achieving fast inference of PFMs on devices with limited computational resources poses significant challenges. To evaluate the efficiency of LitePath, we compare it with 21 other PFMs, namely H-Optimus-1, H0-mini\cite{h0_mini}, EAGLE\cite{neidlinger2025deep}, UNI2\cite{uni}, MUSK\cite{MUSK}, CONCH1.5\cite{CONCH}, TITAN\cite{TITAN}, Phikon2\cite{Phikon2}, CHIEF\cite{CHIEF}, Virchow2\cite{Virchow2}, GPFM\cite{gpfm}, mSTAR\cite{mstar}, Hibou-L\cite{hibou}, Prov-GigaPath\cite{gigapath}, Virchow\cite{virchow}, UNI\cite{uni}, PLIP\cite{plip}, Phikon\cite{Phikon}, CONCH\cite{CONCH}, CTransPath\cite{ctranspath}, and Resnet50\cite{he2016deep}. We deploy each PFM on both an NVIDIA Jetson Orin Nano Super and an RTX 3090. The Jetson Orin Nano Super supports three power modes; we set its power to 25W for all experiments. For each PFM, we generate a dummy input slide containing 30,000 patches, with each patch resized to the target resolution required by the respective PFM. For LitePath, which incorporates patch selection during inference, we assume a selection of 1,000 patches based on estimated attention scores. We measure the latency of feature extraction and MIL prediction for the entire slide using half-precision computation, as it offers faster processing without empirically compromising accuracy. Importantly, the dummy input is generated directly on the CPU to exclude disk I/O, ensuring a fair comparison across devices. Based on the measured latency, we calculate the throughput of each PFM, defined as the number of slides processed per hour. Additionally, we investigate the number of parameters in each PFM to provide further insight into their computational demands.

For a new downstream task, centralized model development comprises five steps: (1) partition the training WSIs into tissue patches; (2) extract patch features using LiteFM; (3) train the task-specific ABMIL; (4) train the APS scoring network to approximate the ABMIL attention scores; and (5) select $k_u$ and $k_a$ on the internal validation set using the inference-only procedure described above. LiteFM distillation is a one-time procedure and is not repeated for each task.

For the evaluated tasks, the intended deployment package contains the frozen LiteFM encoder, task-specific ABMIL and APS weights, and the validation-selected patch configuration. An end-site user therefore needs only to partition an incoming WSI into tissue patches and execute the packaged inference pipeline, without additional model training or hyperparameter selection.

\subsection*{Computing software and hardware}
In this project, we used Python (v3.10.15) and PyTorch\cite{pytorch}(v.2.5.1, CUDA 11.8) for model training and evaluation. OpenSlide (v3.4.1) and ASlide (\url{https://github.com/MrPeterJin/ASlide}) were utilized to process WSIs. For the distillation training of LitePath, we utilized three PFMs as teachers, Virchow2, H-Optimus-1, and UNI2. The pretrained weights of Virchow2 and H-Optimus-1 are available at Hugging Face (\url{https://huggingface.co/paige-ai/Virchow2}, \url{https://huggingface.co/bioptimus/H-optimus-1}). The pretrained weights of UNI2 are available at GitHub (\url{https://github.com/mahmoodlab/UNI}). The distillation training was performed on $4 \times 80$ GB NVIDIA H800 GPUs with the distributed data-parallel (DDP) technique. The downstream tasks were performed on a single NVIDIA RTX 3090 GPU. We used OriginPro (2024), matplotlib (v.3.9.3), and seaborn (v0.13.2) for visualization.

\section*{Data availability}
This study incorporates 33 public datasets for pretraining, including TCGA \cite{TCGA} (\url{https://portal.gdc.cancer.gov/}), CPTAC \cite{CPTAC} (\url{https://proteomic.datacommons.cancer.gov/pdc/}), PANDA \cite{PANDA} (\url{https://www.kaggle.com/c/prostate-cancer-grade-assessment/data}), NADT-Prostate \cite{NADT-Prostate} (\url{https://www.cancerimagingarchive.net/collection/nadt-prostate/}), BCNB \cite{BCNB} (\url{https://bcnb.grand-challenge.org/}), CAMELYON16 \cite{CAMELYON16} (\url{https://camelyon16.grand-challenge.org/Data/}), CAMELYON17 \cite{CAMELYON17} (\url{https://camelyon17.grand-challenge.org/Data/}), BRACS \cite{Bracs} (\url{https://www.bracs.icar.cnr.it/download/}), TIGER2021 \cite{TIGER2021-dataset} (\url{https://tiger.grand-challenge.org/}), MIDOG2022 \cite{MIDDog-dataset} (\url{https://midog.deepmicroscopy.org/download-dataset/}), AGGC2022 \cite{AGGC} (\url{https://aggc22.grand-challenge.org/}), O.B.R \cite{OBR1, OBR2} (\url{https://www.cancerimagingarchive.net/collection/ovarian-bevacizumab-response/}), ACROBAT2023 \cite{acrobat2023-dataset} (\url{https://acrobat.grand-challenge.org/}), AML-C-LMU \cite{AML-dataset} (\url{https://www.cancerimagingarchive.net/collection/aml-cytomorphology_lmu/}), ARCH \cite{ARCH} (\url{https://warwick.ac.uk/fac/cross_fac/tia/data/arch}), BACH \cite{BACH} (\url{https://zenodo.org/records/3632035}), CAMEL \cite{Camel} (\url{https://drive.google.com/open?id=1brr8CnU6ddzAYT157wkdXjbSzoiIDF9y}), DiagSet \cite{DiagSet} (\url{https://ai-econsilio.diag.pl/}), DLBCL \cite{DLBCL} (\url{https://github.com/stanfordmlgroup/DLBCL-Morph}), GTEx \cite{GTEx} (\url{https://gtexportal.org/home/histologyPage}), HunCRC \cite{HunCRC} (\url{https://www.cancerimagingarchive.net/collection/hungarian-colorectal-screening/}), Janowczyk \cite{Janowczyk} (\url{https://andrewjanowczyk.com/use-case-1-nuclei-segmentation/}), LC25000 \cite{LC2500-dataset} (\url{https://academictorrents.com/details/7a638ed187a6180fd6e464b3666a6ea0499af4af}), MIDOG2021 \cite{MIDDog-dataset} (\url{https://imig.science/midog2021/download-dataset/}), OCELOT \cite{Ocelot} (\url{https://zenodo.org/record/7844149}), Oste. Tumor \cite{OsteTumor} (\url{https://www.cancerimagingarchive.net/collection/osteosarcoma-tumor-assessment/}), PAIP2019 \cite{PAIP2019} (\url{https://paip2019.grand-challenge.org/}), PAIP2020 \cite{PAIP2020} (\url{https://paip2020.grand-challenge.org/}), PAIP2021 (\url{https://paip2021.grand-challenge.org/}), Post-NAT-BRCA \cite{Post-NAT-BRCA} (\url{https://www.cancerimagingarchive.net/collection/post-nat-brca/}), SICAPv2 \cite{sicapv2} (\url{https://data.mendeley.com/datasets/9xxm58dvs3/1}), SLN-Breast \cite{SLN-Breast} (\url{https://www.cancerimagingarchive.net/collection/sln-breast/}), SPIE2019 \cite{SPIE2019} (\url{https://breastpathq.grand-challenge.org/}). 
The multi-center clinical evaluation used a PathBench-derived data collection~\cite{PathBench} together with two additional prospectively accrued cohorts from H1. These human WSI datasets are not publicly available because of patient privacy protections, institutional review board requirements and data-use agreements. Requests for access may be submitted to the corresponding authors and will be considered subject to the relevant ethical approvals, institutional policies and data-use agreements.
The open-access downstream evaluations used BRACS~\cite{Bracs}, CAMELYON16 and CAMELYON17~\cite{CAMELYON16,CAMELYON17}, TCGA~\cite{TCGA}, CRC-MSI~\cite{kather2019deep}, PCAM~\cite{veeling2018rotation}, PanCancer-TIL~\cite{saltz2018spatial}, UniToPatho~\cite{barbano2021unitopatho} and WSSS4LUAD~\cite{han2022wsss4luad,han2022multi}. These datasets are available through the repositories or access routes reported in the corresponding source publications.

% \section*{Code availability}
% The code and model weights of this study are available on GitHub at \url{https://github.com/caiyu6666/LitePath}.

\bibliography{sample}

\section*{Acknowledgements}
This work was supported by the National Natural Science Foundation of China (Project No. 62202403), Research Grants Council of the Hong Kong Special Administrative Region, China (Project No. T45-401/22-N and AoE/E-601/24-N), National Key R\&D Program of China (Project No. 2023YFE0204000), HKSAR RGC General Research Fund (GRF) (Project No. 16208823), and Shenzhen Science and Technology Innovation Committee Fund (Project No. KCXFZ20230731094059008).

% We thank Li Liang, Xiuming Zhang, Feng Gao, Yueping Liu, Ronald Cheong Kin Chan, Zhenhui Li, Jing Cui, Yali Xu, Zhengyu Zhang, Du Cai, On Ki Tang, Chenglong Zhao and the HKUST Smart Lab members for their efforts in collecting and establishing PathBench\cite{PathBench}, which was utilized for the evaluation in this study.

\section*{Ethics declaration}
This project has been reviewed and approved by the Human and Artefacts Research Ethics Committee (HAREC) of The Hong Kong University of Science and Technology (Protocol No. HREP-2025-0322) and the Ethics Committee of Nanfang Hospital (Protocol No. NFEC-2024-535, NFEC-2025-403, and NFEC-2025-419). The study was conducted in accordance with the Declaration of Helsinki.

\section*{Author contributions statement}
Y.C., J.M. and H.C. conceived the study. Y.C. designed and implemented the method, pretrained the model, conducted evaluations, implemented deployment on the Jetson device, prepared the original manuscript draft including visualization, revised the deployment software and produced the demo video. C.J. developed the deployment software and graphical user interface, enhanced visualizations, and contributed to Discussion writing and manuscript refinement. Z.Z. provided pathological expertise and designed and coordinated the crossover reader study. Z.Z., C.Z., Z.C., and Ying T. participated as pathologists in the reader study. J.M. curated pretraining data, contributed to the pretraining design, and provided writing suggestions. Yingxue X., Y.W., Z.G., J.M., Z.Z. and Ling L. processed data for downstream tasks. Y.C., F.Z. and Yingxue X. prepared benchmark results for comparison; F.Z. also assisted with visualization design. Z.G. processed prospective data and prepared code for external validation. Yonghao T. and P.D. contributed to the deployment discussion. Li L., X.Z., F.G., Y.L., R.C.K.C., Z.L., J.C., Yali X., C.Y., X.W., D.C., O.K.T. and C.Z. collected pathology data. All authors reviewed the manuscript draft. H.C. and K.-T.C. supervised the research.

\newpage

\section*{Extended Data}
\setcounter{figure}{0}
\setcounter{table}{0}
% \renewcommand{\figurename}{Extended Figure}
% 修改图编号为 Figure A1, A2, ...
\renewcommand{\thefigure}{A\arabic{figure}}
\renewcommand{\thetable}{A\arabic{table}}

\begin{figure}[htbp]
\centering
\begin{minipage}[t]{0.49\textwidth}
\begin{algorithm}[H]
\caption{Inference pipeline of LitePath}
\label{alg:litepath*}
\begin{algorithmic}[1]
\Require $E_{pre}$ and $E_{post}$ \Comment{\parbox[t]{.57\linewidth}{LiteFM backbone is divided into $E_{pre}$ and $E_{post}$}}
\Require $\phi, g, h$ \Comment{\parbox[t]{.57\linewidth}{APS scoring network, aggregator, and output head}}
\Require $k_u, k_a$  \Comment{\parbox[t]{.57\linewidth}{Number of patches selected by uniform and attention-based strategies}}
\Require $\mathbf{X}=\{\mathbf{x}_i\}_{i=1}^N$ \Comment{\parbox[t]{.57\linewidth}{Input WSI containing $N$ patches}}
\bigskip
\State $\mathbf{H}_i = E_{pre} (\mathbf{x}_i)$, for $i=1,...,N$. \Comment{\parbox[t]{.4\linewidth}{Shallow features of all patches}}
\State $\mathcal{S} = \text{APS}(\mathbf{H}, \phi, k_u, k_a)$  \Comment{\parbox[t]{.4\linewidth}{Selected indices ($k_u$ uniform and $k_a$ attention)}}
\State $\mathbf{F}_s = E_{post}(\mathbf{H}_s)$, for $s \in \mathcal{S}$.  \Comment{\parbox[t]{.4\linewidth}{Features of ($k_u+k_a$) selected patches}}
\State $\mathbf{Z} = g({\mathbf{F}_{s \in \mathcal{S}}})$  \Comment{\parbox[t]{.4\linewidth}{Feature aggregation}}
\State $\mathbf{Y} = h(\mathbf{Z})$  \Comment{\parbox[t]{.4\linewidth}{Prediction}}
\end{algorithmic}
\end{algorithm}
\end{minipage}
\hfill
\begin{minipage}[t]{0.49\textwidth}
\begin{algorithm}[H]
\caption{Inference pipeline of conventional PFMs}
\label{alg:PFM}
\begin{algorithmic}[1]
\Require $E$ \Comment{\parbox[t]{.45\linewidth}{PFM backbone}}
\Require $g, h$ \Comment{\parbox[t]{.45\linewidth}{Aggregator and output head}}
\Require $\mathbf{X}=\{\mathbf{x}_i\}_{i=1}^N$ \Comment{\parbox[t]{.45\linewidth}{Input WSI containing $N$ patches}}
\bigskip
\State $\mathbf{F}_i = E (\mathbf{x}_i)$, for $i=1,...,N$. \Comment{\parbox[t]{.45\linewidth}{Features of all patches ($N$)}}
\State $\mathbf{Z} = g({\mathbf{F}})$  \Comment{\parbox[t]{.45\linewidth}{Feature aggregation}}
\State $\mathbf{Y} = h(\mathbf{Z})$    \Comment{\parbox[t]{.45\linewidth}{Prediction}}
\end{algorithmic}
\end{algorithm}
\end{minipage}
\end{figure}

\begin{table}
\centering
\caption{\textbf{Number of slides and processed patches from the 33 datasets used for distillation pretraining.} "-" indicates datasets that provide only ROIs. Dataset selection and processing follow GPFM~\cite{gpfm}.} \label{tab:dataset_distributation}
\begin{tabular}{lll}
\toprule
Dataset Name & Number of Slides & Total Patches \\
\midrule
TCGA\cite{TCGA}&26,285&120,496,200\\
GTExPortal\cite{GTEx}&24,467&31,892,017\\
CPTAC\cite{CPTAC}&7,164&11,768,225\\
CAMELYON17\cite{CAMELYON17}&841&4,612,382\\
HunCRC\cite{HunCRC}&200&3,369,925\\
BRACS\cite{Bracs}&381&2,992,229\\
DiagSet\cite{DiagSet}&825&2,500,385\\
AGGC2022\cite{AGGC}&286&2,130,584\\
CAMELYON16\cite{CAMELYON16}&288&1,706,890\\
DLBCL\cite{DLBCL}&203&1,524,388\\
PAIP2020\cite{PAIP2020}&118&1,362,725\\
O.B.R\cite{OBR1,OBR2}&283&1,159,516\\
PAIP2021&220&1,048,840\\
NADT-Prostate\cite{NADT-Prostate}&1,303&919,847\\
PANDA\cite{PANDA}&7,114&905,206\\
PAIP2019&96&505,356\\
TIGER2021&174&312,835\\
BCNB\cite{BCNB}&1,036&263,734\\
Post-NAT-BRCA\cite{Post-NAT-BRCA}&96&241,547\\
SLN-Breast\cite{SLN-Breast}&129&139,166\\
BACH\cite{BACH}&30&108,256\\
ACROBAT2023&153&76,128\\
MIDOG2022&395&43,342\\
ARCH\cite{ARCH}&-&25,919\\
MIDOG2021&193&24,025\\
LC25000&-&19,678\\
SICAPv2&-&18,783\\
AML-C-LMU&-&18,365\\
CAMEL\cite{Camel}&-&16,744\\
OCELOT\cite{Ocelot}&-&3,201\\
SPIE2019&-&2,579\\
Janowczyk\cite{Janowczyk}&-&2,260\\
Oste. Tumor\cite{OsteTumor}&-&1,391\\
\midrule
Total&72,280&190,212,668\\
\bottomrule
\end{tabular}
\end{table}
\begin{table}[!t]
\centering
\caption{\textbf{Summary of classification tasks in the multi-center clinical evaluation.} H1--H9 denote the nine collaborating hospitals. For each cohort, counts follow the order of the corresponding label categories and are reported as cases (WSIs). Parentheses are omitted when each case contributes one WSI or when the task is defined at the WSI level. Pros., prospective cohort.}
\label{tab:task_summary}
\resizebox{\linewidth}{!}{
\begin{tabular}{lllll}
\toprule
Organ  & Task  & Label set  & Internal cohort  & External/prospective cohort   \\ \midrule
\multirow{5}{*}{Lung}     
    & Primary/Metastatic Classification & Primary, Metastatic      & H1: 389(686), 457(736)                    & \begin{tabular}[c]{@{}l@{}}H5: 237, 256\\H6: 465(744), 361(678)\end{tabular}     \\
    & Cancer Subtyping  & LNET, MIA, ASC, IAC, LUSC, AIS   & H1: 131(170), 121(167), 19(34), 150(181), 123(255), 150(271)  & H1 [Pros.]: 5(23), 150(331), 7(34), 150(647), 70(400), 127(367)    \\ 
    & NSCLC Subtyping  & LUAD, LUSC   & H1: 300, 300  & -  \\ 
    & Lymph Node Metastasis Prediction  & metastasis-positive, metastasis-negative   & H1: 71(231), 250(828)  & -  \\ 
    & IHC Status Prediction - P63  & P63-negative, P63-positive   & H1: 197(249), 90(105)  & H1 [Pros.]: 24(126), 34(171)   \\ 
\cmidrule{1-5} 
\multirow{8}{*}{Breast}           
    & TNM-N Staging (N0/N+)    & N0, N+                                    & H2: 343(916), 125(381)                          & -                            \\
    & pTNM Overall Staging (I/II/III)    & I, II, III                      & H2: 192(451), 232(727), 43(116)                 & -                            \\ 
    & Molecular Subtyping      & Luminal A, Luminal B1, Luminal B2, TNBC, HER-2  & H2: 307(310), 614(618), 243(268), 589(1932), 292(323)  & H9: 102, 89, 24, 101, 102 \\
    & IHC Status Prediction - AR           & AR-negative, AR-positive      & H2: 463(731), 677(841)                          & -    \\
    & IHC Status Prediction - ER           & ER-negative, ER-positive      & H2: 767(1264), 781(786)                         & -    \\
    & IHC Status Prediction - PR           & PR-negative, PR-positive      & H2: 623(1108), 933(950)                         & -    \\
    & IHC Status Prediction - HER2         & HER2-negative, HER2-positive  & H2: 511(743), 833(975)                          & -    \\
    & IHC Status Prediction - CK5          & CK5-negative, CK5-positive    & H2: 753(859), 208(379)                          & -    \\ 
\cmidrule{1-5}
\multirow{9}{*}{Gastric}         
    & Cancer Grading           & G1+G2, G3                                 & H1: 81(82), 318(319)                            & \begin{tabular}[c]{@{}l@{}}H3: 55, 190\\H4: 62, 258\end{tabular} \\
    & Pathological Subtyping   & SRCC, TAC, NOS                            & H1: 163, 166(167), 66(67)               & H3: 59, 0, 195   \\
    & TNM-N Staging (N0/N+)    & N0, N+                                    & H1: 186(188), 212                       & H3: 85, 175      \\
    & Perineural Invasion Detection        & PNI-positive, PNI-negative    & H1: 255(256), 141(142)                  & H3: 156, 76       \\
    & Vascular Invasion Detection          & VI-positive, VI-negative      & H1: 197(198), 198(199)                  & H3: 140, 90       \\ 
    & Normal/Abnormal Classification     & Normal, Abnormal                & H7: 733, 1967                                   & -   \\
    & Intestinal Metaplasia Classification & IM, Not IM                    & H7: 270, 2430                                   & -   \\
    & IHC Status Prediction - HER2         & 0/1+, 2+/3+                   & H1+H3+H4: 549, 126                              & -    \\
    & IHC Status Prediction - S-100        & 0, 1+                         & H1+H3+H4: 90, 270                               & -    \\ 
\cmidrule{1-5}
\multirow{4}{*}{Colorectal}
    & TNM-N Staging (N0/N+)                & N0, N+                        & H8: 367(1848), 230(871)                         & -   \\
    & TNM-T Staging (T1+T2/T3+T4)          & T1+T2, T3+T4                  & H8: 76(319), 519(2391)                          & -   \\
    & TNM-T Staging (T1/T2/T3/T4)          & T1, T2, T3, T4                & H8: 20(75), 56(244), 440(2130), 79(261)         & -   \\
    & Consensus Molecular Subtyping        & CMS1, CMS2, CMS3, CMS4        & H8: 76(372), 239(1061), 86(393), 187(857)       & -    \\ 
\bottomrule
\end{tabular}
}
\end{table}

\begin{table}[!t]
\centering
\caption{\textbf{Summary of survival-analysis tasks in the multi-center clinical evaluation.} H1--H9 denote the nine collaborating hospitals. Cohort size is reported as patients (WSIs), with parentheses omitted when each patient contributes one WSI. Events were defined as recurrence or progression for disease-free survival (DFS), death from any cause for overall survival (OS), and tumor-related death for disease-specific survival (DSS). Event percentages were calculated using the number of patients in each cohort.}
\label{tab:survival_task_summary}
\begin{tabular}{lllll}
\toprule
Organ & Endpoint & Cohort & Patients (WSIs) & Events, n (\%) \\ \midrule
\multirow{3}{*}{Breast}
    & \multirow{2}{*}{DFS}
    & Internal-H2 & 451 (1,270) & 71 (15.7\%) \\
    &
    & External-H9 & 79 & 3 (3.8\%) \\ \cmidrule{2-5}
    & OS
    & Internal-H2 & 451 (1,270) & 59 (13.1\%) \\ \midrule
\multirow{2}{*}{Gastric}
    & DFS & Internal-H3+H4 & 580 & 132 (22.8\%) \\
    & OS  & Internal-H3+H4 & 580 & 206 (35.5\%) \\ \midrule
\multirow{3}{*}{Colorectal}
    & DFS & Internal-H8 & 608 (2,779) & 219 (36.0\%) \\
    & OS  & Internal-H8 & 608 (2,779) & 168 (27.6\%) \\
    & DSS & Internal-H1 & 294 (301) & 42 (14.3\%) \\
\bottomrule
\end{tabular}
\end{table}

% \usepackage{multirow}
% \usepackage{graphicx}
\begin{table}[!t]
\centering
\caption{\textbf{APS configuration and slide information of each cohort.} H1--H9 denote the nine collaborating hospitals. Patch counts per case are reported as the mean, with the 2.5th--97.5th percentile interval in parentheses.}
\label{tab:cohort_summary}
\resizebox{0.68\textwidth}{!}{
\begin{tabular}{lllll}
\toprule
Organ                     & Task                                               & APS ($k_u + k_a$)          & Cohort      & \#patches/case         \\ \midrule
\multirow{9}{*}{Lung}     & \multirow{3}{*}{Primary/Metastatic Classification} & \multirow{3}{*}{2,000 + 0}  & Internal-H1 & 13,495 (507--61,437)    \\
                          &                                                    &                            & External-H5 & 5,007 (109--14,074)     \\
                          &                                                    &                            & External-H6 & 19,817 (469--79,069)    \\ \cmidrule{2-5} 
                          & \multirow{2}{*}{Cancer Subtyping}                  & \multirow{2}{*}{1,000 + 0}  & Internal-H1 & 13,763 (1,005--50,752)   \\
                          &                                                    &                            & Prospective-H1 & 76,646 (2,625--534,209)  \\ \cmidrule{2-5} 
                          & NSCLC Subtyping                                    & 1,900 + 100                 & Internal-H1 & 10,178 (1,745--23,874)   \\ \cmidrule{2-5} 
                          & Lymph Node Metastasis Prediction                   & 1,900 + 100                 & Internal-H1 & 32,229 (5,206--80,441)   \\
\bottomrule
\end{tabular}
}
\end{table}

\begin{table}
\centering
\caption{\textbf{Hyperparameters for distillation pretraining of LiteFM.} Pretraining was conducted on 4$\times$80 GB H800 GPUs.} \label{tab:pretraining}
\begin{tabular}{lll}
\toprule
&Hyperparameter & Value \\
\midrule
\multirow{7}{*}{Model} 
     & Input size & 224 \\
     & Patch size & 16 \\
     & Embedding dim & 384 \\
     & Attention blocks & 12 \\
     & Heads & 6 \\
     & MLP ratio & 4 \\
     & Output dim & 1024 \\
     & Proj. dim for teachers & 2560, 1536, 1536 \\
     & \textbf{\#Params.}     & \textbf{22.06M}    \\
\midrule
\multirow{12}{*}{Optimization}
     & Iterations   &  100,000  \\
     & Optimizer    &  AdamW    \\
     & Total batch size         & 2,048 \\
     & Weight decay             & 0.05  \\
     & Gradient clip            & 1.0   \\
     & Learning rate            & 3e-3  \\
     & Scheduler                & CosineLRScheduler \\
     & Warmup steps             & 5000  \\
     & Warmup lr init.          & 1e-6  \\
     & Minimum lr               & 1e-5  \\
     & $\alpha, \beta, \gamma$  & 0.4, 0.3, 0.3 \\
\bottomrule
\end{tabular}
\end{table}

\begin{table}
\centering
\caption{\textbf{Hyperparameters for training ABMIL on classification and survival tasks.}}
\label{tab:abmil}
\begin{tabular}{lll}
\toprule
Hyperparameter          & Classification & Survival \\
\midrule
Hidden dim              & 512           & 512 \\
Dropout                 & 0.25          & 0.25 \\
Batch size              & 1             & 1 \\
Epochs                  & 50            & 30 \\
Optimizer               & Adam          & Adam \\
Learning rate           & 2e-4          & 2e-4 \\
Scheduler               & Cosine        & Cosine \\
Weight decay            & 1e-5          & 1e-5 \\
Loss                    & Cross-entropy & Discrete-time NLL ($\alpha=0$) \\
Early stopping          & Yes           & No \\
\bottomrule
\end{tabular}
\end{table}

\begin{table}
\centering
\caption{\textbf{Hyperparameters for training APS scoring network.}}
\label{tab:aps}
\begin{tabular}{ll}
\toprule
Hyperparameter          & Value             \\
\midrule
Hidden dim              & 512               \\
Dropout                 & 0.25              \\
Batch size              & 1                 \\
Epochs                  & 100               \\
Optimizer               & Adam              \\
Learning rate           & 2e-4              \\
Scheduler               & Cosine            \\
Weight decay            & 1e-5              \\
Loss                    & Soft cross-entropy \\
Temperature             & 0.7               \\
\textbf{\#Params. }     & \textbf{0.46M}    \\
\bottomrule
\end{tabular}
\end{table}

\begin{table}
     \centering
     \caption{\textbf{Hyperparameters for training linear probe.}}
     \label{tab:linear_probe}
     \begin{tabular}{ll}
     \toprule
     Hyperparameter          & Value             \\
     \midrule
     Batch size              & 64                 \\
     Epochs                  & 500               \\
     Early stopping          & Yes               \\
     Optimizer               & Adam              \\
     Learning rate           & 2e-4              \\
     Scheduler               & Cosine            \\
     Weight decay            & 1e-5              \\
     Loss                    & Cross-entropy      \\
     \bottomrule
     \end{tabular}
\end{table}

\begin{table}
\centering
\caption{\textbf{Overview of four models in LiteFM family.}}
\label{tab:litepath_family}
\begin{tabular}{llll}
\toprule
Model Name      & Backbone  & \#Params. & Teachers               \\
\midrule
LiteFM-L      & ViT-B     & 86.59M    & Virchow2, H-Optimus-1, UNI2  \\
LiteFM        & ViT-S     & 22.06M    & Virchow2, H-Optimus-1, UNI2  \\
LiteVirchow2    & ViT-S     & 22.06M    & Virchow2                     \\
LiteFM-S      & ViT-Ti    & 5.72M     & Virchow2, H-Optimus-1, UNI2  \\
\bottomrule
\end{tabular}
\end{table}

\begin{table}[!t]
    \centering
    \caption{\textbf{D-Score comparison of models on 45 cohorts.} Note that the inclusion of additional models may alter the D-Scores of existing models due to performance and FLOPs normalization.}
    \label{tab:dscore}
    \resizebox{\linewidth}{!}{
    \begin{tabular}{llcccccccccccc}
    \toprule
    Organ & Cohort & LitePath & H0-mini & Virchow2 & mSTAR & GPFM & H-Optimus-1 & UNI & UNI2 & Phikon2 & CTransPath & Prov-GigaPath & Resnet50 \\
    \midrule
    \multirow{9}{*}{Lung} & Primary/Metastatic Classification [Internal-H1] & 0.9231 & \textbf{0.9334} & 0.8619 & 0.8919 & 0.8800 & 0.8365 & 0.8680 & 0.8246 & 0.9089 & 0.9039 & 0.8481 & 0.0000 \\
     & Primary/Metastatic Classification [External-H5] & 0.9099 & 0.8972 & 0.8333 & \textbf{0.9330} & 0.8424 & 0.7389 & 0.8891 & 0.8470 & 0.8838 & 0.8437 & 0.7223 & 0.0000 \\
     & Primary/Metastatic Classification [External-H6] & \textbf{0.9067} & 0.8801 & 0.8760 & 0.8906 & 0.8525 & 0.8159 & 0.8665 & 0.7958 & 0.8470 & 0.8101 & 0.7553 & 0.0000 \\
     & Cancer Subtyping [Internal-H1] & 0.5976 & 0.6127 & 0.6239 & 0.8143 & \textbf{0.9199} & 0.6381 & 0.7813 & 0.6387 & 0.8464 & 0.7770 & 0.8161 & 0.0000 \\
     & Cancer Subtyping [Prospective-H1] & 0.6946 & 0.7308 & 0.7303 & 0.3437 & 0.2334 & \textbf{0.8365} & 0.1559 & 0.5871 & 0.0000 & 0.0832 & 0.4174 & 0.5833 \\
     & NSCLC Subtyping [Internal-H1] & \textbf{0.8931} & 0.8712 & 0.8400 & 0.8021 & 0.7625 & 0.7612 & 0.7967 & 0.8701 & 0.6004 & 0.6715 & 0.7122 & 0.0000 \\
     & Lymph Node Metastasis Prediction [Internal-H1] & \textbf{0.9505} & 0.9107 & 0.7414 & 0.7987 & 0.7324 & 0.8365 & 0.7755 & 0.7202 & 0.7938 & 0.9490 & 0.6942 & 0.0000 \\
\bottomrule
    \end{tabular}
    }
    \end{table}

\begin{figure}[hp]
\centering
\includegraphics[width=\linewidth]{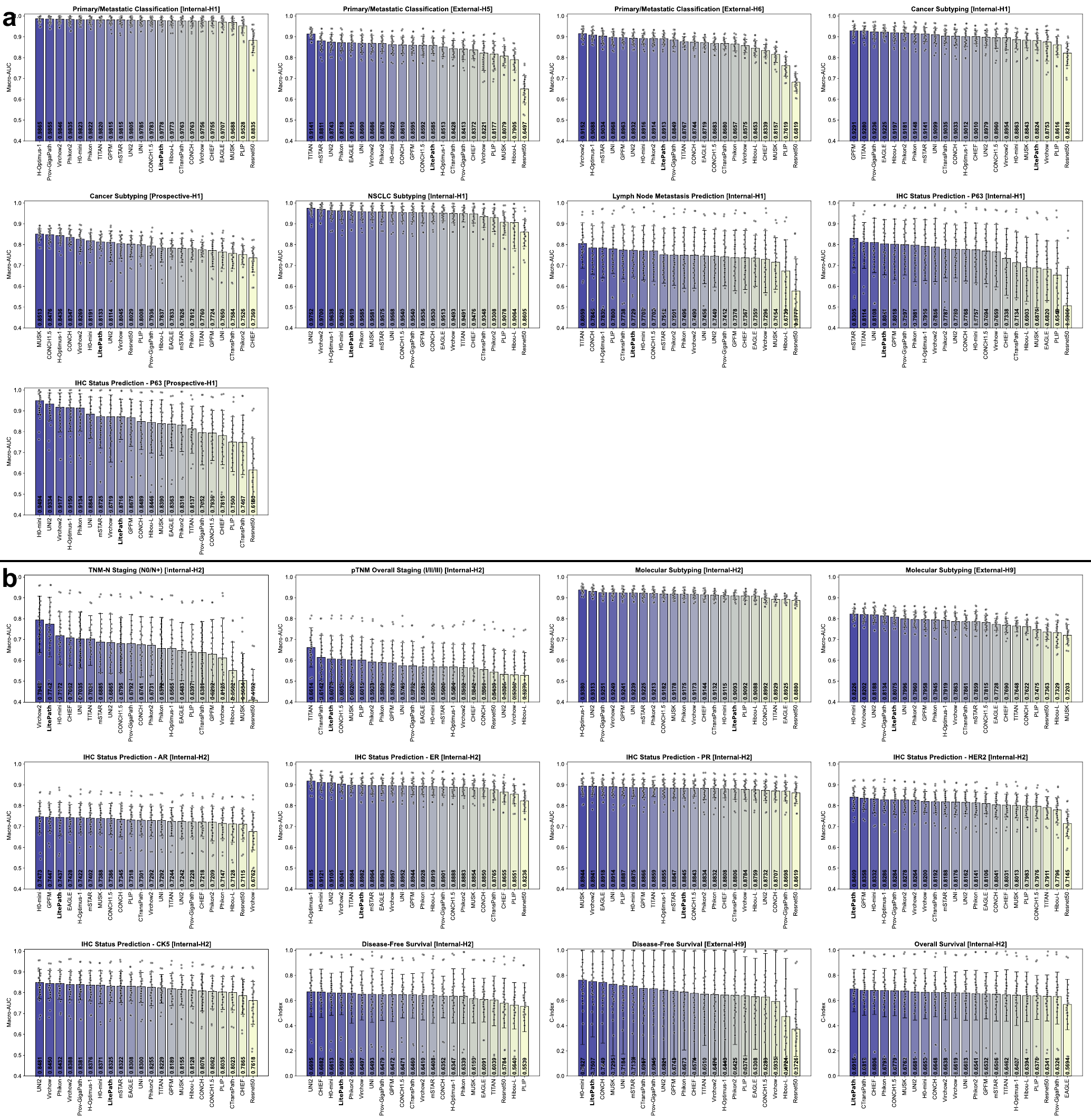}
\caption{\textbf{Performance comparison of PFMs (Part I).} 
\textbf{a}, Results on lung cancer tasks. 
\textbf{b}, Results on breast cancer tasks.}
\label{fig:auc_hist1}
\end{figure}

\begin{figure}[hp]
\centering
\includegraphics[width=\linewidth]{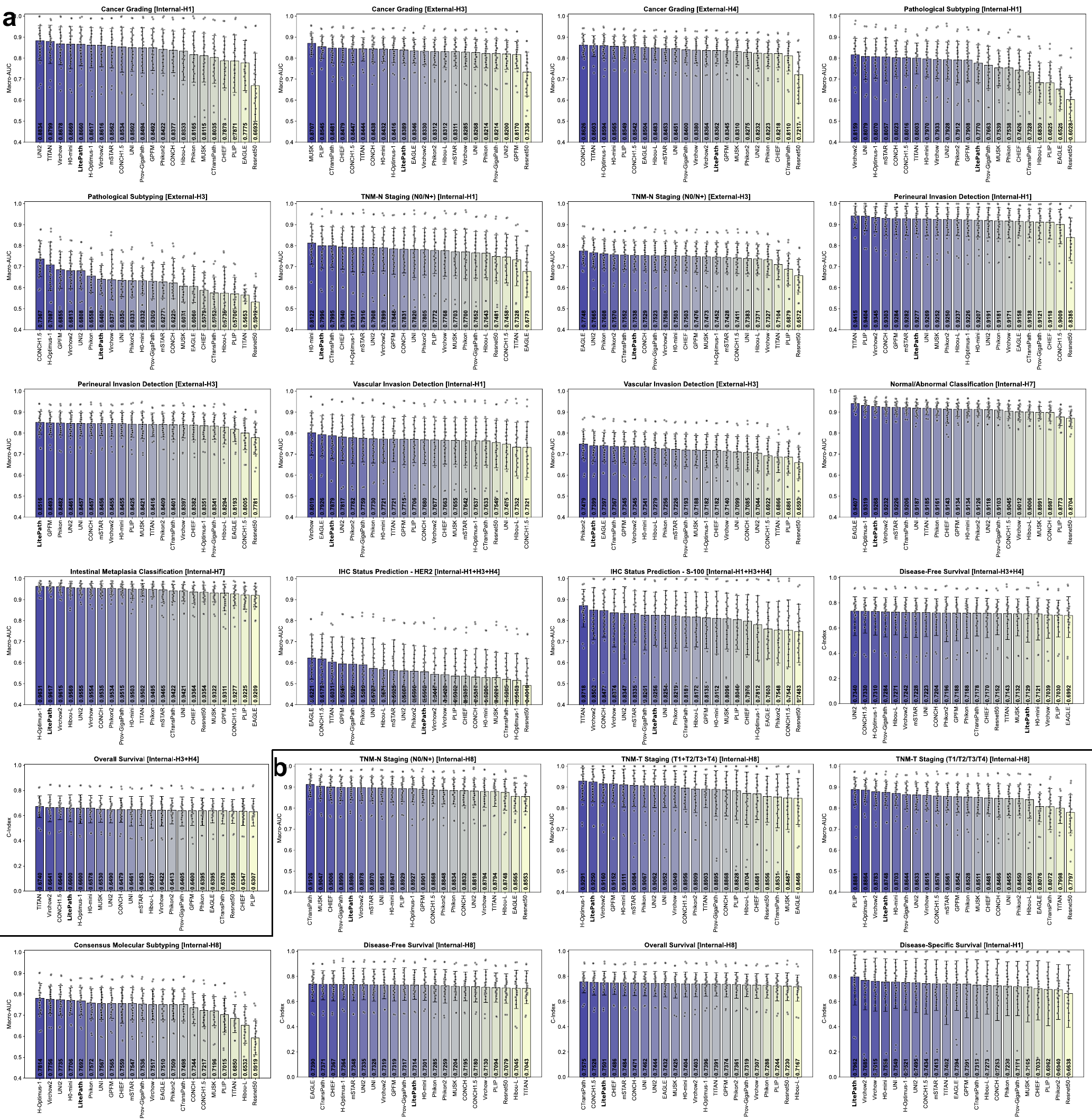}
\caption{\textbf{Performance comparison of PFMs (Part II).}  
\textbf{a}, Results on gastric cancer tasks. 
\textbf{b}, Results on colorectal cancer tasks.
}
\label{fig:auc_hist2}
\end{figure}

\begin{figure}[hp]
\centering
\includegraphics[width=\linewidth]{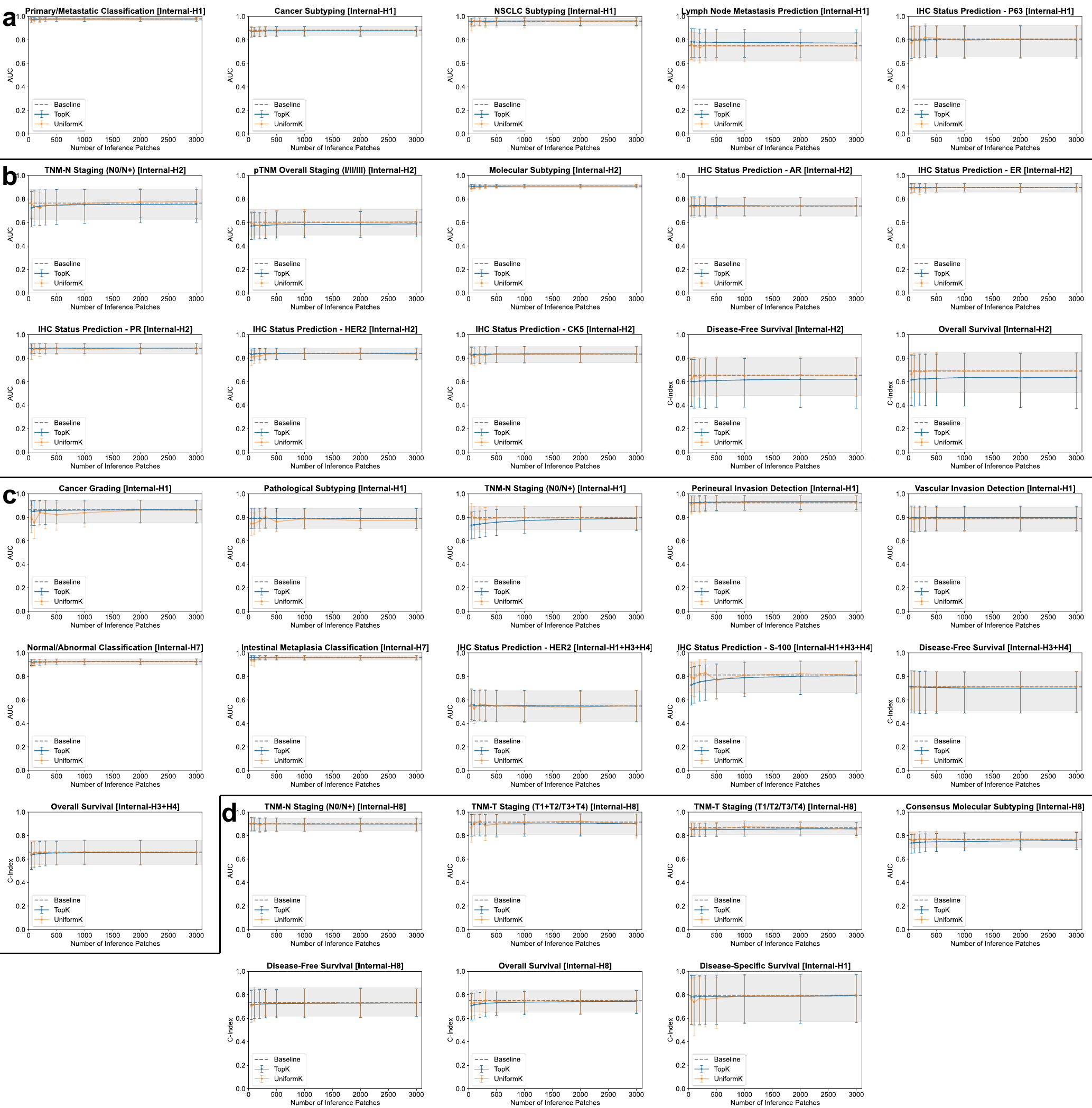}
\caption{\textbf{Performance of inference with partial patches.}
\textbf{a--d}, Results on lung, breast, gastric, and colorectal cancer tasks, respectively. Each panel shows mean AUC or C-index with 95\% confidence intervals for three inference strategies evaluated across varying k: baseline (inference using all patches), top-k (inference using k patches with the highest final attention scores from ABMIL), and uniform-k (inference using k patches selected uniformly by index).}
\label{fig:topk}
\end{figure}

\begin{figure}[hp]
\centering
\includegraphics[width=\linewidth]{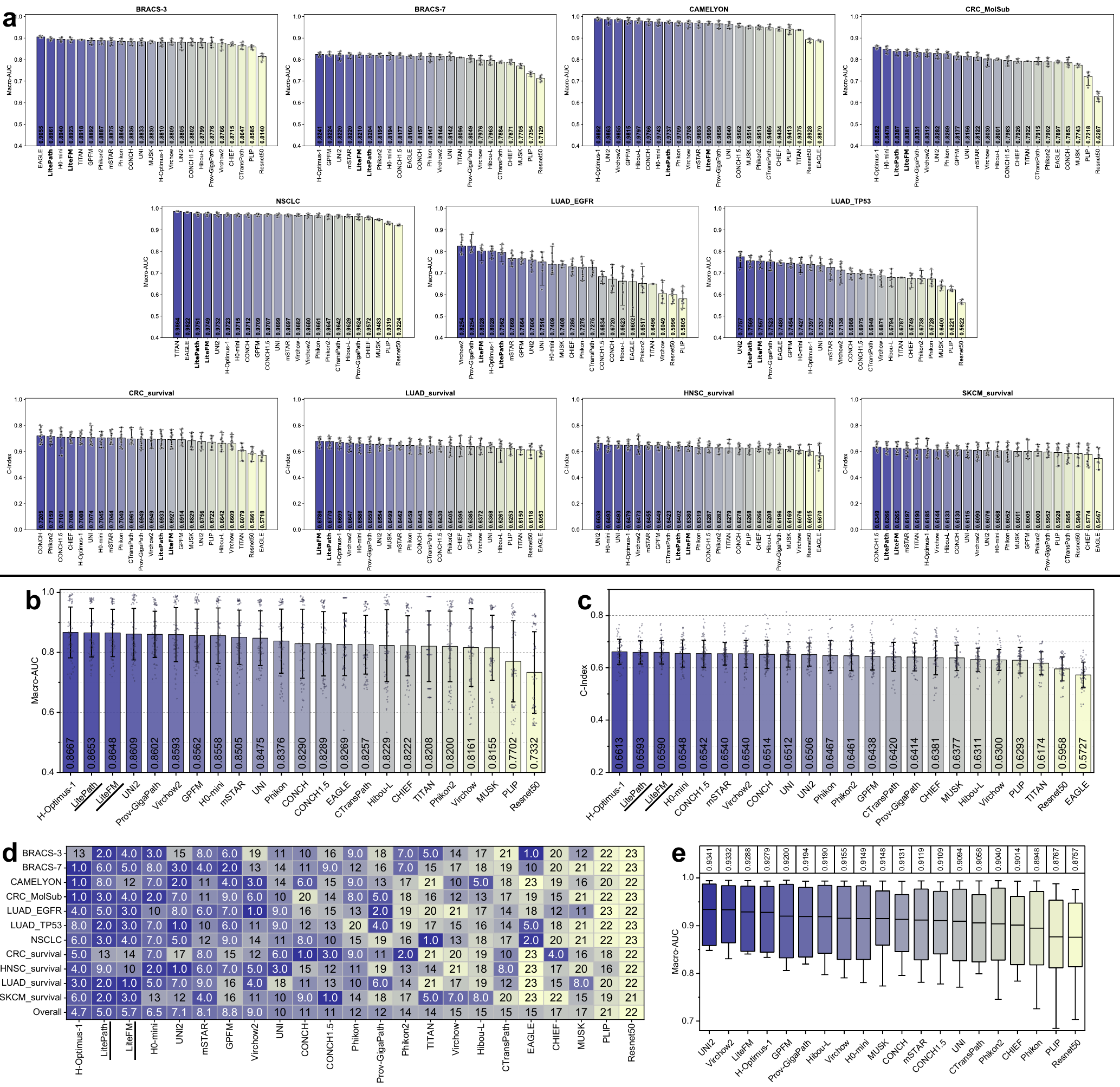}
\caption{\textbf{Comparison of PFMs on public tasks.} 
\textbf{a}, Performance across 7 slide-level classification and 4 slide-level survival tasks. 
\textbf{b}, Average Macro-AUC across public slide-level classification tasks. 
\textbf{c}, Average C-index across public slide-level survival tasks. 
\textbf{d}, Model performance rankings across public slide-level tasks, with the overall average rank.
\textbf{e}, Average Macro-AUC across 5 patch-level ROI classification tasks. Task-wise performance is shown in Extended Data Table~\ref{tab:roi_classification}.
}
\label{fig:auc_hist3}
\end{figure}

\begin{table}[!t]
\centering
\caption{Mean AUC (95\% CI) of models on Primary/Metastatic Classification of lung cancer. One internal cohort (H1) and two external cohorts (H5, H6) are used for evaluation.}
\label{tab:nanfang_primary_metastatic}
\begin{tabular}{lccc}
\toprule
Model & Internal (H1) & External (H5) & External (H6) \\
\midrule
Resnet50 & 0.8835 (0.8236--0.9329) & 0.6497 (0.5824--0.7129) & 0.6819 (0.6393--0.7225) \\
CTransPath & 0.9763 (0.9560--0.9913) & 0.8428 (0.8010--0.8825) & 0.8680 (0.8415--0.8922) \\
CONCH & 0.9763 (0.9524--0.9922) & 0.8610 (0.8190--0.8990) & 0.8744 (0.8434--0.9000) \\
Phikon & 0.9822 (0.9595--0.9968) & 0.8718 (0.8326--0.9093) & 0.8657 (0.8321--0.8951) \\
PLIP & 0.9528 (0.9189--0.9799) & 0.8177 (0.7403--0.8772) & 0.7619 (0.7108--0.8038) \\
UNI & 0.9785 (0.9579--0.9932) & 0.8690 (0.8253--0.9082) & 0.8968 (0.8601--0.9235) \\
Virchow & 0.9756 (0.9524--0.9921) & 0.8221 (0.7385--0.8771) & 0.8575 (0.8239--0.8880) \\
\bottomrule
\end{tabular}
\end{table}

\begin{table}[!t]
\centering
\caption{Mean AUC (95\% CI) of models on Cancer Subtyping of lung cancer. One internal cohort (H1) and one prospective cohort (H1) are used for evaluation.}
\label{tab:nanfang_lung_finegrained}
\begin{tabular}{lcc}
\toprule
Model & Internal (H1) & Prospective (H1) \\
\midrule
Resnet50 & 0.8218 (0.7623--0.8689) & 0.8029 (0.7418--0.8425) \\
CTransPath & 0.9035 (0.8698--0.9331) & 0.7584 (0.6846--0.8074) \\
CONCH & 0.9033 (0.8610--0.9364) & 0.8347 (0.7998--0.8732) \\
Phikon & 0.9148 (0.8757--0.9472) & 0.7812 (0.7302--0.8201) \\
PLIP & 0.8616 (0.8138--0.9011) & 0.8008 (0.7439--0.8423) \\
UNI & 0.9099 (0.8691--0.9415) & 0.7650 (0.6846--0.8278) \\
Virchow & 0.8753 (0.7836--0.9313) & 0.8045 (0.7350--0.8460) \\
Prov-GigaPath & 0.9236 (0.8902--0.9525) & 0.7936 (0.7475--0.8336) \\
Hibou-L & 0.9197 (0.8816--0.9496) & 0.7837 (0.7166--0.8289) \\
mSTAR & 0.9141 (0.8775--0.9441) & 0.7826 (0.7153--0.8299) \\
GPFM & \textbf{0.9291 (0.9012--0.9546)} & 0.7724 (0.6934--0.8199) \\
Virchow2 & 0.8954 (0.8374--0.9377) & 0.8269 (0.7664--0.8721) \\
CHIEF & 0.9010 (0.8672--0.9316) & 0.7369 (0.6687--0.7847) \\
Phikon2 & 0.9181 (0.8863--0.9459) & 0.7526 (0.7045--0.7948) \\
CONCH1.5 & 0.8960 (0.8453--0.9371) & 0.8476 (0.8082--0.8771) \\
TITAN & 0.9280 (0.8920--0.9524) & 0.7760 (0.7407--0.8124) \\
MUSK & 0.8843 (0.8466--0.9169) & \textbf{0.8513 (0.8232--0.8755)} \\
UNI2 & 0.8979 (0.8590--0.9312) & 0.8114 (0.7191--0.8658) \\
H-Optimus-1 & 0.9012 (0.8538--0.9386) & 0.8436 (0.8069--0.8770) \\
H0-mini & 0.8863 (0.8429--0.9201) & 0.8191 (0.7769--0.8565) \\
EAGLE & 0.9225 (0.9008--0.9419) & 0.7833 (0.7413--0.8210) \\
LitePath & 0.8824 (0.8374--0.9172) & 0.8133 (0.7794--0.8476) \\
\midrule
LiteFM-S & 0.8838 (0.8303--0.9208) & 0.8034 (0.7722--0.8333) \\
LiteVirchow2 & 0.8884 (0.8541--0.9189) & 0.7997 (0.7528--0.8350) \\
LiteFM & 0.8829 (0.8384--0.9168) & 0.8122 (0.7760--0.8484) \\
LiteFM-L & 0.8847 (0.8430--0.9182) & 0.8207 (0.7933--0.8479) \\
\bottomrule
\end{tabular}
\end{table}

\begin{table}[!t]
\centering
\caption{Mean AUC (95\% CI) of models on NSCLC Subtyping and Lymph Node Metastasis Prediction of lung cancer. One internal cohort (H1) is used for evaluation.}
\label{tab:nanfang-lung-nsclc-lymph}
\begin{tabular}{lcc}
\toprule
Model & NSCLC Subtyping & Lymph Node Metastasis \\
\midrule
Resnet50 & 0.8605 (0.7910--0.9202) & 0.5777 (0.4158--0.7325) \\
CTransPath & 0.9348 (0.8831--0.9742) & 0.7738 (0.6361--0.8932) \\
CONCH & 0.9530 (0.9116--0.9841) & 0.7845 (0.6582--0.8908) \\
Phikon & 0.9585 (0.9197--0.9856) & 0.7496 (0.5907--0.8822) \\
PLIP & 0.9078 (0.8465--0.9568) & 0.7800 (0.6613--0.8842) \\
UNI & 0.9568 (0.9131--0.9852) & 0.7456 (0.6128--0.8684) \\
Virchow & 0.9493 (0.9055--0.9825) & 0.7296 (0.5691--0.8673) \\
Prov-GigaPath & 0.9540 (0.9112--0.9839) & 0.7412 (0.6021--0.8653) \\
Hibou-L & 0.9064 (0.7811--0.9756) & 0.6739 (0.4719--0.8236) \\
mSTAR & 0.9575 (0.9107--0.9875) & 0.7512 (0.6152--0.8756) \\
GPFM & 0.9536 (0.9050--0.9852) & 0.7378 (0.5786--0.8689) \\
Virchow2 & 0.9700 (0.9371--0.9927) & 0.7490 (0.5907--0.8811) \\
CHIEF & 0.9476 (0.9038--0.9809) & 0.7367 (0.5847--0.8686) \\
Phikon2 & 0.9308 (0.8632--0.9757) & 0.7501 (0.6050--0.8784) \\
CONCH1.5 & 0.9540 (0.9127--0.9830) & 0.7702 (0.6238--0.8902) \\
TITAN & 0.9491 (0.9085--0.9799) & \textbf{0.8059 (0.6868--0.9055)} \\
MUSK & 0.9581 (0.9196--0.9852) & 0.7154 (0.5855--0.8333) \\
UNI2 & \textbf{0.9752 (0.9464--0.9940)} & 0.7449 (0.5960--0.8726) \\
H-Optimus-1 & 0.9638 (0.9265--0.9901) & 0.7839 (0.6403--0.9034) \\
H0-mini & 0.9625 (0.9195--0.9887) & 0.7703 (0.6300--0.8909) \\
EAGLE & 0.9513 (0.9150--0.9783) & 0.7355 (0.6001--0.8450) \\
LitePath & 0.9619 (0.9209--0.9887) & 0.7729 (0.6442--0.8848) \\
\midrule
LiteFM-S & 0.9578 (0.9162--0.9856) & 0.7385 (0.6050--0.8541) \\
LiteVirchow2 & 0.9646 (0.9298--0.9894) & 0.7463 (0.6132--0.8640) \\
LiteFM & 0.9603 (0.9195--0.9870) & 0.7509 (0.6211--0.8658) \\
LiteFM-L & 0.9597 (0.9179--0.9899) & 0.7596 (0.6261--0.8773) \\
\bottomrule
\end{tabular}
\end{table}

\begin{table}[!t]
\centering
\caption{Mean AUC (95\% CI) of models on IHC Status Prediction - P63 of lung cancer. One internal cohort (H1) and one prospective cohort (H1) are used for evaluation.}
\label{tab:nanfang_lung_p63}
\begin{tabular}{lcc}
\toprule
Model & Internal (H1) & Prospective (H1) \\
\midrule
Resnet50 & 0.5066 (0.3278--0.6867) & 0.6163 (0.4639--0.7674) \\
CTransPath & 0.7134 (0.5446--0.8607) & 0.7487 (0.5940--0.8787) \\
CONCH & 0.7768 (0.6255--0.9053) & 0.8489 (0.7139--0.9449) \\
Phikon & 0.7981 (0.6220--0.9256) & 0.9134 (0.8026--0.9820) \\
PLIP & 0.6540 (0.4778--0.8172) & 0.7500 (0.6083--0.8713) \\
UNI & 0.8108 (0.6655--0.9268) & 0.8843 (0.7677--0.9667) \\
Virchow & 0.7659 (0.5965--0.8974) & 0.8719 (0.6540--0.9770) \\
Prov-GigaPath & 0.7997 (0.6542--0.9163) & 0.7952 (0.6402--0.9200) \\
Hibou-L & 0.6903 (0.5276--0.8347) & 0.8446 (0.6972--0.9498) \\
mSTAR & \textbf{0.8305 (0.6882--0.9375)} & 0.8725 (0.7212--0.9640) \\
GPFM & 0.8018 (0.6316--0.9231) & 0.8675 (0.7301--0.9555) \\
Virchow2 & 0.7886 (0.6227--0.9255) & 0.9177 (0.7988--0.9863) \\
CHIEF & 0.7338 (0.5574--0.8772) & 0.7815 (0.6414--0.9018) \\
Phikon2 & 0.7787 (0.6403--0.8996) & 0.8318 (0.7047--0.9338) \\
CONCH1.5 & 0.7694 (0.6127--0.8977) & 0.7936 (0.6618--0.9022) \\
TITAN & 0.8114 (0.6858--0.9066) & 0.8137 (0.6959--0.9231) \\
MUSK & 0.6876 (0.4968--0.8494) & 0.8390 (0.6870--0.9526) \\
UNI2 & 0.7780 (0.6397--0.8947) & 0.9334 (0.8547--0.9855) \\
H-Optimus-1 & 0.7916 (0.6276--0.9194) & 0.9150 (0.8012--0.9869) \\
H0-mini & 0.7757 (0.6161--0.9036) & \textbf{0.9494 (0.8819--0.9916)} \\
EAGLE & 0.6820 (0.5317--0.8179) & 0.8363 (0.7227--0.9291) \\
LitePath & 0.8041 (0.6553--0.9202) & 0.8716 (0.7627--0.9554) \\
\midrule
LiteFM-S & 0.7630 (0.6080--0.8914) & 0.8800 (0.7757--0.9583) \\
LiteVirchow2 & 0.7794 (0.6151--0.9093) & 0.8723 (0.7649--0.9548) \\
LiteFM & 0.8072 (0.6578--0.9223) & 0.8696 (0.7621--0.9534) \\
LiteFM-L & 0.7959 (0.6451--0.9188) & 0.8465 (0.7302--0.9370) \\
\bottomrule
\end{tabular}
\end{table}

\begin{table}[!t]
\centering
\caption{Mean AUC (95\% CI) of models on TNM-N Staging (N0/N+) and pTNM Overall Staging (I/II/III) of breast cancer. One internal cohort (H2) is used for evaluation.}
\label{tab:zj1-c1_breast_tnm}
\begin{tabular}{lcc}
\toprule
Model & TNM-N Staging (N0/N+) & pTNM Overall Staging (I/II/III) \\
\midrule
Resnet50 & 0.4102 (0.2721--0.5560) & 0.5438 (0.4330--0.6525) \\
CTransPath & 0.6381 (0.4829--0.7879) & 0.6142 (0.5027--0.7206) \\
CONCH & 0.6741 (0.5209--0.8107) & 0.5560 (0.4489--0.6708) \\
Phikon & 0.6572 (0.4568--0.8069) & 0.5890 (0.4820--0.6917) \\
PLIP & 0.6397 (0.4982--0.7705) & 0.6015 (0.4906--0.7098) \\
UNI & 0.7035 (0.5637--0.8317) & 0.5743 (0.4631--0.6843) \\
Virchow & 0.6125 (0.4388--0.8043) & 0.5306 (0.4281--0.6407) \\
Prov-GigaPath & 0.6792 (0.5091--0.8185) & 0.5732 (0.4510--0.6897) \\
Hibou-L & 0.5502 (0.4112--0.6870) & 0.5279 (0.4234--0.6372) \\
mSTAR & 0.6885 (0.5410--0.8249) & 0.5686 (0.4559--0.6818) \\
GPFM & 0.6292 (0.4225--0.7969) & 0.5876 (0.4790--0.6923) \\
Virchow2 & \textbf{0.7949 (0.6375--0.9077)} & 0.5652 (0.4518--0.6765) \\
CHIEF & 0.7082 (0.5751--0.8333) & 0.5642 (0.4502--0.6787) \\
Phikon2 & 0.6731 (0.5285--0.8060) & 0.5923 (0.4828--0.7004) \\
CONCH1.5 & 0.6795 (0.5420--0.8047) & 0.6053 (0.4945--0.7185) \\
TITAN & 0.7031 (0.5767--0.8139) & \textbf{0.6614 (0.5672--0.7565)} \\
MUSK & 0.5034 (0.3744--0.6316) & 0.6026 (0.4946--0.7137) \\
UNI2 & 0.6865 (0.5372--0.8232) & 0.5326 (0.4210--0.6467) \\
H-Optimus-1 & 0.6565 (0.5068--0.7952) & 0.5684 (0.4626--0.6763) \\
H0-mini & 0.7172 (0.5826--0.8333) & 0.5690 (0.4580--0.6819) \\
EAGLE & 0.6483 (0.5029--0.7814) & 0.5693 (0.4694--0.6599) \\
LitePath & 0.7742 (0.6178--0.9034) & 0.6070 (0.4923--0.7170) \\
\midrule
LiteFM-S & 0.7665 (0.6247--0.8832) & 0.6083 (0.4954--0.7162) \\
LiteVirchow2 & 0.8129 (0.7004--0.9041) & 0.6004 (0.5011--0.7007) \\
LiteFM & 0.7657 (0.6262--0.8839) & 0.6013 (0.4909--0.7110) \\
LiteFM-L & 0.7607 (0.6263--0.8751) & 0.5991 (0.4746--0.7143) \\
\bottomrule
\end{tabular}
\end{table}

\begin{table}[!t]
\centering
\caption{Mean AUC (95\% CI) of models on Molecular Subtyping of breast cancer. One internal cohort (H2) and one external cohort (H9) are used for evaluation.}
\label{tab:zj1-c1_breast_molsubtype}
\begin{tabular}{lcc}
\toprule
Model & Internal (H2) & External (H9) \\
\midrule
Resnet50 & 0.8880 (0.8670--0.9077) & 0.7363 (0.7001--0.7706) \\
CTransPath & 0.9132 (0.8939--0.9316) & 0.7861 (0.7532--0.8171) \\
CONCH & 0.8992 (0.8775--0.9196) & 0.7622 (0.7292--0.7935) \\
Phikon & 0.9175 (0.8982--0.9356) & 0.7945 (0.7532--0.8277) \\
PLIP & 0.9092 (0.8881--0.9285) & 0.7475 (0.7092--0.7786) \\
UNI & 0.9239 (0.9049--0.9411) & 0.7996 (0.7592--0.8377) \\
Virchow & 0.9173 (0.8990--0.9343) & 0.7863 (0.7464--0.8281) \\
Prov-GigaPath & 0.9251 (0.9077--0.9416) & 0.8134 (0.7817--0.8433) \\
Hibou-L & 0.9088 (0.8881--0.9277) & 0.7329 (0.6631--0.7885) \\
mSTAR & 0.9225 (0.9017--0.9408) & 0.7859 (0.7453--0.8340) \\
GPFM & 0.9241 (0.9055--0.9410) & 0.7958 (0.7469--0.8323) \\
Virchow2 & 0.9249 (0.9065--0.9414) & 0.8202 (0.7850--0.8503) \\
CHIEF & 0.9144 (0.8949--0.9327) & 0.7690 (0.7342--0.8022) \\
Phikon2 & 0.9213 (0.9032--0.9383) & 0.7960 (0.7594--0.8322) \\
CONCH1.5 & 0.9182 (0.9006--0.9346) & 0.7815 (0.7455--0.8164) \\
TITAN & 0.8929 (0.8777--0.9086) & 0.7648 (0.7364--0.7921) \\
MUSK & 0.9178 (0.8993--0.9351) & 0.7203 (0.6763--0.7720) \\
UNI2 & 0.9313 (0.9148--0.9465) & 0.8188 (0.7816--0.8554) \\
H-Optimus-1 & \textbf{0.9380 (0.9232--0.9523)} & 0.7919 (0.7588--0.8315) \\
H0-mini & 0.9115 (0.8969--0.9256) & \textbf{0.8226 (0.7933--0.8495)} \\
EAGLE & 0.8925 (0.8760--0.9089) & 0.7728 (0.7443--0.8010) \\
LitePath & 0.9093 (0.8932--0.9242) & 0.8075 (0.7778--0.8363) \\
\midrule
LiteFM-S & 0.9055 (0.8881--0.9221) & 0.7961 (0.7668--0.8228) \\
LiteVirchow2 & 0.9112 (0.8957--0.9262) & 0.8112 (0.7830--0.8364) \\
LiteFM & 0.9106 (0.8947--0.9256) & 0.8037 (0.7733--0.8333) \\
LiteFM-L & 0.9137 (0.8971--0.9296) & 0.8214 (0.7926--0.8466) \\
\bottomrule
\end{tabular}
\end{table}

\begin{table}[!t]
\centering
\caption{Mean AUC (95\% CI) of models on IHC status prediction tasks (AR, ER, PR, HER2, CK5) of breast cancer. One internal cohort (H2) is used for evaluation.}
\label{tab:breast_ihc_all_h2}
\resizebox{\linewidth}{!}{
\begin{tabular}{lccccc}
\toprule
Model & AR & ER & PR & HER2 & CK5 \\
\midrule
Resnet50 & 0.7115 (0.6389--0.7812) & 0.8236 (0.7728--0.8697) & 0.8619 (0.8095--0.9078) & 0.7145 (0.6520--0.7732) & 0.7618 (0.6479--0.8552) \\
CTransPath & 0.7301 (0.6585--0.7978) & 0.8765 (0.8276--0.9174) & 0.8806 (0.8311--0.9213) & 0.8192 (0.7653--0.8678) & 0.8023 (0.7168--0.8761) \\
CONCH & 0.7345 (0.6501--0.8089) & 0.8850 (0.8432--0.9216) & 0.8707 (0.8272--0.9096) & 0.8041 (0.7453--0.8596) & 0.8076 (0.7120--0.8888) \\
Phikon & 0.7147 (0.6412--0.7847) & 0.8928 (0.8527--0.9293) & 0.8832 (0.8380--0.9233) & 0.8141 (0.7552--0.8660) & 0.8432 (0.7690--0.9069) \\
PLIP & 0.7318 (0.6621--0.7975) & 0.8551 (0.8093--0.8960) & 0.8887 (0.8483--0.9248) & 0.7983 (0.7427--0.8491) & 0.8035 (0.7208--0.8762) \\
UNI & 0.7292 (0.6549--0.7987) & 0.8952 (0.8571--0.9289) & 0.8914 (0.8497--0.9288) & 0.8176 (0.7563--0.8714) & 0.8300 (0.7521--0.8986) \\
Virchow & 0.6762 (0.5873--0.7676) & 0.8957 (0.8571--0.9304) & 0.8784 (0.8313--0.9202) & 0.8200 (0.7673--0.8681) & 0.8450 (0.7724--0.9085) \\
\bottomrule
\end{tabular}
}
\end{table}

\begin{table}[!t]
\centering
\caption{Mean C-index (95\% CI) of models on survival tasks (disease-free survival and overall survival) of breast cancer. One internal cohort (H2) and one external cohort (H9) are used for evaluation.}
\label{tab:zj1_breast_survival}
\begin{tabular}{lccc}
\toprule
Model & DFS (Internal H2) & DFS (External H9) & OS (Internal H2) \\
\midrule
Resnet50 & 0.5784 (0.3815--0.7776) & 0.3726 (0.1560--0.6923) & 0.6341 (0.4368--0.8040) \\
CTransPath & 0.6460 (0.4604--0.8149) & 0.6967 (0.1094--1.0000) & 0.6813 (0.4832--0.8488) \\
CONCH & 0.6352 (0.4627--0.7937) & 0.7449 (0.2615--0.9872) & 0.6648 (0.4716--0.8307) \\
Phikon & 0.6339 (0.4173--0.8550) & 0.6673 (0.2063--1.0000) & 0.6798 (0.4704--0.8557) \\
PLIP & 0.5539 (0.3518--0.7411) & 0.6376 (0.1587--0.9481) & 0.6378 (0.4494--0.8003) \\
UNI & 0.6493 (0.4263--0.8582) & 0.7184 (0.1746--1.0000) & 0.6603 (0.4430--0.8397) \\
Virchow & 0.6497 (0.4446--0.8348) & 0.5935 (0.1500--1.0000) & 0.6619 (0.4594--0.8280) \\
\bottomrule
\end{tabular}
\end{table}

\begin{table}[!t]
\centering
\caption{Mean AUC (95\% CI) of models on Cancer Grading of gastric cancer. One internal cohort (H1) and two external cohorts (H3, H4) are used for evaluation.}
\label{tab:nanfang_gastric_grade}
\begin{tabular}{lccc}
\toprule
Model & Internal (H1) & External (H3) & External (H4) \\
\midrule
Resnet50 & 0.6693 (0.4992--0.8235) & 0.7336 (0.6193--0.8283) & 0.7217 (0.5854--0.8292) \\
CTransPath & 0.8035 (0.6799--0.9075) & 0.8481 (0.7767--0.9053) & 0.8110 (0.7378--0.8754) \\
CONCH & 0.8377 (0.7312--0.9283) & 0.8438 (0.7809--0.9009) & \textbf{0.8626 (0.8053--0.9130)} \\
Phikon & 0.8165 (0.6850--0.9254) & 0.8214 (0.7534--0.8836) & 0.8223 (0.7602--0.8771) \\
PLIP & 0.7871 (0.6367--0.9106) & 0.8545 (0.7988--0.9037) & 0.8549 (0.7995--0.9040) \\
UNI & 0.8502 (0.7371--0.9405) & 0.8268 (0.7559--0.8897) & 0.8451 (0.7827--0.8986) \\
Virchow & 0.8678 (0.7686--0.9486) & 0.8285 (0.7593--0.8926) & 0.8380 (0.7759--0.8939) \\
\bottomrule
\end{tabular}
\end{table}

\begin{table}[!t]
\centering
\caption{Mean AUC (95\% CI) of models on Pathological Subtyping of gastric cancer. One internal cohort (H1) and one external cohort (H3) are used for evaluation.}
\label{tab:nanfang_gastric_pathsubtype}
\begin{tabular}{lcc}
\toprule
Model & Internal (H1) & External (H3) \\
\midrule
Resnet50 & 0.6028 (0.5014--0.7084) & 0.5312 (0.4525--0.6028) \\
CTransPath & 0.7328 (0.6386--0.8238) & 0.5752 (0.5004--0.6507) \\
CONCH & 0.8023 (0.7120--0.8823) & 0.6225 (0.5476--0.7384) \\
Phikon & 0.7538 (0.6563--0.8441) & 0.6558 (0.5774--0.7359) \\
PLIP & 0.6825 (0.5859--0.7794) & 0.5706 (0.4985--0.6492) \\
UNI & 0.8079 (0.6975--0.8939) & 0.6350 (0.5569--0.7168) \\
Virchow & 0.7933 (0.7023--0.8789) & 0.6377 (0.5425--0.7786) \\
Prov-GigaPath & 0.7663 (0.6551--0.8567) & 0.6309 (0.5467--0.7185) \\
Hibou-L & 0.6830 (0.5897--0.7776) & 0.5735 (0.4807--0.6977) \\
mSTAR & 0.8057 (0.6913--0.8972) & 0.6277 (0.5501--0.7108) \\
GPFM & 0.7908 (0.6904--0.8777) & 0.6855 (0.5964--0.7704) \\
Virchow2 & \textbf{0.8159 (0.7290--0.8943)} & 0.6813 (0.5946--0.7719) \\
CHIEF & 0.7426 (0.6348--0.8375) & 0.5879 (0.5011--0.6706) \\
Phikon2 & 0.7912 (0.6823--0.8802) & 0.6337 (0.5551--0.7170) \\
CONCH1.5 & 0.8016 (0.7179--0.8775) & \textbf{0.7367 (0.6520--0.8228)} \\
TITAN & 0.8003 (0.7245--0.8720) & 0.5653 (0.5355--0.5943) \\
MUSK & 0.7539 (0.6685--0.8349) & 0.6071 (0.5317--0.6866) \\
UNI2 & 0.7928 (0.7000--0.8770) & 0.6808 (0.5956--0.7640) \\
H-Optimus-1 & 0.8070 (0.7042--0.8925) & 0.7087 (0.5870--0.8172) \\
H0-mini & 0.7970 (0.7099--0.8809) & 0.6332 (0.5703--0.6960) \\
EAGLE & 0.6528 (0.5556--0.7500) & 0.6060 (0.5296--0.7007) \\
LitePath & 0.7770 (0.6831--0.8633) & 0.6400 (0.5683--0.7055) \\
\midrule
LiteFM-S & 0.7882 (0.6854--0.8806) & 0.6267 (0.5518--0.7053) \\
LiteVirchow2 & 0.7974 (0.7092--0.8824) & 0.6457 (0.5832--0.7051) \\
LiteFM & 0.7921 (0.6995--0.8762) & 0.6339 (0.5650--0.6996) \\
LiteFM-L & 0.7979 (0.7156--0.8806) & 0.6286 (0.5591--0.6992) \\
\bottomrule
\end{tabular}
\end{table}

\begin{table}[!t]
\centering
\caption{Mean AUC (95\% CI) of models on TNM-N Staging (N0/N+) of gastric cancer. One internal cohort (H1) and one external cohort (H3) are used for evaluation.}
\label{tab:nanfang_gastric_tnm-n}
\begin{tabular}{lcc}
\toprule
Model & Internal (H1) & External (H3) \\
\midrule
Resnet50 & 0.7481 (0.6312--0.8510) & 0.6572 (0.5784--0.7336) \\
CTransPath & 0.7995 (0.6966--0.8894) & 0.7552 (0.6857--0.8201) \\
CONCH & 0.7831 (0.6701--0.8831) & 0.7529 (0.6809--0.8198) \\
Phikon & 0.7700 (0.6419--0.8788) & 0.7608 (0.6899--0.8278) \\
PLIP & 0.7772 (0.6673--0.8782) & 0.6879 (0.6125--0.7602) \\
UNI & 0.7820 (0.6485--0.8887) & 0.7383 (0.6656--0.8080) \\
Virchow & 0.7768 (0.6543--0.8781) & 0.7327 (0.6584--0.8046) \\
Prov-GigaPath & 0.7652 (0.6454--0.8670) & 0.7523 (0.6801--0.8198) \\
Hibou-L & 0.7643 (0.6468--0.8659) & 0.7371 (0.6655--0.8046) \\
mSTAR & 0.7916 (0.6726--0.8894) & 0.7428 (0.6696--0.8122) \\
GPFM & 0.7848 (0.6637--0.8819) & 0.7476 (0.6726--0.8157) \\
Virchow2 & 0.7899 (0.6794--0.8846) & 0.7508 (0.6813--0.8183) \\
CHIEF & 0.7940 (0.6879--0.8856) & 0.7503 (0.6790--0.8177) \\
Phikon2 & 0.7805 (0.6604--0.8810) & 0.7570 (0.6890--0.8204) \\
CONCH1.5 & 0.7458 (0.6223--0.8540) & 0.7411 (0.6670--0.8115) \\
TITAN & 0.7328 (0.6089--0.8455) & 0.7104 (0.6374--0.7783) \\
MUSK & 0.7703 (0.6567--0.8718) & 0.7473 (0.6778--0.8130) \\
UNI2 & 0.7908 (0.6786--0.8870) & 0.7665 (0.6964--0.8322) \\
H-Optimus-1 & 0.7917 (0.6758--0.8888) & 0.7452 (0.6711--0.8139) \\
H0-mini & \textbf{0.8122 (0.7113--0.9024)} & 0.7503 (0.6790--0.8153) \\
EAGLE & 0.6773 (0.5558--0.7992) & \textbf{0.7748 (0.7120--0.8374)} \\
LitePath & 0.7996 (0.6917--0.8916) & 0.7538 (0.6837--0.8201) \\
\midrule
LiteFM-S & 0.7811 (0.6697--0.8794) & 0.7618 (0.6933--0.8249) \\
LiteVirchow2 & 0.7990 (0.6909--0.8953) & 0.7503 (0.6807--0.8157) \\
LiteFM & 0.7965 (0.6903--0.8907) & 0.7534 (0.6809--0.8210) \\
LiteFM-L & 0.8025 (0.6970--0.8971) & 0.7576 (0.6897--0.8210) \\
\bottomrule
\end{tabular}
\end{table}

\begin{table}[!t]
\centering
\caption{Mean AUC (95\% CI) of models on Perineural Invasion Detection of gastric cancer. One internal cohort (H1) and one external cohort (H3) are used for evaluation.}
\label{tab:nanfang_gastric_perineural}
\begin{tabular}{lcc}
\toprule
Model & Internal (H1) & External (H3) \\
\midrule
Resnet50 & 0.8385 (0.7293--0.9320) & 0.7781 (0.7003--0.8477) \\
CTransPath & 0.9138 (0.8370--0.9732) & 0.8401 (0.7694--0.8974) \\
CONCH & 0.9303 (0.8482--0.9886) & 0.8457 (0.7870--0.8974) \\
Phikon & 0.9181 (0.8330--0.9808) & 0.8482 (0.7901--0.9007) \\
PLIP & 0.9404 (0.8713--0.9884) & 0.8425 (0.7799--0.8981) \\
UNI & 0.9269 (0.8467--0.9863) & 0.8457 (0.7863--0.8974) \\
Virchow & 0.9171 (0.8333--0.9772) & 0.8397 (0.7772--0.8937) \\
Prov-GigaPath & 0.9121 (0.8290--0.9756) & 0.8341 (0.7703--0.8893) \\
Hibou-L & 0.9237 (0.8473--0.9808) & 0.8294 (0.7485--0.8907) \\
mSTAR & 0.9282 (0.8487--0.9876) & 0.8456 (0.7885--0.8972) \\
GPFM & 0.9284 (0.8519--0.9825) & 0.8493 (0.7921--0.9007) \\
Virchow2 & 0.9345 (0.8657--0.9844) & 0.8455 (0.7844--0.8992) \\
CHIEF & 0.9118 (0.8310--0.9733) & 0.8382 (0.7676--0.8975) \\
Phikon2 & 0.9250 (0.8495--0.9802) & 0.8409 (0.7814--0.8950) \\
CONCH1.5 & 0.9009 (0.8087--0.9725) & 0.8005 (0.7213--0.8679) \\
TITAN & \textbf{0.9415 (0.8815--0.9886)} & 0.8416 (0.7839--0.8957) \\
MUSK & 0.9252 (0.8479--0.9830) & 0.8421 (0.7733--0.9005) \\
UNI2 & 0.9191 (0.8372--0.9800) & 0.8461 (0.7853--0.8993) \\
H-Optimus-1 & 0.9226 (0.8433--0.9803) & 0.8351 (0.7698--0.8906) \\
H0-mini & 0.9207 (0.8413--0.9832) & 0.8455 (0.7865--0.8986) \\
EAGLE & 0.9158 (0.8459--0.9729) & 0.8193 (0.7596--0.8744) \\
LitePath & 0.9277 (0.8557--0.9828) & \textbf{0.8516 (0.7946--0.9041)} \\
\midrule
LiteFM-S & 0.9259 (0.8527--0.9811) & 0.8520 (0.7931--0.9035) \\
LiteVirchow2 & 0.9277 (0.8573--0.9843) & 0.8485 (0.7931--0.8991) \\
LiteFM & 0.9249 (0.8505--0.9811) & 0.8518 (0.7941--0.9033) \\
LiteFM-L & 0.9226 (0.8381--0.9879) & 0.8405 (0.7834--0.8940) \\
\bottomrule
\end{tabular}
\end{table}

\begin{table}[!t]
\centering
\caption{Mean AUC (95\% CI) of models on Vascular Invasion Detection of gastric cancer. One internal cohort (H1) and one external cohort (H3) are used for evaluation.}
\label{tab:nanfang_gastric_vascular}
\begin{tabular}{lcc}
\toprule
Model & Internal (H1) & External (H3) \\
\midrule
Resnet50 & 0.7549 (0.6353--0.8619) & 0.6593 (0.5826--0.7337) \\
CTransPath & 0.7633 (0.6476--0.8693) & 0.7367 (0.6649--0.8030) \\
CONCH & 0.7680 (0.6551--0.8705) & 0.7085 (0.6254--0.7845) \\
Phikon & 0.7730 (0.6577--0.8764) & 0.7256 (0.6517--0.7968) \\
PLIP & 0.7706 (0.6592--0.8744) & 0.6861 (0.6100--0.7584) \\
UNI & 0.7475 (0.6287--0.8578) & 0.7099 (0.6299--0.7833) \\
Virchow & \textbf{0.8019 (0.6956--0.8964)} & 0.7140 (0.6354--0.7883) \\
Prov-GigaPath & 0.7759 (0.6630--0.8770) & 0.7193 (0.6409--0.7949) \\
Hibou-L & 0.7332 (0.5957--0.8517) & 0.7279 (0.6540--0.7968) \\
mSTAR & 0.7642 (0.6468--0.8697) & 0.7226 (0.6435--0.7959) \\
GPFM & 0.7715 (0.6492--0.8786) & 0.7345 (0.6596--0.8035) \\
Virchow2 & 0.7677 (0.6456--0.8756) & 0.7345 (0.6559--0.8061) \\
CHIEF & 0.7663 (0.6528--0.8703) & 0.7182 (0.6443--0.7883) \\
Phikon2 & 0.7782 (0.6545--0.8858) & \textbf{0.7479 (0.6774--0.8121)} \\
CONCH1.5 & 0.7321 (0.5901--0.8538) & 0.6922 (0.6057--0.7752) \\
TITAN & 0.7721 (0.6641--0.8710) & 0.6866 (0.6125--0.7570) \\
MUSK & 0.7655 (0.6519--0.8726) & 0.7188 (0.6449--0.7885) \\
UNI2 & 0.7817 (0.6701--0.8825) & 0.7044 (0.6199--0.7831) \\
H-Optimus-1 & 0.7637 (0.6435--0.8717) & 0.7182 (0.6355--0.7934) \\
H0-mini & 0.7721 (0.6609--0.8755) & 0.7341 (0.6563--0.8026) \\
EAGLE & 0.7926 (0.6906--0.8824) & 0.7397 (0.6659--0.8039) \\
LitePath & 0.7879 (0.6772--0.8875) & 0.7399 (0.6646--0.8092) \\
\midrule
LiteFM-S & 0.7846 (0.6609--0.8870) & 0.7387 (0.6484--0.8096) \\
LiteVirchow2 & 0.7960 (0.6903--0.8918) & 0.7497 (0.6808--0.8144) \\
LiteFM & 0.7853 (0.6750--0.8854) & 0.7401 (0.6658--0.8086) \\
LiteFM-L & 0.7904 (0.6831--0.8851) & 0.7418 (0.6690--0.8095) \\
\bottomrule
\end{tabular}
\end{table}

\begin{table}[!t]
\centering
\caption{Mean AUC (95\% CI) of models on two gastric cancer tasks (Normal/Abnormal Classification and Intestinal Metaplasia) of gastric cancer. One internal cohort (H7) is used for evaluation for both tasks.}
\label{tab:stomach_four_tasks_h7}
\begin{tabular}{lcc}
\toprule
Model & Normal/Abnormal & Intestinal Metaplasia \\
\midrule
Resnet50 & 0.8704 (0.8319--0.9052) & 0.9354 (0.8899--0.9686) \\
CTransPath & 0.9206 (0.8925--0.9450) & 0.9422 (0.8909--0.9720) \\
CONCH & 0.8987 (0.8576--0.9299) & 0.9535 (0.9196--0.9755) \\
Phikon & 0.9161 (0.8848--0.9438) & 0.9495 (0.9135--0.9733) \\
PLIP & 0.8773 (0.8364--0.9135) & 0.9225 (0.8544--0.9629) \\
UNI & 0.9187 (0.8820--0.9454) & 0.9421 (0.8940--0.9713) \\
Virchow & 0.9012 (0.8628--0.9337) & 0.9554 (0.9208--0.9766) \\
Prov-GigaPath & 0.9103 (0.8751--0.9402) & 0.9515 (0.9220--0.9726) \\
Hibou-L & 0.9006 (0.8668--0.9306) & 0.9569 (0.9216--0.9758) \\
mSTAR & 0.9226 (0.8897--0.9480) & 0.9465 (0.8869--0.9736) \\
GPFM & 0.9134 (0.8798--0.9414) & 0.9311 (0.8470--0.9734) \\
Virchow2 & 0.9232 (0.8908--0.9484) & 0.9615 (0.9337--0.9792) \\
CHIEF & 0.9143 (0.8480--0.9449) & 0.9364 (0.8817--0.9673) \\
Phikon2 & 0.9126 (0.8834--0.9391) & 0.9534 (0.9238--0.9761) \\
CONCH1.5 & 0.9045 (0.8700--0.9345) & 0.9277 (0.8715--0.9678) \\
TITAN & 0.9185 (0.8870--0.9481) & 0.9502 (0.9166--0.9769) \\
MUSK & 0.8991 (0.8685--0.9264) & 0.9322 (0.8896--0.9630) \\
UNI2 & 0.9118 (0.8763--0.9437) & 0.9555 (0.9245--0.9775) \\
H-Optimus-1 & 0.9319 (0.9017--0.9559) & \textbf{0.9631 (0.9426--0.9780)} \\
H0-mini & 0.9134 (0.8822--0.9423) & 0.9503 (0.9063--0.9742) \\
EAGLE & \textbf{0.9407 (0.9152--0.9634)} & 0.9209 (0.8657--0.9628) \\
LitePath & 0.9268 (0.9013--0.9494) & 0.9617 (0.9406--0.9779) \\
\midrule
LiteFM-S & 0.9214 (0.8921--0.9456) & 0.9643 (0.9450--0.9797) \\
LiteVirchow2 & 0.9231 (0.8919--0.9488) & 0.9627 (0.9390--0.9788) \\
LiteFM & 0.9261 (0.9011--0.9484) & 0.9609 (0.9388--0.9774) \\
LiteFM-L & 0.9273 (0.8980--0.9523) & 0.9579 (0.9294--0.9777) \\
\bottomrule
\end{tabular}
\end{table}

\begin{table}[!t]
\centering
\caption{Mean AUC (95\% CI) of models on IHC Status Prediction — HER2 and S-100 of gastric cancer. One internal cohort (H1+H3+H4) is used for evaluation.}
\label{tab:stomach_ihc_her2_s100}
\begin{tabular}{lcc}
\toprule
Model & HER2 & S-100 \\
\midrule
Resnet50 & 0.4919 (0.3634--0.6189) & 0.7483 (0.6007--0.8776) \\
CTransPath & 0.5205 (0.3963--0.6440) & 0.8181 (0.6728--0.9333) \\
CONCH & 0.5317 (0.3975--0.6660) & 0.8477 (0.7216--0.9449) \\
Phikon & 0.5899 (0.4587--0.7151) & 0.8219 (0.6687--0.9400) \\
PLIP & 0.5362 (0.4018--0.6657) & 0.8046 (0.6634--0.9197) \\
UNI & 0.5727 (0.4242--0.7178) & 0.8254 (0.6687--0.9433) \\
Virchow & 0.5429 (0.4190--0.6659) & 0.8374 (0.6745--0.9487) \\
Prov-GigaPath & 0.5929 (0.4683--0.7116) & 0.8261 (0.6605--0.9404) \\
Hibou-L & 0.5671 (0.4440--0.6871) & 0.8172 (0.6745--0.9356) \\
mSTAR & 0.5625 (0.3878--0.7089) & 0.8335 (0.6988--0.9422) \\
GPFM & 0.5945 (0.4617--0.7213) & 0.8135 (0.6667--0.9347) \\
Virchow2 & 0.5447 (0.4212--0.6663) & 0.8502 (0.7096--0.9568) \\
CHIEF & 0.5357 (0.4105--0.6574) & 0.7976 (0.6375--0.9231) \\
Phikon2 & 0.5590 (0.4262--0.6803) & 0.7548 (0.5979--0.8918) \\
CONCH1.5 & 0.6179 (0.5011--0.7342) & 0.7542 (0.5798--0.8987) \\
TITAN & 0.6031 (0.4980--0.7223) & \textbf{0.8718 (0.7692--0.9489)} \\
MUSK & 0.5291 (0.4102--0.6493) & 0.8096 (0.6643--0.9305) \\
UNI2 & 0.5607 (0.4309--0.6869) & 0.8347 (0.6177--0.9618) \\
H-Optimus-1 & 0.5153 (0.3844--0.6425) & 0.7812 (0.6121--0.9194) \\
H0-mini & 0.5296 (0.3879--0.6613) & 0.8112 (0.6409--0.9409) \\
EAGLE & \textbf{0.6221 (0.5210--0.7318)} & 0.7603 (0.6186--0.8885) \\
LitePath & 0.5562 (0.4257--0.6779) & 0.8256 (0.6825--0.9351) \\
\midrule
LiteFM-S & 0.5272 (0.3906--0.6547) & 0.8215 (0.6750--0.9370) \\
LiteVirchow2 & 0.5205 (0.3285--0.6838) & 0.7774 (0.5854--0.9264) \\
LiteFM & 0.5482 (0.4133--0.6731) & 0.8135 (0.6669--0.9329) \\
LiteFM-L & 0.5456 (0.4016--0.6827) & 0.8211 (0.6667--0.9421) \\
\bottomrule
\end{tabular}
\end{table}

\begin{table}[!t]
\centering
\caption{Mean C-index (95\% CI) of models on survival tasks (disease-free survival and overall survival) of gastric cancer. One internal cohort (H3+H4) is used for evaluation for both tasks.}
\label{tab:ynch_gastric_survival}
\begin{tabular}{lcc}
\toprule
Model & DFS & OS \\
\midrule
Resnet50 & 0.7152 (0.5462--0.8283) & 0.6358 (0.5369--0.7281) \\
CTransPath & 0.7178 (0.5364--0.8404) & 0.6370 (0.5337--0.7353) \\
CONCH & 0.7204 (0.5405--0.8321) & 0.6479 (0.5471--0.7385) \\
Phikon & 0.7188 (0.5181--0.8340) & 0.6395 (0.5229--0.7406) \\
PLIP & 0.7030 (0.5283--0.8208) & 0.6307 (0.5169--0.7345) \\
UNI & 0.7223 (0.5314--0.8367) & 0.6461 (0.5171--0.7532) \\
Virchow & 0.7039 (0.5184--0.8238) & 0.6422 (0.5149--0.7503) \\
Prov-GigaPath & 0.7284 (0.5287--0.8410) & 0.6405 (0.5180--0.7495) \\
Hibou-L & 0.7271 (0.5451--0.8472) & 0.6437 (0.5014--0.7581) \\
mSTAR & 0.7228 (0.5251--0.8375) & 0.6453 (0.5198--0.7524) \\
GPFM & 0.7188 (0.4981--0.8461) & 0.6400 (0.5228--0.7447) \\
Virchow2 & 0.7242 (0.5249--0.8426) & 0.6641 (0.5487--0.7665) \\
CHIEF & 0.7170 (0.5232--0.8355) & 0.6347 (0.5207--0.7338) \\
Phikon2 & 0.7196 (0.4911--0.8470) & 0.6413 (0.5319--0.7394) \\
CONCH1.5 & 0.7330 (0.5599--0.8400) & 0.6640 (0.5655--0.7560) \\
TITAN & 0.7143 (0.5084--0.8493) & \textbf{0.6740 (0.5830--0.7580)} \\
MUSK & 0.7132 (0.5366--0.8306) & 0.6530 (0.5478--0.7599) \\
UNI2 & \textbf{0.7340 (0.5404--0.8481)} & 0.6490 (0.5403--0.7530) \\
H-Optimus-1 & 0.7310 (0.5438--0.8450) & 0.6600 (0.5390--0.7644) \\
H0-mini & 0.7121 (0.5048--0.8384) & 0.6578 (0.5433--0.7607) \\
EAGLE & 0.6992 (0.5244--0.8492) & 0.6395 (0.5401--0.7361) \\
LitePath & 0.7129 (0.4872--0.8480) & 0.6600 (0.5504--0.7628) \\
\midrule
LiteFM-S & 0.7077 (0.5083--0.8414) & 0.6580 (0.5504--0.7569) \\
LiteVirchow2 & 0.7072 (0.4988--0.8413) & 0.6571 (0.5492--0.7580) \\
LiteFM & 0.7111 (0.5079--0.8413) & 0.6596 (0.5520--0.7598) \\
LiteFM-L & 0.7234 (0.5113--0.8536) & 0.6570 (0.5462--0.7602) \\
\bottomrule
\end{tabular}
\end{table}

\begin{table}[!t]
\centering
\caption{Mean AUC (95\% CI) of models on four colon cancer tasks (TNM-N Staging N0/N+, TNM-T Staging T1+T2/T3+T4, TNM-T Staging T1/T2/T3/T4, and Consensus Molecular Subtyping) of colorectal cancer. One internal cohort (H8) is used for evaluation for all tasks.}
\label{tab:colon_five_tasks_h8}
\resizebox{\linewidth}{!}{
\begin{tabular}{lcccc}
\toprule
Model & TNM-N (N0/N+) & TNM-T (T1+T2/T3+T4) & TNM-T (T1/T2/T3/T4) & CMS \\
\midrule
Resnet50 & 0.8553 (0.7826--0.9166) & 0.8556 (0.7330--0.9519) & 0.7797 (0.6815--0.8637) & 0.5919 (0.5120--0.6726) \\
CTransPath & \textbf{0.9126 (0.8566--0.9586)} & 0.8531 (0.7064--0.9626) & 0.8072 (0.6977--0.8932) & 0.7498 (0.6725--0.8223) \\
CONCH & 0.8832 (0.8113--0.9420) & 0.8868 (0.7624--0.9751) & 0.8466 (0.7575--0.9155) & 0.7344 (0.6607--0.8042) \\
Phikon & 0.8834 (0.8190--0.9390) & 0.9067 (0.8046--0.9782) & 0.8528 (0.7767--0.9129) & 0.7572 (0.6824--0.8269) \\
PLIP & 0.8929 (0.8311--0.9448) & 0.8828 (0.7437--0.9817) & \textbf{0.8881 (0.8154--0.9440)} & 0.7015 (0.6186--0.7814) \\
UNI & 0.8961 (0.8310--0.9500) & 0.9052 (0.8028--0.9798) & 0.8455 (0.7650--0.9091) & 0.7567 (0.6788--0.8307) \\
Virchow & 0.8794 (0.8036--0.9375) & 0.9049 (0.8038--0.9789) & 0.8615 (0.7937--0.9207) & 0.7517 (0.6773--0.8201) \\
\bottomrule
\end{tabular}
}
\end{table}

\begin{table}[!t]
\centering
\caption{Mean C-index (95\% CI) of models on survival tasks (disease-free survival, overall survival, and disease-specific survival) of colorectal cancer. One internal cohort (H8) is used for DFS and OS evaluation, and one internal cohort (H1) is used for DSS evaluation.}
\label{tab:colorectal_survival}
\begin{tabular}{lccc}
\toprule
Model & DFS & OS & DSS \\
\midrule
Resnet50 & 0.7079 (0.5860--0.8208) & 0.7230 (0.6075--0.8279) & 0.6638 (0.3938--0.8919) \\
CTransPath & 0.7371 (0.6131--0.8451) & \textbf{0.7575 (0.6694--0.8366)} & 0.7313 (0.4674--0.9318) \\
CONCH & 0.7195 (0.6066--0.8339) & 0.7471 (0.6469--0.8357) & 0.7253 (0.4490--0.9360) \\
Phikon & 0.7285 (0.5963--0.8576) & 0.7288 (0.6273--0.8176) & 0.7238 (0.4874--0.9157) \\
PLIP & 0.7094 (0.5946--0.8270) & 0.7244 (0.6052--0.8220) & 0.6962 (0.4074--0.9136) \\
UNI & 0.7328 (0.6030--0.8577) & 0.7462 (0.6455--0.8360) & 0.7549 (0.5457--0.9354) \\
Virchow & 0.7130 (0.5928--0.8547) & 0.7307 (0.6254--0.8217) & 0.7615 (0.5232--0.9459) \\
\bottomrule
\end{tabular}
\end{table}

\begin{table}[!t]
\centering
\caption{Mean AUC (95\% CI) of models on five ROI classification tasks (CRC-MSI, PCAM, PanCancer-TIL, UniToPatho, WSSS4LUAD).}
\label{tab:roi_classification}
\resizebox{\linewidth}{!}{
\begin{tabular}{lcccccc}
\toprule
Model & CRC-MSI & PCAM & PanCancer-TIL & UniToPatho & WSSS4LUAD & Overall \\
\midrule
UNI2 & \textbf{0.8596 (0.8542--0.8650)} & \textbf{0.9868 (0.9854--0.9883)} & 0.9793 (0.9780--0.9804) & 0.8514 (0.8407--0.8609) & 0.9937 (0.9916--0.9955) & \textbf{0.9341 (0.8454--0.9948)} \\
Virchow2 & 0.8356 (0.8246--0.8449) & 0.9817 (0.9775--0.9851) & 0.9796 (0.9778--0.9812) & \textbf{0.8751 (0.8547--0.8945)} & 0.9943 (0.9924--0.9960) & 0.9332 (0.8278--0.9953) \\
LiteFM & 0.8448 (0.8383--0.8508) & 0.9822 (0.9806--0.9841) & 0.9784 (0.9771--0.9796) & 0.8449 (0.8347--0.8535) & 0.9938 (0.9919--0.9955) & 0.9288 (0.8386--0.9949) \\
H-Optimus-1 & 0.8375 (0.8279--0.8457) & 0.9813 (0.9795--0.9835) & \textbf{0.9802 (0.9789--0.9814)} & 0.8474 (0.8367--0.8577) & 0.9928 (0.9905--0.9949) & 0.9279 (0.8310--0.9942) \\
GPFM & 0.8081 (0.8016--0.8145) & 0.9832 (0.9751--0.9876) & 0.9783 (0.9771--0.9794) & 0.8358 (0.8260--0.8445) & \textbf{0.9947 (0.9930--0.9963)} & 0.9200 (0.8044--0.9957) \\
Prov-GigaPath & 0.8245 (0.8152--0.8361) & 0.9779 (0.9727--0.9813) & 0.9660 (0.9640--0.9680) & 0.8363 (0.8271--0.8446) & 0.9923 (0.9897--0.9946) & 0.9194 (0.8178--0.9937) \\
Hibou-L & 0.8004 (0.7924--0.8079) & 0.9828 (0.9790--0.9853) & 0.9729 (0.9710--0.9750) & 0.8459 (0.8357--0.8553) & 0.9930 (0.9908--0.9949) & 0.9190 (0.7952--0.9941) \\
\bottomrule
\end{tabular}
}
\end{table}

\begin{table}[!t]
\centering
\caption{Mean AUC (95\% CI) on BRACS-3 and BRACS-7.}
\label{tab:public_bracs_subtyping}
\begin{tabular}{lcc}
\toprule
Model & BRACS-3 & BRACS-7 \\
\midrule
Resnet50 & 0.8140 (0.7552--0.8700) & 0.7129 (0.6596--0.7643) \\
CTransPath & 0.8647 (0.8059--0.9194) & 0.7884 (0.7335--0.8383) \\
CONCH & 0.8836 (0.8295--0.9340) & 0.8157 (0.7672--0.8595) \\
Phikon & 0.8846 (0.8271--0.9344) & 0.8147 (0.7555--0.8655) \\
PLIP & 0.8585 (0.7981--0.9155) & 0.7354 (0.6801--0.7865) \\
UNI & 0.8833 (0.8277--0.9326) & 0.8142 (0.7587--0.8649) \\
Virchow & 0.8809 (0.8225--0.9319) & 0.7976 (0.7385--0.8521) \\
Prov-GigaPath & 0.8776 (0.8153--0.9336) & 0.8049 (0.7523--0.8549) \\
Hibou-L & 0.8799 (0.8210--0.9320) & 0.7963 (0.7411--0.8478) \\
mSTAR & 0.8875 (0.8306--0.9363) & 0.8220 (0.7700--0.8684) \\
GPFM & 0.8892 (0.8297--0.9382) & 0.8224 (0.7731--0.8662) \\
Virchow2 & 0.8766 (0.8117--0.9299) & 0.8144 (0.7655--0.8587) \\
CHIEF & 0.8715 (0.8137--0.9221) & 0.7871 (0.7284--0.8436) \\
Phikon2 & 0.8887 (0.8326--0.9360) & 0.8195 (0.7671--0.8680) \\
CONCH1.5 & 0.8802 (0.8235--0.9323) & 0.8177 (0.7730--0.8594) \\
TITAN & 0.8918 (0.8357--0.9394) & 0.8096 (0.7621--0.8586) \\
MUSK & 0.8830 (0.8312--0.9330) & 0.7705 (0.7155--0.8224) \\
UNI2 & 0.8805 (0.8153--0.9348) & 0.8220 (0.7681--0.8672) \\
H-Optimus-1 & 0.8810 (0.8189--0.9332) & \textbf{0.8241 (0.7739--0.8692)} \\
H0-mini & 0.8940 (0.8387--0.9413) & 0.8194 (0.7700--0.8655) \\
EAGLE & \textbf{0.9055 (0.8496--0.9507)} & 0.8160 (0.7641--0.8625) \\
LitePath & 0.8961 (0.8420--0.9446) & 0.8204 (0.7709--0.8647) \\
\midrule
LiteFM & 0.8923 (0.8386--0.9412) & 0.8210 (0.7711--0.8652) \\
\bottomrule
\end{tabular}
\end{table}

\begin{table}[!t]
\centering
\caption{Mean AUC (95\% CI) on CAMELYON.}
\label{tab:public_camelyon}
\begin{tabular}{lc}
\toprule
Model & CAMELYON \\
\midrule
Resnet50 & 0.8928 (0.8319--0.9488) \\
CTransPath & 0.9486 (0.9017--0.9842) \\
CONCH & 0.9766 (0.9410--0.9958) \\
Phikon & 0.9709 (0.9357--0.9958) \\
PLIP & 0.9413 (0.8862--0.9800) \\
UNI & 0.9640 (0.9155--0.9943) \\
Virchow & 0.9708 (0.9266--0.9940) \\
Prov-GigaPath & 0.9658 (0.9226--0.9925) \\
Hibou-L & 0.9797 (0.9471--0.9983) \\
mSTAR & 0.9693 (0.9247--0.9952) \\
GPFM & 0.9815 (0.9462--0.9997) \\
Virchow2 & 0.9855 (0.9639--0.9995) \\
CHIEF & 0.9434 (0.8988--0.9779) \\
Phikon2 & 0.9513 (0.9022--0.9897) \\
CONCH1.5 & 0.9562 (0.9172--0.9861) \\
TITAN & 0.9375 (0.8944--0.9755) \\
MUSK & 0.9514 (0.9071--0.9857) \\
UNI2 & 0.9863 (0.9478--0.9999) \\
H-Optimus-1 & \textbf{0.9892 (0.9692--1.0000)} \\
H0-mini & 0.9743 (0.9345--0.9957) \\
EAGLE & 0.8870 (0.8270--0.9384) \\
LitePath & 0.9737 (0.9456--0.9939) \\
\midrule
LiteFM & 0.9690 (0.9368--0.9923) \\
\bottomrule
\end{tabular}
\end{table}

\begin{table}[!t]
\centering
\caption{Mean AUC (95\% CI) on TCGA NSCLC, LUAD EGFR Prediction, LUAD TP53 Prediction, and CRC MolSub.}
\label{tab:public_tcga_classification}
\resizebox{\linewidth}{!}{
\begin{tabular}{lcccc}
\toprule
Model & NSCLC & LUAD\_EGFR & LUAD\_TP53 & CRC MolSub \\
\midrule
Resnet50 & 0.9224 (0.8878--0.9518) & 0.5996 (0.3643--0.8241) & 0.5622 (0.4385--0.6883) & 0.6287 (0.5406--0.7146) \\
CTransPath & 0.9642 (0.9391--0.9828) & 0.7275 (0.5403--0.9062) & 0.6948 (0.5705--0.8095) & 0.7915 (0.7230--0.8543) \\
CONCH & 0.9712 (0.9481--0.9865) & 0.6720 (0.4435--0.8887) & 0.6988 (0.5747--0.8128) & 0.7853 (0.7095--0.8537) \\
Phikon & 0.9661 (0.9398--0.9856) & 0.7275 (0.5220--0.9193) & 0.6728 (0.5430--0.8000) & 0.8269 (0.7619--0.8825) \\
PLIP & 0.9319 (0.8980--0.9599) & 0.5805 (0.3288--0.8345) & 0.6221 (0.4943--0.7488) & 0.7218 (0.6331--0.7948) \\
UNI & 0.9699 (0.9442--0.9874) & 0.7516 (0.5208--0.9529) & 0.7337 (0.6121--0.8432) & 0.8156 (0.7525--0.8743) \\
Virchow & 0.9682 (0.9444--0.9861) & 0.6049 (0.3809--0.8171) & 0.6871 (0.5512--0.8150) & 0.8030 (0.7271--0.8664) \\
\bottomrule
\end{tabular}
}
\end{table}

\begin{table}[!t]
\centering
\caption{Mean C-index (95\% CI) on survival analysis of TCGA CRC, HNSC, LUAD, and SKCM.}
\label{tab:public_survival}
\resizebox{\linewidth}{!}{
\begin{tabular}{lcccc}
\toprule
Model & TCGA-CRC & TCGA-HNSC & TCGA-LUAD & TCGA-SKCM \\
\midrule
Resnet50 & 0.5861 (0.4241--0.7284) & 0.6015 (0.4881--0.7172) & 0.6118 (0.4684--0.7450) & 0.5840 (0.4247--0.7361) \\
CTransPath & 0.6961 (0.5443--0.8322) & 0.6423 (0.5448--0.7325) & 0.6440 (0.5021--0.7809) & 0.5856 (0.4427--0.7202) \\
CONCH & \textbf{0.7205 (0.5569--0.8585)} & 0.6278 (0.5130--0.7347) & 0.6443 (0.4898--0.7711) & 0.6130 (0.5026--0.7195) \\
Phikon & 0.7040 (0.5419--0.8385) & 0.6319 (0.5171--0.7506) & 0.6459 (0.5081--0.7743) & 0.6052 (0.4559--0.7323) \\
PLIP & 0.6722 (0.5217--0.8091) & 0.6268 (0.5122--0.7267) & 0.6253 (0.4891--0.7440) & 0.5928 (0.4193--0.7268) \\
UNI & 0.7074 (0.5550--0.8549) & 0.6493 (0.5473--0.7483) & 0.6368 (0.5018--0.7600) & 0.6115 (0.4856--0.7289) \\
Virchow & 0.6609 (0.5121--0.8011) & 0.6076 (0.4980--0.7158) & 0.6372 (0.4866--0.7726) & 0.6144 (0.4758--0.7304) \\
\bottomrule
\end{tabular}
}
\end{table}

\end{document}